\documentclass[letterpaper, 10pt, conference, runningheads]{svproc}

\usepackage{moreverb,url}
\usepackage[colorlinks,bookmarksopen,bookmarksnumbered,citecolor=red,urlcolor=red]{hyperref}
\usepackage{graphicx}
\usepackage{graphics} % for pdf, bitmapped graphics files
\usepackage{epsfig} % for postscript graphics files
\usepackage{times} % assumes new font selection scheme installed
\usepackage{amsmath} % assumes amsmath package installed
\usepackage{amssymb}  % assumes amsmath package installed
\usepackage{mathtools}
\usepackage{algorithm}
\usepackage{algpseudocode}
\usepackage{booktabs}
\usepackage{float}

\usepackage[hang,flushmargin]{footmisc}

\newtheorem{assumption}{Assumption}

\algdef{SE}[DOWHILE]{Do}{doWhile}{\algorithmicdo}[1]{\algorithmicwhile\ #1}%

\usepackage{comment}

\newcommand{\robotradius}{r} % added
\newcommand{\sensorrange}{R} % added

\newcommand{\robotposition}{\mathbf{x}} % added
\newcommand{\robotpositionmodel}{\mathbf{y}} % added

\newcommand{\robotpositionunicycle}{\overline{\mathbf{x}}} % added
\newcommand{\robotorientation}{\psi} % added
\newcommand{\robotpositionunicyclemodel}{\overline{\mathbf{y}}} % added
\newcommand{\robotorientationmodel}{\varphi} % added

\newcommand{\goalposition}{\mathbf{x}_d} % added
\newcommand{\goalpositionmodel}{\mathbf{y}_d}

\newcommand{\enclosingworkspace}{\mathcal{W}_e} % added
\newcommand{\workspace}{\mathcal{W}} % added
\newcommand{\enclosingfreespace}{\mathcal{F}_e} % added
\newcommand{\freespace}{\mathcal{F}} % added
\newcommand{\freespacesemantic}{\mathcal{F}_{sem}^\hybridmode} % added
\newcommand{\freespacemapped}{\mathcal{F}_{map}^\hybridmode} % added
\newcommand{\freespacemappedhat}{\hat{\mathcal{F}}_{map}^\hybridmode} % added
\newcommand{\freespacemodel}{\mathcal{F}_{model}^\hybridmode} % added

\newcommand{\obstacle}{\tilde{O}} % added
\newcommand{\obstacleset}{\tilde{\mathcal{O}}} % added
\newcommand{\obstacledilated}{O} % added
\newcommand{\obstaclesetdilated}{\mathcal{O}} % added

\newcommand{\knownobstacle}{\tilde{P}} % added
\newcommand{\knownobstacleset}{\tilde{\mathcal{P}}} % added
\newcommand{\knownobstaclesetindex}{\mathcal{N}_{\mathcal{P}}}
\newcommand{\knownobstaclesetphysical}{\knownobstacleset_{\hybridmode}} % added
\newcommand{\knownobstaclesetdilatedmappedindex}{\mathcal{J}^\hybridmode}
\newcommand{\knownobstaclesetdilatedmappedintrusionindex}{\mathcal{J}_{\mathcal{B}}^\hybridmode}
\newcommand{\knownobstaclesetdilatedmappeddiskindex}{\mathcal{J}_{\mathcal{D}}^\hybridmode}
\newcommand{\knownobstacledilated}{P} % added
\newcommand{\knownobstaclesetdilated}{\mathcal{P}} % added
\newcommand{\knownobstaclecardinality}{N_P} % added

\newcommand{\knownobstaclesetdilatedsemantic}{\mathcal{P}_{sem}^\hybridmode} % added
\newcommand{\knownobstaclesetdilatedmapped}{\mathcal{P}_{map}^\hybridmode} % added
\newcommand{\knownobstacledilatedmappedintrusion}{B} % added
\newcommand{\knownobstaclesetdilatedmappedintrusion}{\mathcal{B}_{map}^\hybridmode} % added
\newcommand{\knownobstacledilatedmappeddisk}{D} % added
\newcommand{\knownobstaclesetdilatedmappeddisk}{\mathcal{D}_{map}^\hybridmode} % added

\newcommand{\unknownobstacle}{\tilde{C}} % added
\newcommand{\unknownobstacleset}{\tilde{\mathcal{C}}} % added
\newcommand{\unknownobstaclesetindex}{\mathcal{N}_{\mathcal{C}}}
\newcommand{\unknownobstacledilated}{C} % added
\newcommand{\unknownobstaclesetdilated}{\mathcal{C}} % added
\newcommand{\unknownobstaclecardinality}{N_C} % added

\newcommand{\unknownobstaclesetdilatedsemantic}{\mathcal{C}_{sem}} % added
\newcommand{\unknownobstaclesetdilatedsemanticindex}{\mathcal{J}_{\mathcal{C}}}
\newcommand{\unknownobstaclesetdilatedmapped}{\mathcal{C}_{map}} % added

\newcommand{\betaclearance}{\varepsilon}
\newcommand{\implicitgeneric}{\beta}

\newcommand{\diffeogeneric}{\mathbf{h}} % added
\newcommand{\diffeo}{\mathbf{h}^{\hybridmode}} % added
\newcommand{\diffeounicycle}{\overline{\diffeogeneric}^{\hybridmode}}
\newcommand{\diffeopurging}[1]{\mathbf{h}^{\hybridmode}_{#1}} % added
\newcommand{\diffeocomposition}{\mathbf{g}^{\hybridmode}} % added
\newcommand{\diffeoroot}{\hat{\mathbf{h}}^{\hybridmode}} % added

\newcommand{\diffeocenter}[1]{\mathbf{x}_{#1}^*} % added
\newcommand{\diffeoradius}[1]{\rho_{#1}} % added

\newcommand{\triangletree}[1]{\mathcal{T}_{#1}} % added
\newcommand{\trianglevertices}[1]{\mathcal{V}_{#1}} % added
\newcommand{\triangleedges}[1]{\mathcal{E}_{#1}} % added
\newcommand{\freespacemappedpurging}[1]{\mathcal{F}_{map,#1}^\hybridmode} % added
\newcommand{\innerpolygon}[1]{\mathcal{Q}_{#1}} % added
\newcommand{\outerpolygon}[1]{\overline{\mathcal{Q}}_{#1}} % added
\newcommand{\innerpolygonimplicit}[1]{\gamma_{#1}} % added
\newcommand{\outerpolygonimplicit}[1]{\delta_{#1}} % added
\newcommand{\innerpolygonsigma}[1]{\sigma_{\innerpolygonimplicit{#1}}} % added
\newcommand{\outerpolygonsigma}[1]{\sigma_{\outerpolygonimplicit{#1}}} % added
\newcommand{\innerpolygontune}[1]{\mu_{\innerpolygonimplicit{#1}}}
\newcommand{\innerpolygondistance}[1]{\epsilon_{#1}}
\newcommand{\outerpolygontune}[1]{\mu_{\outerpolygonimplicit{#1}}}
\newcommand{\switch}[1]{\sigma_{#1}} % added
\newcommand{\switchresidual}{\sigma_d}
\newcommand{\deformingfactor}[1]{\nu_{#1}} % added
\newcommand{\sharednormal}[1]{\mathbf{n}_{#1}}

\newcommand{\controlfullyactuated}{\mathbf{u}} % added
\newcommand{\controlunicycle}{\overline{\controlfullyactuated}} % added
\newcommand{\controlfullyactuatedmodel}{\mathbf{v}} % added
\newcommand{\controlunicyclemodel}{\overline{\controlfullyactuatedmodel}} % added
\newcommand{\linearinput}{v} % added
\newcommand{\angularinput}{\omega} % added
\newcommand{\linearinputmodel}{\hat{\linearinput}} % added
\newcommand{\angularinputmodel}{\hat{\angularinput}} % added
\newcommand{\angletransform}{\xi^{\hybridmode}} % added
\newcommand{\directionvector}{\mathbf{e}}
\newcommand{\projection}[2]{\mathrm{\Pi}_{#1}(#2)} % added
\newcommand{\localfreespace}[1]{\mathcal{LF}(#1)} % added
\newcommand{\localgoallinear}{\mathbf{y}_{d,\parallel}(\robotpositionunicyclemodel)} % added
\newcommand{\localgoalangular}{\mathbf{y}_{d,G}(\robotpositionunicyclemodel)} % added

\newcommand{\hybridsystem}{\mathsf{H}} % added
\newcommand{\hybridsystemunicycle}{\overline{\hybridsystem}} % added
\newcommand{\hybridmode}{\mathcal{I}} % added

\newcommand{\hybridfreespacemode}{\mathcal{F}^\hybridmode} % added
\newcommand{\hybridfreespacemodeprime}{\mathcal{F}^{\hybridmode'}} % added
\newcommand{\hybridfreespacemodesemantic}{\mathcal{F}_{sem}^\hybridmode} % added
\newcommand{\hybridfreespacemodemapped}{\mathcal{F}_{map}^\hybridmode} % added
\newcommand{\hybridfreespacemodemodel}{\mathcal{F}_{model}^\hybridmode} % added
\newcommand{\hybridfreespacemodemodelprime}{\mathcal{F}_{model}^{\hybridmode'}} % added
\newcommand{\hybriddomain}{\mathsf{D}} % added
\newcommand{\hybridgraph}{\mathsf{\Gamma}} % added
\newcommand{\hybridguardrestriction}{G} % added
\newcommand{\hybridguard}{\mathsf{\hybridguardrestriction}} % added
\newcommand{\hybridresetrestriction}{R} % added
\newcommand{\hybridreset}{\mathsf{\hybridresetrestriction}} % added
\newcommand{\hybridfieldrestriction}{U} % added
\newcommand{\hybridfield}{\mathsf{\hybridfieldrestriction}} % added
\newcommand{\hybriddomainunicycle}{\overline{\hybriddomain}}
\newcommand{\hybridguardrestrictionunicycle}{\overline{\hybridguardrestriction}} % added
\newcommand{\hybridguardunicycle}{\overline{\hybridguard}} % added
\newcommand{\hybridresetrestrictionunicycle}{\overline{\hybridresetrestriction}} % added
\newcommand{\hybridresetunicycle}{\overline{\hybridreset}} % added
\newcommand{\hybridfieldrestrictionunicycle}{\overline{\hybridfieldrestriction}} % added
\newcommand{\hybridfieldunicycle}{\overline{\hybridfield}} % added

\newcommand{\ball}[2]{\mathsf{B}(#1,#2)}
\newcommand{\ballclosure}[2]{\overline{\mathsf{B}(#1,#2)}}

\begin{document}

\title{Reactive Navigation in Partially Familiar Planar Environments Using Semantic Perceptual Feedback}
\titlerunning{Vasilopoulos, et al.}
\authorrunning{Vasilopoulos, et al.}
\author{Vasileios Vasilopoulos\inst{1}, Georgios Pavlakos\inst{2}, Karl Schmeckpeper\inst{3}, Kostas Daniilidis\inst{3} and Daniel E. Koditschek\inst{1}% <-this % stops a space
\institute{Department of Electrical and Systems Engineering, University of Pennsylvania, \\ Philadelphia, PA 19104,
        {\tt\small \{vvasilo,kod\}@seas.upenn.edu} \and
        Department of Electrical Engineering and Computer Sciences, UC Berkeley, \\ Berkeley, CA 94720,
        {\tt\small pavlakos@berkeley.edu} \and
        Department of Computer and Information Science, University of Pennsylvania, \\ Philadelphia, PA 19104,
        {\tt\small karls@seas.upenn.edu, kostas@cis.upenn.edu}}%
}%

\maketitle

\begin{abstract}
This paper solves the planar navigation problem by recourse to an online reactive scheme that exploits recent advances in SLAM and visual object recognition to recast prior geometric knowledge in terms of an offline catalogue of familiar objects. The resulting vector field planner guarantees convergence to an arbitrarily specified goal, avoiding collisions along the way with fixed but arbitrarily placed instances from the catalogue as well as completely unknown fixed obstacles so long as they are strongly convex and well separated. We illustrate the generic robustness properties of such deterministic reactive planners as well as the relatively modest computational cost of this algorithm by supplementing an extensive numerical study with physical implementation on both a wheeled and legged platform in different settings.
\end{abstract}


\allowdisplaybreaks

\section{Introduction}

\begin{figure}[t]
\centering
\includegraphics[width=1.0\textwidth]{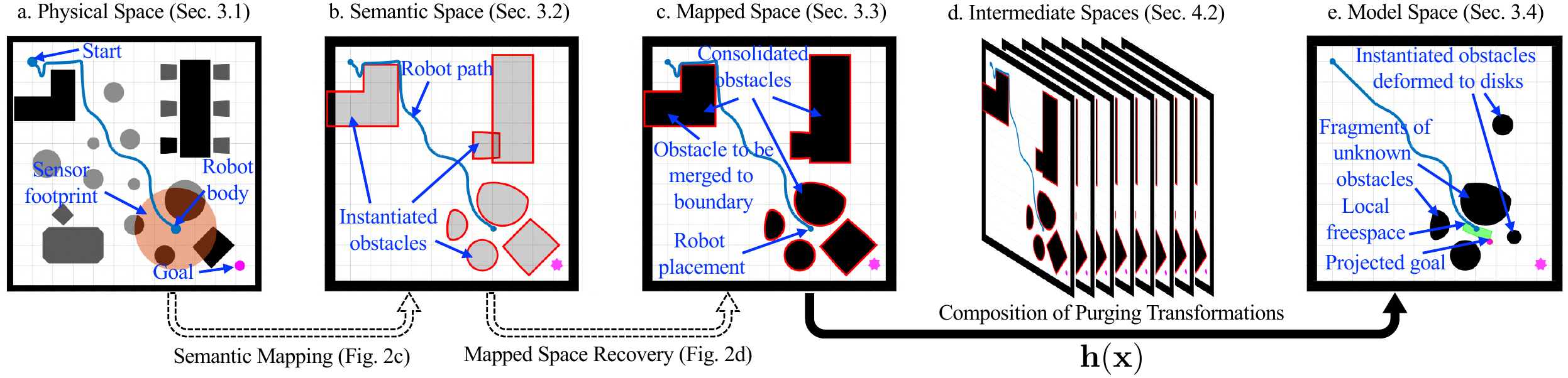}
\caption{Snapshot Illustration of Key Realtime Computation and Associated Models: The robot moves in the physical space (a - Section \ref{subsec:physical_space}), depicted as the blue trace of its centroid, toward a goal (pink) discovering along the way (black) both familiar objects of known geometry but unknown location (dark grey) and unknown obstacles (light grey), with an onboard sensor of limited range (orange disk). These obstacles are localized, dilated and stored permanently in the semantic space (b - Section \ref{subsec:semantic_space}) if they have familiar geometry, or temporarily, with just the corresponding sensed fragments, if they are unknown. The consolidated obstacles (resolved in real time from the unions of overlapping localized familiar obstacles), along with the sensed fragments of the unknown obstacles, are then stored in the mapped space (c - Section \ref{subsec:mapped_space}). A nonlinear change of coordinates, $\diffeogeneric(\robotposition)$, into a topologically equivalent but geometrically simplified model space (e - Section \ref{subsec:model_space}, depicting the robot's placement and prior trajectory amongst the $\diffeogeneric$-deformed convex images of the mapped obstacles) is computed instantaneously each time a new perceptual event instantiates more obstacles to be localized in the semantic space, thus redefining the mapped space. The map, $\diffeogeneric$, is a diffeomorphism, computed via composition of ``purging'' transformations between intermediate spaces (d - Section \ref{subsec:leaf_purging}) that abstract the consolidated localized polygonal obstacles by successively pruning away their geometric details to yield topologically equivalent disks. A doubly reactive control scheme for convex environments \cite{arslan_kod_WAFR2016} defines a vector field on the model space which is transformed in realtime through the diffeomorphism to generate the input in the physical space (Section \ref{sec:controller}).} \label{fig:diffeo_idea}
\end{figure}

\begin{figure}
\centering
\includegraphics[width=0.9\textwidth]{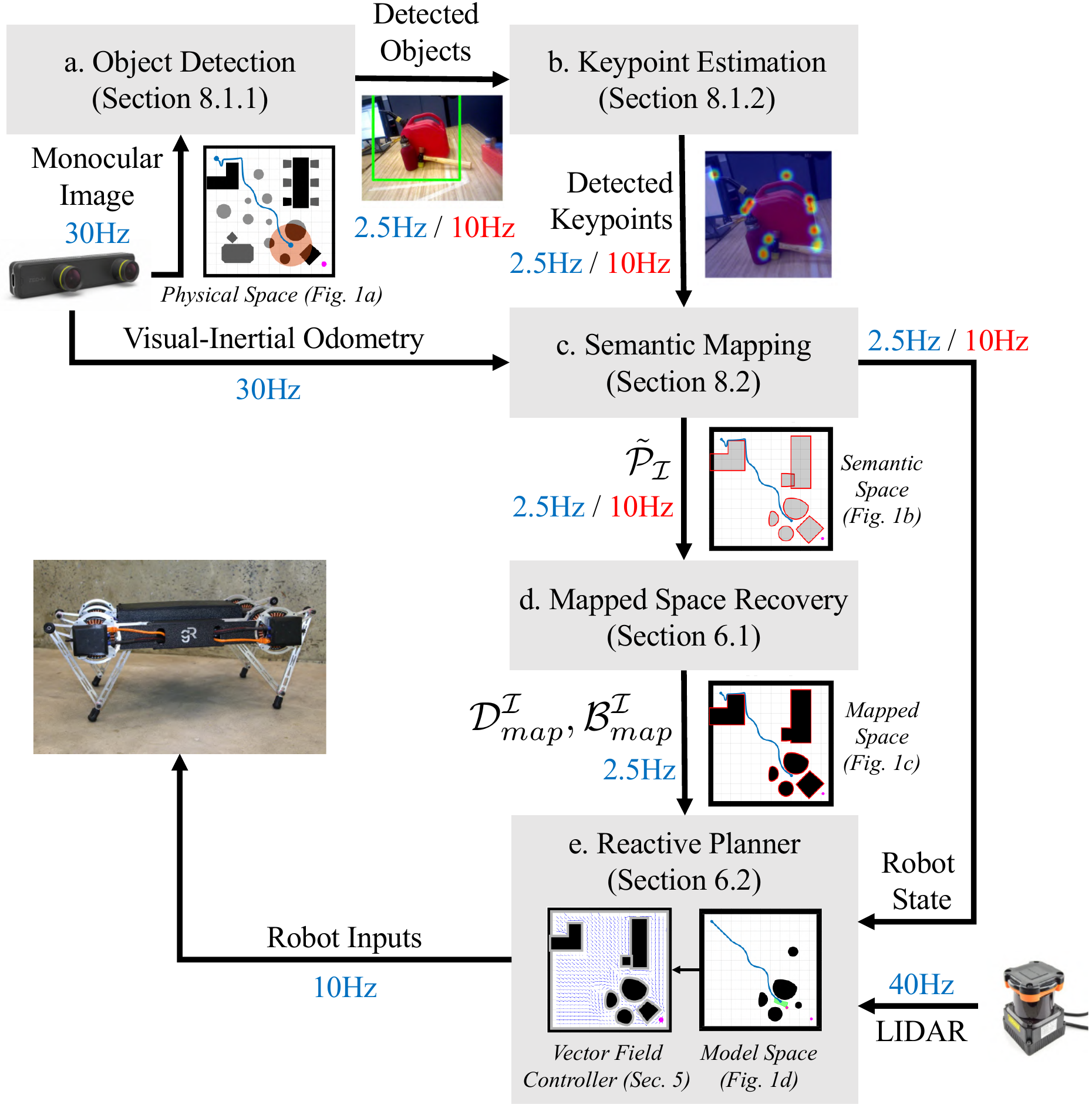}
\caption{A summary of our online reactive planning architecture. Using the camera image, two separate neural network architectures (configured in serial and run either onboard at 2.5Hz, or offboard at 10Hz) (a) detect familiar obstacles \cite{yolov3} (Section \ref{subsubsec:object_detection}) and (b) localize corresponding semantic keypoints \cite{Pavlakos2017} (Section \ref{subsubsec:keypoint_localization}). (c) The keypoint locations on the image and an egomotion estimate provided by visual inertial odometry are used by the semantic mapping module \cite{Bowman2017} (Section \ref{subsec:semantic_mapping}) to provide updated robot ($\robotposition$) and obstacle poses ($\knownobstaclesetphysical$) on the plane. (d) The mapped space tracking algorithm (Section \ref{subsec:mapped_space_recovery} - Algorithm \ref{algorithm:mapped_space_recovery}), run onboard at 2.5Hz, uses $\knownobstaclesetphysical$ to generate the list of obstacles in the mapped space $\knownobstaclesetdilatedmappeddisk, \knownobstaclesetdilatedmappedintrusion$. (e) The reactive planning module (Section \ref{subsec:reactive_planning} - Algorithm \ref{algorithm:reactive_planning}), run onboard at 10Hz, uses $\knownobstaclesetdilatedmappeddisk, \knownobstaclesetdilatedmappedintrusion$, along with LIDAR data for unknown obstacles, to provide the robot inputs and close the control loop.} \label{fig:algorithm}
\end{figure}

This paper advances the formally demonstrable capabilities of online reactive motion planners by disentangling the topology of navigation from the geometry of perception. Specifically, we appeal to a semantically aware perceptual oracle for the instantaneous recognition and localization of previously memorized objects and abstract away their geometric details in real time as they are encountered, yielding a sequence of topologically representative motion planning problems solvable by recourse to purely reactive online methods. Recent advances in SLAM and visual object recognition afford a working empirical realization of this provably correct scheme in both quasi-static and highly dynamic physical planar robots.

\subsection{Motivation and Prior Work}
Even as legged \cite{wooden_etal_ICRA2010,johnson_hale_haynes_kod_SSRR2011,Ilhan_Johnson_Koditschek_2018} and aerial \cite{amazon_prime,tang_kumar_2018,mohta_2018,gao_2019} robots engage increasingly realistic, unstructured environments, intuition suggests that prior experience ought to yield deterministic navigation guarantees, postponing statistical predictions of performance to estimated \cite{trautman_ma_murray_krause_IJRR2015}, learned \cite{henry_vollmer_ferris_fox_ICRA2010} or simulated \cite{karaman_frazzoli_ICRA2012} characterizations of truly bewilderingly dense or moving environments. Similarly, sampling-based methods, motivated by the typically high dimensional configuration spaces arising from combined task and motion planning \cite{garrett_ijrr_2018}, can achieve asymptotic optimality \cite{vegabrown_2018}, but no guarantee of convergence (or task completion) under partial prior knowledge or limited sampling, and their probabilistic completeness guarantees can be slow to be realized in practice when confronting settings with narrow passages \cite{noreen-khan-habib-2016}, even in 2D environments. More importantly, our recent parallel work \cite{vasilopoulos_pavlakos_bowman_caporale_daniilidis_pappas_koditschek_2020}, that uses the reactive planning principles presented in this paper, shows that existing state-of-the-art path replanning algorithms for unknown 2D environments \cite{otte-2015} can cycle repeatedly in the presence of both unforeseen obstacles and narrow passages as they search for alternative openings, before eventually (and after protracted cycling) reporting failure (incorrectly) and halting.

\subsubsection{Reactive Navigation}

Heretofore, deterministically safe, convergent reactive methods have required substantial prior knowledge of a static environment, whether encoded using navigation functions \cite{rimon1992,filippidis_kyriakopoulos_2012,Lionis_Papageorgiou_Kyriakopoulos_2008}, harmonic potential functions \cite{Vlantis_Vrohidis_Bechlioulis_Kyriakopoulos_2018,conner_2011} or pre-computed sequences of ``funnels'' \cite{Majumdar_Tedrake_2017}. In contrast, sensor-driven planners in this general tradition \cite{paranjape_etal_IJRR2015,johnson_hale_haynes_kod_SSRR2011,simmons_ICRA1996,fiorini_shiller_IJRR1998,vandenberg_guy_lin_manocha_ISRR2011,vandenberg_lin_manocha_ICRA2008,brock_khatib_ICRA1999,borenstein_koren_TRA1991,borenstein_koren_TSMC1989,khatib_1986} have guaranteed collision avoidance but have offered no assurance of convergence to a designated goal.

Recent advances in the theory of sensor-based navigation \cite{arslan_kod_WAFR2016,arslan_kod_ICRA2016B,arslan_kod_ICRA2017} relying on the properties of metric projections on convex sets \cite{Kuntz-1994} (and other parallel approaches \cite{Ilhan_Johnson_Koditschek_2018,Paternain_Koditschek_Ribeiro_2017,Ataka_Lam_Althoefer_2018,Chang_Marsden_2014}) add the key feature of guaranteed convergence to a designated goal, by trading away prior knowledge for the presumption of simplicity: unknown obstacles can be successfully negotiated in real time without losing global convergence guarantees if they are ``round'' (i.e., very strongly convex in a sense made precise in \cite{Arslan_Koditschek_2018}). 

However, this presumption, along with the additional requirement for enough separation between the obstacles in the workspace, limit the domain of application for such methods to geometrically simple environments and might prohibit successful navigation in complicated, unstructured environments with non-convex geometry. Hence, other reactive approaches either seek to appropriately modify the input reference signal to account for unanticipated (potentially non-convex) disturbances \cite{Revzen_Ilhan_Koditschek_2012}, or rely on stochastic frameworks that are empirically shown to improve performance with non-convex obstacles \cite{Reverdy_Ilhan_Koditschek_2015}, with no guarantees of convergence.

\subsubsection{Realtime Perception}

In this work, we address these shortcomings by appeal to an agent's memory evoked by execution-time perceptual cues. Recent advances in semantic SLAM \cite{Atanasov_2016,Bowman2017} and object pose extraction using convolutional neural net architectures \cite{Kar_Tulsiani_Carreira_Malik_2015,Kong_Lin_Lucey_2017,Pavlakos2017} now provide an avenue for systematically composing partial prior knowledge about the robot's workspace within a deterministic framework well suited to the vector field planning methods reviewed above.

Contrasting recent work has recruited end-to-end learning to achieve obstacle avoiding reactions within semantically labeled representations of familiar environments \cite{Gupta_2017}, or supplemented such deep-learned representations with reference paths \cite{Kumar_2018}, or optimally generated waypoint sequences \cite{bansal_2019} that guide the robot to its destination. Although such approaches cannot guarantee safe convergence to the robot's destination, they promote the importance of landmark-based navigation, already highlighted by parallel work in biology \cite{cowan_navigation}. However, characteristically, the input to such architectures is raw visual data thereby generating egocentric reactions that are hostage to the experience of one particular environment. In contrast, our compositional use of semantically tagged, learned-object recognizers affords systematic re-use across many different environments and achieves formal deterministic guarantees as well --- at least up to their (admittedly still far from formally justifiable) idealization as perfect realtime perceptual oracles.

\subsubsection{Topologically Informed Navigation}

Work on the topology of motion planning \cite{farber_topology,Farber_Grant_Lupton_Oprea_2019} has overtaken earlier investigation of reactive (i.e., vector field) navigation planners \cite{Rimon_1990,Rimon_Koditschek_1989} to the point that, comparatively, only preliminary results on their intrinsic limitations have been reported \cite{Baryshnikov_Shapiro_2014}. It seems clear that our success in achieving such strong results for a broad class of partially known environments is due to the simplicity of the problem class (punctured two dimensional manifolds have the homotopy type of a bouquet of circles), but we are not in a position to opine firmly on the likely limitations of this approach in higher dimensional settings.

Recently, several contributions have focused on either finding invariants for homology classes to facilitate optimal path search in known environments \cite{Bhattacharya_Ghrist_Kumar_2015}, exploiting data to enforce topological constraints \cite{Pokorny_Kragic_2015}, or conceptualizing sensor measurements related to the shape of an object in a topologically meaningful way using persistent homology \cite{Mueller_Birk_2018}. In contrast, we extract geometric and topological information about the robot's workspace at execution time in order to construct a map between a geometrically complicated mapped space and a (topologically equivalent but geometrically simple) model space that can be used for planning purposes. To this end, we employ methods from the field of computational geometry for implicit description of geometric shape using R-functions \cite{shapiro2007}, convex decomposition \cite{keil-convex-decomposition} and logic operations with polygons \cite{egenhofer_1991,clementini_1993,douglas_1973}.

\subsection{Summary of Contributions}
We consider the navigation problem in a 2D workspace cluttered with unknown convex obstacles, along with ``familiar'' non-convex obstacles that belong to classes of known geometries, but whose number and placement are \`a-priori unknown. We assume a limited-range onboard sensor and a catalogue of known obstacles, along with a ``mapping oracle'' for their online identification and localization in the physical workspace. This framework allows the robot to explore the geometry and topology of its workspace in real time as it navigates toward its goal, by recognizing and incorporating in its stored semantic map ``familiar'' obstacles, whose number and placement are otherwise unknown, awaiting discovery at execution time.

Based on the aforementioned description, we propose a representation of the environment taking the form of a ``multi-layer'' collection of topological spaces whose realtime interaction can be exploited to integrate the geometrically naive sensor driven methods of \cite{arslan_kod_WAFR2016} with the offline geometrically sensitive methods of \cite{rimon1992}. 

Specifically, we adapt the construction of \cite{Rimon_Koditschek_1989} to generate a realtime smooth change of coordinates (a {\it diffeomorphism}) of the mapped space of the environment into a (locally) topologically equivalent but geometrically more favorable model space, relative to which the sensor-based reactive methods of \cite{arslan_kod_WAFR2016} can be directly applied. We prove that the conjugate vector field defined by appropriately transforming the reactive model space back through this diffeomorphism induces a vector field on the robot's physical configuration space that inherits the same formal guarantees of obstacle avoidance and convergence. Since the robot's knowledge about the geometry and topology of its workspace at execution time is constantly updated, we adopt a hybrid dynamical systems description of our navigation framework, and show that the resulting hybrid system both inherits the consistency properties outlined in \cite{Johnson_Burden_Koditschek_2016} and safely drives the robot to the goal without violating given command limits. 

We extend the construction to the case of a differential drive robot, by pulling back the extended field over planar rigid transformations introduced for this purpose in \cite{arslan_kod_WAFR2016} through a suitable polar coordinate transformation of the tangent lift of our original planar diffeomorphism and demonstrate, once again, that the physical differential drive robot inherits the same obstacle avoidance and convergence properties as those guaranteed for the geometrically simple model robot \cite{arslan_kod_WAFR2016}.

We believe that this is the first {\it doubly-reactive controller} \cite{arslan_kod_WAFR2016} (i.e., a navigation framework wherein not only the robot's trajectory but also the control vector field that generates it are computed online at execution time) that can handle arbitrary polygonal shapes in real time without the need for specific separation assumptions between the familiar obstacles, by combining perception and object recognition for the familiar obstacles with local range measurements (e.g., LIDAR) for the unknown obstacles, to yield provably correct navigation in geometrically complicated environments. Furthermore, unlike RRT-based \cite{lavalle_kuffner_2001} or PRM-based \cite{kavraki_1996} algorithms, and similarly to other vector-field based approaches, our framework is capable of solving the overall ``kinodynamic'' problem online, instead of executing separate trajectory and motion planning, for both a fully actuated particle and a differential drive robot.

Finally, by coupling the semantic SLAM framework of \cite{Bowman2017} and the object detection pipeline of \cite{Pavlakos2017} with our reactive planning architecture, we are able to localize against isolated semantic cues while navigating, instead of localizing against entire scenes \cite{Gupta_2017} or visual geometric features \cite{hesch_2014}. Therefore, by training just on data from the objects the robot is expected to encounter, we introduce modularity and robustness in our approach, while simultaneously performing online planning that does not rely on specific features of a deep network architecture (e.g., number or type of layers), \cite{Gupta_2017,Kumar_2018,bansal_2019}.

\subsection{Organization of the Paper}
\label{subsec:organization}
The paper is organized as follows. Section \ref{sec:problem_formulation} describes the problem and establishes our assumptions. Section \ref{sec:environment_representation} describes the physical, semantic, mapped and model planning spaces (summarized in Fig. \ref{fig:diffeo_idea}) used in the diffeomorphism construction between the mapped and model spaces, whose properties are established next in Section \ref{sec:diffeomorphism}. Section \ref{sec:controller} provides the formal hybrid systems description framework and the correctness proofs for both a fully actuated (Theorem \ref{theorem:hybrid_fullyactuated}) and differential drive (Theorem \ref{theorem:hybrid_unicycle}) velocity controlled planar robot, comprising the central theoretical contribution of this paper. 

Based on these results, Section \ref{sec:online_algorithms} continues with a description of the implemented mapped space recovery and reactive planning algorithms, for both a fully actuated and a differential drive robot, shown in Fig. \ref{fig:algorithm}-(d),(e). Section \ref{sec:numerical_results} presents a variety of illustrative numerical studies, and Section \ref{sec:experimental_setup} continues with a brief description of the experimental setup, realizing the deployed perception (relying on prior work and shown in Fig. \ref{fig:algorithm}-(a),(c)) and motion planning (Fig. \ref{fig:algorithm}-(d),(e)) algorithms on both the Turtlebot \cite{turtlebot} and the Minitaur \cite{ghostminitaur} robot. Section \ref{sec:experiments} continues with our experimental results, and Section \ref{sec:conclusion} concludes by summarizing our findings and sketching some of the future work now in progress building on these results. Finally, the Appendix provides details of computational methods and proofs.
\section{Problem Formulation}
\label{sec:problem_formulation}

\begin{table}
    \centering
    \begin{tabular}{l l}
        \toprule
        \toprule
        $\workspace \subset  \mathbb{R}^2$ & Closed, compact, polygonal, potentially non-convex workspace \\
        $\enclosingworkspace \subset  \mathbb{R}^2$ & Enclosing workspace \eqref{eq:enclosing_workspace} \\
        $\freespace \subset \workspace$ & Freespace \eqref{eq:free_space} \\
        $\enclosingfreespace \subset \enclosingworkspace$ & Enclosing freespace \eqref{eq:enclosing_freespace} \\
        $\robotradius \in \mathbb{R}$ & Robot radius \\
        $\sensorrange \in \mathbb{R}$ & Sensor range \\
        $\goalposition \in \freespace$ & Goal location \\
        $\obstacleset := \{ \obstacle_1, \obstacle_2, \ldots \} \subseteq \mathbb{R}^2$ & Set of fixed, disjoint obstacles \\
        $\knownobstacleset := \{\knownobstacle_i\}_{i \in \knownobstaclesetindex} \subseteq \obstacleset$ & Set of ``familiar'', polygonal obstacles, \\
        & indexed by the set $\knownobstaclesetindex := \{1,\ldots,\knownobstaclecardinality\} \subset \mathbb{N}$ \\
        $\unknownobstacleset := \obstacleset \backslash \knownobstacleset = \{ \unknownobstacle_i \}_{i \in \unknownobstaclesetindex}$ & Set of completely unknown obstacles, \\
        & indexed by the set $\unknownobstaclesetindex := \{1,\ldots,\unknownobstaclecardinality\} \subset \mathbb{N}$ \\
        $\obstaclesetdilated, \knownobstaclesetdilated, \unknownobstaclesetdilated$ & Set of obstacles in $\obstacleset, \knownobstacleset, \unknownobstacleset$ respectively, \\
        & dilated by the robot radius, $\robotradius$ \\
        \hline
    \end{tabular}
    \caption{Key symbols used throughout this paper, associated with the Problem Formulation in Section \ref{sec:problem_formulation}. See also Table \ref{table:notation_environment} for notation associated with the environment representation in Section \ref{sec:environment_representation}, Table \ref{table:notation_diffeo} for notation associated with the diffeomorphism construction in Section \ref{sec:diffeomorphism}, and Table \ref{table:notation_controller} for notation associated with our reactive controller in Section \ref{sec:controller}.}
    \label{table:notation}
\end{table}

Similarly to \cite{vasilopoulos_koditschek_WAFR2018}, we consider a disk-shaped robot with radius $\robotradius>0$, centered at $\robotposition \in \mathbb{R}^2$, navigating a closed, compact, polygonal, potentially non-convex workspace $\workspace \subset \mathbb{R}^2$, with known outer boundary $\partial \workspace$, towards a target location $\goalposition \in \workspace$. The robot is assumed to possess a sensor with fixed range $\sensorrange$, capable of recognizing ``familiar'' objects, as well as estimating the distance of the robot to nearby obstacles\footnote{For our hardware implementation, this idealized sensor is reduced to a combination of a LIDAR for distance measurements to obstacles and a monocular camera for object recognition and pose identification.}. We also define the {\it enclosing workspace}, as the convex hull of the closure of the workspace $\workspace$:
\begin{equation}
    \enclosingworkspace := \left\{ \robotposition \in \mathbb{R}^2 \, | \, \robotposition \in \text{Conv}(\overline{\workspace}) \right\} \label{eq:enclosing_workspace}
\end{equation}

The workspace is cluttered by a finite, unknown number of fixed, disjoint obstacles, denoted by $\obstacleset:=\{\obstacle_1,\obstacle_2,\ldots\}$. By convention, the set $\obstacleset$ also includes potentially non-convex ``intrusions'' of the boundary of the physical workspace $\workspace$ into the enclosing workspace $\enclosingworkspace$, that can be described as the connected components of $\enclosingworkspace \backslash \workspace$. We use the notation in \cite{arslan_kod_WAFR2016} and define the {\em freespace} as
\begin{equation}
\freespace := \left\{ \robotposition \in \enclosingworkspace \, \Big| \, \ballclosure{\robotposition}{\robotradius} \subseteq \enclosingworkspace \, \backslash \, \bigcup_i \obstacle_i \right\} \label{eq:free_space}
\end{equation}
where $\ball{\robotposition}{\robotradius}$ is the open ball centered at $\robotposition$ with radius $\robotradius$, and $\ballclosure{\robotposition}{\robotradius}$ denotes its closure. Similarly to the enclosing workspace, $\enclosingworkspace$, we define the {\it enclosing freespace}, $\enclosingfreespace$ as
\begin{equation}
    \enclosingfreespace := \left\{ \robotposition \in \mathbb{R}^2 \, | \, \robotposition \in \text{Conv}(\overline{\freespace}) \right\} \label{eq:enclosing_freespace}
\end{equation}

Although none of the positions of any obstacles in $\obstacleset$ are \`{a}-priori known, a subset $\knownobstacleset:=\{\knownobstacle_i\}_{i \in \knownobstaclesetindex} \subseteq \obstacleset$ of these obstacles, indexed by $\knownobstaclesetindex:=\{1,\ldots,\knownobstaclecardinality\} \subset \mathbb{N}$, is assumed to be ``familiar'' in the sense of having an \`{a}-priori known, readily recognizable, potentially non-convex, polygonal geometry (i.e., belonging to a known catalogue of {\it geometry classes}), which the robot can identify and localize instantaneously from online sensory measurement, as described in Section \ref{sec:experimental_setup}. We require that this subset also includes all connected components of $\enclosingworkspace \backslash \workspace$. The remaining obstacles in $\unknownobstacleset:=\obstacleset\backslash\knownobstacleset$, indexed by $\unknownobstaclesetindex:=\{1,\ldots,\unknownobstaclecardinality\} \subset \mathbb{N}$ are assumed to be strictly convex but are in all other regards (location and specific shape) completely unknown to the robot, while nevertheless satisfying a curvature condition given in \cite[Assumption 2]{arslan_kod_WAFR2016}. This condition is interpreted in \cite{arslan_kod_WAFR2016} as the requirement for each convex obstacle $\unknownobstacle \in \unknownobstacleset$ to be sufficiently ``round'', since mere obstacle convexity with flat edges can still generate undesired local minima when using a greedy reactive controller.

To simplify our notation, we neglect the robot dimensions, by dilating each obstacle in $\obstacleset$ by $\robotradius$, and assume that the robot operates in $\mathcal{F}$. We denote the set of dilated obstacles derived from $\obstacleset, \knownobstacleset$ and $\unknownobstacleset$, by $\obstaclesetdilated, \knownobstaclesetdilated$ and $\unknownobstaclesetdilated$ respectively. Since obstacles in $\knownobstacleset$ are polygonal, and dilations of polygonal obstacles are not in general polygonal, we approximate obstacles in $\knownobstaclesetdilated$ with conservative polygonal supersets. Note that since the set $\knownobstacleset$ is required to contain all connected components of $\enclosingworkspace \backslash \workspace$, that describe non-convex ``intrusions'' of the boundary of the physical workspace $\workspace$ into the enclosing workspace $\enclosingworkspace$, the set $\knownobstaclesetdilated$ is similarly required to contain the dilations of these intrusions. For obstacles in $\unknownobstaclesetdilated$ we require the following separation assumptions, introduced in \cite{arslan_kod_WAFR2016}.
\begin{assumption} \label{assumption:convex}
Each obstacle $\unknownobstacledilated_i \in \unknownobstaclesetdilated$ has a positive clearance $d(\unknownobstacledilated_i,\unknownobstacledilated_j) > 0$ from any obstacle $\unknownobstacledilated_j \in \unknownobstaclesetdilated$, with $i \neq j$, and a positive clearance $d(\unknownobstacledilated_i, \partial \freespace) > 0$ from the boundary of the freespace $\freespace$.
\end{assumption}

Then, similarly to \cite{rimon1992}, we describe each polygonal obstacle $\knownobstacledilated_i \in \knownobstaclesetdilated \subseteq \obstaclesetdilated$ by an {\it obstacle function}, a real-valued map providing an implicit representation of the form
\begin{equation}
\knownobstacledilated_i = \{ \robotposition \in \mathbb{R}^2 \, | \, \implicitgeneric_i(\robotposition) \leq 0 \}
\end{equation}
that the robot can construct online from the catalogued geometry after it has localized $\knownobstacledilated_i$, as detailed in Appendix \ref{appendix:implicit}. We also require the following technical assumption.
\begin{assumption} \label{assumption:beta}
For each $\knownobstacledilated_i \in \knownobstaclesetdilated$, there exists $\betaclearance_i>0$ such that the set $S_{\implicitgeneric_i} := \{\robotposition \, | \, \implicitgeneric_i(\robotposition) \leq \betaclearance_i \}$ has a positive clearance $d(S_{\implicitgeneric_i},\unknownobstacledilated) > 0$ from any obstacle $\unknownobstacledilated \in \unknownobstaclesetdilated$.
\end{assumption}

Note that Assumptions \ref{assumption:convex} and \ref{assumption:beta} constrain the shape (convex) and placements (sufficiently separated) only of obstacles that have never previously been encountered. Familiar (polygonal, dilated by $\robotradius$) obstacles $\knownobstacledilated_i \in \knownobstaclesetdilated$, while fixed, can be placed completely arbitrarily with no further prior information: in particular, they can overlap unrestrictedly, with no jeopardy to our formal results, because we rely on the sensor oracle to recognize and locate them in real time. Obstacles in $\knownobstaclesetdilated$ are similarly allowed to overlap with the boundary of the enclosing freespace $\partial \enclosingfreespace$. To control the scope of the present paper, we simply assume that a path to the goal always exists, i.e., the robot operates in a non-adversarial environment.

\begin{assumption} \label{assumption:non_adversarial}
The freespace $\freespace$ is path-connected.
\end{assumption}

Finally, in Section \ref{subsec:control_modes}, we impose the technical Assumption \ref{assumption:saddles} precluding the possibility that any of the (topologically unavoidable) unstable saddle points of our control law coincide with a catalogued ``knot point'' of any familiar obstacle (a condition that we conjecture should be generic in the configuration space of obstacle placements).

Based on these assumptions and further positing first-order, fully-actuated robot dynamics $\dot{\robotposition} = \controlfullyactuated(\robotposition)$, the problem consists of finding a Lipschitz continuous controller $\controlfullyactuated:\freespace \rightarrow \mathbb{R}^2$, that leaves the freespace $\freespace$ positively invariant and asymptotically steers almost all configurations in $\freespace$ to the given goal $\goalposition \in \freespace$. We have also summarized key symbols used throughout this paper in Table \ref{table:notation}.
\section{Navigational Representation of the Environment}
\label{sec:environment_representation}

\begin{table}[htbp]
    \centering
    \begin{tabular}{l l}
        \toprule
        \toprule
        $\knownobstaclesetphysical := \{ \knownobstacle_i\}_{i \in \hybridmode} \subseteq \knownobstacleset$ & Set of (constantly updated) instantiated familiar polygonal \\
        & obstacles, Section \ref{subsec:physical_space} \\
        $\hybridmode \subseteq \knownobstaclesetindex$ & Index set of the $|\hybridmode|$ presently instantiated obstacles in $\knownobstaclesetphysical$, \\
        & Section \ref{subsec:physical_space} \\
        $\freespacesemantic$ & Semantic space corresponding to $\hybridmode \in 2^{\knownobstaclesetindex}$, Section \ref{subsec:semantic_space} \\
        $\freespacemapped$ & Mapped space corresponding to $\hybridmode \in 2^{\knownobstaclesetindex}$, Section \ref{subsec:mapped_space} \\
        $\freespacemodel$ & Model space corresponding to $\hybridmode \in 2^{\knownobstaclesetindex}$, Section \ref{subsec:model_space} \\
        $\knownobstaclesetdilatedsemantic := \bigsqcup_{i \in \hybridmode} \knownobstacledilated_i $ & Set of familiar, polygonal obstacles instantiated in the \\
        & semantic space, Section \ref{subsec:semantic_space} \\
        $\unknownobstaclesetdilatedsemantic := \{ \unknownobstacledilated_i \}_{i \in \unknownobstaclesetdilatedsemanticindex} \subseteq \unknownobstaclesetdilated$ & Set of unknown obstacles in the semantic space, \\
        & indexed by $\unknownobstaclesetdilatedsemanticindex \subseteq \unknownobstaclesetindex$, Section \ref{subsec:semantic_space} \\
        $\knownobstaclesetdilatedmapped := \bigcup_{i \in \hybridmode} \knownobstacledilated_i = \{ \knownobstacledilated_i\}_{i \in \knownobstaclesetdilatedmappedindex}$ & Set of consolidated familiar obstacles in the mapped space, \\
        & indexed by $\knownobstaclesetdilatedmappedindex$, Section \ref{subsec:mapped_space} \\
        $\unknownobstaclesetdilatedmapped :=\unknownobstaclesetdilatedsemantic$ & Set of unknown obstacles in the mapped space, \\
        & indexed by $\unknownobstaclesetdilatedsemanticindex \subseteq \unknownobstaclesetindex$, Section \ref{subsec:mapped_space} \\
        $\knownobstaclesetdilatedmappedintrusion := \{ \knownobstacledilatedmappedintrusion_i \}_{i \in \knownobstaclesetdilatedmappedintrusionindex }$ & Connected components of $\knownobstaclesetdilatedmapped$ to be merged into $\partial \enclosingfreespace$, \\
        & indexed by $\knownobstaclesetdilatedmappedintrusionindex$, Section \ref{subsec:mapped_space} \\
        $\knownobstaclesetdilatedmappeddisk := \{ \knownobstacledilatedmappeddisk_i \}_{i \in \knownobstaclesetdilatedmappeddiskindex }$ & Connected components of $\knownobstaclesetdilatedmapped$ to be deformed into disks, \\
        & indexed by $\knownobstaclesetdilatedmappeddiskindex$, Section \ref{subsec:mapped_space} \\
        $\diffeocenter{i} \in \mathbb{R}^2, i \in \knownobstaclesetdilatedmappeddiskindex$ & Centers of the $|\knownobstaclesetdilatedmappeddiskindex|$ disks in the model space, Section \ref{subsec:model_space} \\
        $\diffeoradius{i} \in \mathbb{R}, i \in \knownobstaclesetdilatedmappeddiskindex$ & Radii of the $|\knownobstaclesetdilatedmappeddiskindex|$ disks in the model space, Section \ref{subsec:model_space} \\
        \hline
    \end{tabular}
    \caption{Key symbols related to the environment representation in Section \ref{sec:environment_representation}.}
    \label{table:notation_environment}
\end{table}

In this Section, we introduce associated notation for the four distinct representations of the environment that we will refer to as {\it planning spaces} and use in the construction of our algorithm. Fig. \ref{fig:diffeo_idea} illustrates the role of these spaces and the transformations that relate them in constructing and analyzing a realtime generated vector field that guarantees safe passage to the goal. The new technical contribution is an adaptation  of the methods of \cite{Rimon_Koditschek_1989} to the realtime construction of a diffeomorphism, $\diffeogeneric$, where the requirement for fast, online performance demands an algorithm that is as simple as possible and with few tunable parameters. Hence, since the reactive controller in \cite{arslan_kod_WAFR2016} is designed to (provably) handle convex shapes, sensed  obstacles not recognized by the semantic SLAM process are simply assumed to be convex (implemented by designing $\diffeogeneric$ to resolve to the identity transformation in the  neighborhood of ``unfamiliar'' objects) and the control response defaults to that prior construction.

\subsection{Physical Space}
\label{subsec:physical_space}
The {\it physical space} is a complete description of the geometry of the unknown actual world and while inaccessible to the robot is used for purposes of analysis. It describes the enclosing workspace $\enclosingworkspace$, punctured with the obstacles $\obstacleset$. This gives rise to the freespace $\freespace$, given in \eqref{eq:free_space}, consisting of all placements of the robot's centroid that entail no intersections of its body with any interior obstacles or intrusions from the boundary. The robot navigates this space toward the goal, discovering and localizing new obstacles along the way. Those discovered obstacles which are not convex  are (by assumption) ``familiar'' and are then ``instantiated'' --- recalled, and registered from memory to populate the accumulating record of discovery in the semantic space --- as we next discuss. Those which are ``unfamiliar'' are presumed convex and registered as such in the companion spaces next to be presented.\footnote{Although we make no use in the present paper of the discovered unfamiliar objects beyond simply avoiding them, future work now in progress relaxes the convexity requirement to build up in memory an increasingly complete geometric description (treated in the same manner as in the ``familiar'' case)  from whatever subsequent encounters ensue along the way to the goal. \label{footnote:memory}} 

We denote by $\knownobstaclesetphysical := \{ \knownobstacle_i\}_{i \in \hybridmode} \subseteq \knownobstacleset$ the finite set of (constantly updated) physically ``instantiated'' familiar objects, indexed by $\hybridmode \subseteq \knownobstaclesetindex$, that drives the construction of the semantic, mapped and model spaces described next. As explained in Section \ref{subsec:control_hybrid}, such elements $\hybridmode$ of the power set $2^{\knownobstaclesetindex}$ also index the modes of our hybrid system.

\subsection{Semantic Space}
\label{subsec:semantic_space}
The {\it semantic space} $\freespacesemantic$ records the robot's evolving information about the environment aggregated from the raw sensor data about the observable portions of a subset of unrecognized (and therefore, presumed convex) obstacles from $\unknownobstaclesetdilated$, together with the polygonal boundaries of the $|\hybridmode|$ familiar obstacles, that are instantiated at the moment the sensory data triggers the identification and localization a familiar obstacle.
\begin{definition}
    \label{definition:instantiation}
    A familiar obstacle $\knownobstacle \in \knownobstacleset$ is considered to be ``instantiated'', if it has been sensed using the robot's sensor of fixed range $\sensorrange$, recognized, localized, and its dilation $\knownobstacledilated \in \knownobstaclesetdilated$ is permanently included in the semantic space.\footnote{This implies that there exists a time $t_{\knownobstacledilated} > 0$, such that $\ballclosure{\robotposition^{t_{\knownobstacledilated}}}{\sensorrange} \cap \knownobstacle \neq \varnothing$ and $\ballclosure{\robotposition^t}{\sensorrange} \cap \knownobstacle = \varnothing$, for all $t < t_{\knownobstacledilated}$, with $\robotposition^t$ denoting the robot position at time $t$.}
\end{definition}
We denote the {\it set of unrecognized obstacles in the semantic space} by $\unknownobstaclesetdilatedsemantic:= \{ \unknownobstacledilated_i \}_{i \in \unknownobstaclesetdilatedsemanticindex}$, indexed by $\unknownobstaclesetdilatedsemanticindex \subseteq \unknownobstaclesetindex$, and the {\it set of familiar obstacles in the semantic space} by $\knownobstaclesetdilatedsemantic := \bigsqcup_{i \in \hybridmode} \knownobstacledilated_i$.

It is important to note that this environment is constantly updated, both by discovering and storing new familiar obstacles in the semantic map and by discarding old information and storing new information regarding obstacles in $\unknownobstaclesetdilated$.$^{\ref{footnote:memory}}$ Here, the robot is treated as a point particle, since all obstacles are dilated by $\robotradius$ in the passage from the workspace to the freespace representation of valid placements.

\subsection{Mapped Space}
\label{subsec:mapped_space}
Although the semantic space contains all the relevant geometric information (identity and pose) about the obstacles the robot has encountered, it does not explicitly contain any topological information about the explored environment, as represented by the disjoint union operation in the definition of $\knownobstaclesetdilatedsemantic$. This is because Assumption \ref{assumption:beta} does not exclude overlaps between obstacles in $\knownobstaclesetdilated$. Their algorithmically effective consolidation in real time reduces the number while increasing the geometric complexity of the actual freespace obstacles the robot must negotiate along the way to the goal. To do so, we need therefore to take unions of overlapping obstacles in $\knownobstaclesetdilatedsemantic$, making up $\knownobstaclesetdilatedmapped := \bigcup_{i \in \hybridmode} \knownobstacledilated_i = \{ \knownobstacledilated_i\}_{i \in \knownobstaclesetdilatedmappedindex}$ (i.e., a new set of {\it consolidated familiar obstacles} indexed by $\knownobstaclesetdilatedmappedindex$ with $|\knownobstaclesetdilatedmappedindex| \leq |\hybridmode|$), as well as copies of the sensed fragments of unknown obstacles from $\unknownobstaclesetdilatedsemantic$ (i.e., $\unknownobstaclesetdilatedmapped := \unknownobstaclesetdilatedsemantic$) to form the {\it mapped space} $\freespacemapped$ as
\begin{equation}
    \freespacemapped := \enclosingfreespace \backslash (\knownobstaclesetdilatedmapped \cup \unknownobstaclesetdilatedmapped)
\end{equation}
Note that, by Assumption \ref{assumption:beta}, the convex obstacles are assumed to be far enough away from the familiar obstacles, such that no overlap occurs in the above union.

Next, we focus on the connected components of $\knownobstaclesetdilatedmapped$; since Assumption \ref{assumption:beta} allows overlaps between obstacles in $\knownobstaclesetdilated$ and the boundary of the enclosing freespace $\partial \enclosingfreespace$, for any connected component $\knownobstacledilated$ of $\knownobstaclesetdilatedmapped$ such that $\knownobstacledilated \cap \partial \enclosingfreespace \neq \varnothing$, we take $\knownobstacledilatedmappedintrusion:= \knownobstacledilated \cap \enclosingfreespace$ and include $\knownobstacledilatedmappedintrusion$ in a new set $\knownobstaclesetdilatedmappedintrusion$, indexed by $\knownobstaclesetdilatedmappedintrusionindex$. The rest of the connected components in $\knownobstaclesetdilatedmapped$, which do not intersect $\partial \enclosingfreespace$, are included in a separate set $\knownobstaclesetdilatedmappeddisk$, indexed by $\knownobstaclesetdilatedmappeddiskindex$. The idea here is that obstacles in $\knownobstaclesetdilatedmappedintrusion$ should be merged to the boundary of the enclosing freespace $\partial \enclosingfreespace$, and obstacles in $\knownobstaclesetdilatedmappeddisk$ should be deformed to disks, since $\freespacemapped$ and $\freespacemodel$ need to be diffeomorphic.

\subsection{Model Space}
\label{subsec:model_space}
The {\it model space} $\freespacemodel$ has the same boundary as $\enclosingfreespace$ (i.e. $\partial \freespacemodel:=\partial \enclosingfreespace$) and consists of copies of the sensed fragments of the $|\unknownobstaclesetdilatedsemanticindex|$ unrecognized visible convex obstacles in $\unknownobstaclesetdilatedmapped$, and a collection of $|\knownobstaclesetdilatedmappeddiskindex|$ Euclidean disks corresponding to the $|\knownobstaclesetdilatedmappeddiskindex|$ consolidated obstacles in $\knownobstaclesetdilatedmappeddisk$ that are deformed to disks. The centers $\{\diffeocenter{i}\}_{i \in \knownobstaclesetdilatedmappeddiskindex}$ and radii $\{\diffeoradius{i}\}_{i \in \knownobstaclesetdilatedmappeddiskindex}$ of the $|\knownobstaclesetdilatedmappeddiskindex|$ disks are chosen so that $\ballclosure{\diffeocenter{i}}{\diffeoradius{i}}$ is contained in the interior of $\knownobstacledilatedmappeddisk_i \in \knownobstaclesetdilatedmappeddisk$, as required in \cite{Rimon_Koditschek_1989}. The obstacles in $\knownobstaclesetdilatedmappedintrusion$ are merged into $\partial \enclosingfreespace$, to make $\freespacemapped$ and $\freespacemodel$ topologically equivalent, through a map $\diffeo:\freespacemapped \rightarrow \freespacemodel$. We can, therefore, write $\freespacemodel$ as
\begin{equation}
    \freespacemodel = \enclosingfreespace \Big \backslash \left( \bigcup_{i \in \knownobstaclesetdilatedmappeddiskindex}  \ballclosure{\diffeocenter{i}}{\diffeoradius{i}} \cup \unknownobstaclesetdilatedmapped \right)
\end{equation}

Note here that we need to distinguish between familiar but unanticipated obstacles in $\knownobstaclesetdilatedmappeddisk$ and $\knownobstaclesetdilatedmappedintrusion$, and completely unknown convex obstacles in $\unknownobstaclesetdilatedmapped$, because our algorithm handles them quite differently. From a practical point of view, the robot instantiates familiar obstacles using its full sensor suite equipped with a learned catalogue, while completely unknown obstacles that the robot can only partially observe are handled solely based on distance measurements (e.g., LIDAR). More importantly, from a formal point of view, this distinction is needed in the definition of our global diffeomorphism (to be presented next in Section \ref{sec:diffeomorphism}), since only the familiar obstacles affect its construction. Moreover, the previously unseen unfamiliar obstacles cannot be simply registered in the model space, but they must be formally mapped there through that diffeomorphism, which is constructed so that it defaults to the identity transform on their boundaries.

\subsection{Implicit Representation of Obstacles}
\label{subsec:implicit}
We note here that the construction of the map $\diffeo$ between the mapped space $\freespacemapped$ and the model space $\freespacemodel$, described next in Section \ref{sec:diffeomorphism} and shown in Fig. \ref{fig:diffeo_idea}, relies on the existence of smooth implicit functions $\implicitgeneric : \mathbb{R}^2 \rightarrow \mathbb{R}$ for polygons, constructed such that $\implicitgeneric(\robotposition) = 0$ implies that $\robotposition$ lies on the boundary of the polygon. Although the construction of such functions is a separate problem on its own, here we derive implicit representations using so-called ``R-functions'' from the constructive solid geometry literature \cite{shapiro2007}. In Appendix \ref{appendix:implicit}, we describe the method used for the construction of such implicit functions for polygonal obstacles that have the desired property of being analytic everywhere except for the polygon vertices. We have already successfully implemented such constructions in a similar setting for star-shaped obstacles in \cite{vasilopoulos_koditschek_WAFR2018}. 

\section{The Diffeomorphism Between the Mapped and Model Spaces}
\label{sec:diffeomorphism}

In this Section, we describe our method of constructing the diffeomorphism, $\diffeo$, between the mapped space $\freespacemapped$ and the model space $\freespacemodel$. We assume that the robot has already recognized, localized and stored the $|\knownobstaclesetdilatedmappedindex|$ consolidated familiar polygonal obstacles in $\knownobstaclesetdilatedmapped$ in its map, and has subsequently identified obstacles to be merged to the boundary of the enclosing freespace $\partial \enclosingfreespace$, stored in $\knownobstaclesetdilatedmappedintrusion$, and obstacles to be deformed to disks, stored in $\knownobstaclesetdilatedmappeddisk$.

\begin{table}[htbp]
    \centering
    \begin{tabular}{l l}
        \toprule
        \toprule
        $\triangletree{\knownobstacledilated_i}$ & Tree of triangles, constructed from the dual graph of the triangulation \\
        & of $\knownobstacledilated_i$, Section \ref{subsec:obstacle_representation} \\
        $\trianglevertices{\knownobstacledilated_i}$ & Set of vertices of $\triangletree{\knownobstacledilated_i}$ identified with triangles in the triangulation \\
        &of $\knownobstacledilated_i$, Section \ref{subsec:obstacle_representation} \\
        $ \triangleedges{\knownobstacledilated_i}$ & Set of edges of $\triangletree{\knownobstacledilated_i}$ encoding triangle adjacency in the triangulation \\
        & of $\knownobstacledilated_i$, Section \ref{subsec:obstacle_representation} \\
        $\freespacemappedpurging{j_i} \subset \mathbb{R}^2$ & Mapped space before the purging of leaf triangle $j_i \in \trianglevertices{\knownobstacledilated_i}$, \\
        & Section \ref{subsec:leaf_purging} \\
        $\freespacemappedpurging{p(j_i)} \subset \mathbb{R}^2$ & Mapped space after the purging of leaf triangle $j_i \in \trianglevertices{\knownobstacledilated_i}$, Section \ref{subsec:leaf_purging} \\
        $\freespacemappedhat \subset \mathbb{R}^2$ & Mapped space after the purging of all leaf triangles, Sections \ref{subsec:leaf_purging}, \ref{subsec:root_purging} \\
        $\diffeocenter{j_i} \in \mathbb{R}^2$ & Admissible center for the purging transformation of a leaf triangle \\
        & $j_i \in \trianglevertices{\knownobstacledilated_i}$, Definition \ref{definition:center} \\
        $\diffeocenter{i} \in \mathbb{R}^2$ & Admissible center for the transformation of a root triangle \\
        & $r_i \in \trianglevertices{\knownobstacledilated_i}$, Definitions \ref{definition:centers_root_disk}, \ref{definition:centers_root_boundary} \\
        $\robotposition_{1j_i}, \robotposition_{2j_i}, \robotposition_{3j_i}$ & Vertices of a leaf triangle $j_i \in \trianglevertices{\knownobstacledilated_i}$, Section \ref{subsec:leaf_purging} \\
        $\robotposition_{1r_i}, \robotposition_{2r_i}, \robotposition_{3r_i}$ & Vertices of a root triangle $r_i \in \trianglevertices{\knownobstacledilated_i}$, Section \ref{subsec:root_purging} \\
        $\innerpolygon{j_i} \subset \mathbb{R}^2$ & Quadrilateral $\robotposition_{3j_i}\robotposition_{1j_i}\diffeocenter{j_i}\robotposition_{2j_i}\robotposition_{3j_i}$ associated with a leaf triangle \\
        & $j_i \in \trianglevertices{\knownobstacledilated_i}$, Section \ref{subsec:leaf_purging} \\
        $\innerpolygon{r_i} \subset \mathbb{R}^2$ & Quadrilateral $\robotposition_{3r_i}\robotposition_{1r_i}\diffeocenter{i}\robotposition_{2r_i}\robotposition_{3r_i}$ associated with a root triangle \\
        & $r_i \in \trianglevertices{\knownobstacledilated_i}$, Section \ref{subsec:root_purging} \\
        $\outerpolygon{j_i} \subset \mathbb{R}^2$ & Admissible polygonal collar associated with a leaf triangle $j_i \in \trianglevertices{\knownobstacledilated_i}$, \\
        & Definition \ref{definition:collars} \\
        $\outerpolygon{r_i} \subset \mathbb{R}^2$ & Admissible polygonal collar associated with a root triangle $r_i \in \trianglevertices{\knownobstacledilated_i}$, \\
        & Definitions \ref{definition:collars_root_disk}, \ref{definition:collars_root_boundary} \\
        $\innerpolygonimplicit{j_i}, \innerpolygonimplicit{r_i}: \mathbb{R}^2 \rightarrow \mathbb{R}$ & Implicit function associated with $\innerpolygon{j_i}$ \eqref{eq:implicit_gamma_ji} or $\innerpolygon{r_i}$ \eqref{eq:implicit_gamma_ri} \\
        $\outerpolygonimplicit{j_i}, \outerpolygonimplicit{r_i}: \mathbb{R}^2 \rightarrow \mathbb{R}$ & Implicit function associated with $\outerpolygon{j_i}$ \eqref{eq:implicit_delta_ji} or $\outerpolygon{r_i}$ \eqref{eq:implicit_delta_ri} \\
        $\innerpolygonsigma{j_i}, \innerpolygonsigma{r_i}: \mathbb{R}^2 \rightarrow [0,1]$ & Auxiliary $C^\infty$ switch associated with $\innerpolygon{j_i}$ \eqref{eq:sigma_gamma_ji} or $\innerpolygon{r_i}$ \eqref{eq:sigma_gamma_ri} \\
        $\outerpolygonsigma{j_i}, \outerpolygonsigma{r_i}: \mathbb{R}^2 \rightarrow [0,1]$ & Auxiliary $C^\infty$ switch associated with $\outerpolygon{j_i}$ \eqref{eq:sigma_delta_ji} or $\outerpolygon{r_i}$ \eqref{eq:sigma_delta_ri} \\
        $\switch{j_i}, \switch{r_i}: \mathbb{R}^2 \rightarrow [0,1]$ & $C^\infty$ switch of the transformation of a leaf triangle $j_i$ \eqref{eq:sigma_ji} or a root \\
        & triangle $r_i$ \eqref{eq:sigma_ri} \\
        $\deformingfactor{j_i}, \deformingfactor{r_i}: \mathbb{R}^2 \rightarrow [0,1]$ & Deforming factor for a leaf triangle $j_i$ \eqref{eq:deforming_factor_purging} or a root \\
        & triangle $r_i$ \eqref{eq:deforming_factor_disk}, \eqref{eq:deforming_factor_root_boundary} \\
        $\diffeopurging{j_i}:\freespacemappedpurging{j_i} \rightarrow \freespacemappedpurging{p(j_i)}$ & Purging transformation mapping a leaf triangle $j_i$ to its \\
        & parent $p(j_i)$ \eqref{eq:map_purging} \\
        $\diffeocomposition: \freespacemapped \rightarrow \freespacemappedhat$ & Composition of purging transformations mapping $\freespacemapped$ to $\freespacemappedhat$, \\
        & Section \ref{subsubsec:leaf_purging_composition} \\
        $\diffeoroot: \freespacemappedhat \rightarrow \freespacemodel$ & Diffeomorphism between $\freespacemappedhat$ and $\freespacemodel$ \eqref{eq:map_root} \\
        $\diffeo : \freespacemapped \rightarrow \freespacemodel$ & Diffeomorphism between $\freespacemapped$ and $\freespacemodel$ \eqref{eq:diffeo_final} \\
        \hline
    \end{tabular}
    \caption{Key symbols related to the diffeomorphism construction from $\freespacemapped$ to $\freespacemodel$, described in Section \ref{sec:diffeomorphism}.}
    \label{table:notation_diffeo}
\end{table}

The idea is then to compose a sequence of ``purging'' diffeomorphisms, which coincide with the identity map except on a small ``collar'' around each component of $\knownobstaclesetdilatedmapped$, that produce successively less complicated isolated shapes. The final simplified shapes are then conveniently deformed into a disk if the corresponding obstacle belongs in $\knownobstaclesetdilatedmappeddisk$, or into the boundary of $\enclosingfreespace$ if the corresponding obstacle belongs in $\knownobstaclesetdilatedmappedintrusion$, in order to generate the model space $\freespacemodel$. Before describing these transformations, we first provide some background on the used obstacle representation methodology in Section \ref{subsec:obstacle_representation}, and provide associated notation in Table \ref{table:notation_diffeo}.

\subsection{Obstacle Representation}
\label{subsec:obstacle_representation}
In order to construct the map $\diffeo$ between the mapped space and the model space, we assume that the robot has access to the triangulation of each one of the obstacles stored in both $\knownobstaclesetdilatedmappeddisk$ and $\knownobstaclesetdilatedmappedintrusion$. This triangulation can be efficiently constructed online upon recognition of each obstacle, using the {\it Two Ears Theorem} \cite{meisters_1975}, and the associated {\it Ear Clipping Method} \cite{elgindy_1993}\footnote{The authors thank Prof. Elon Rimon for pointing out these results.}.

\begin{figure}[H]
\centering
\includegraphics[width=1.0\textwidth]{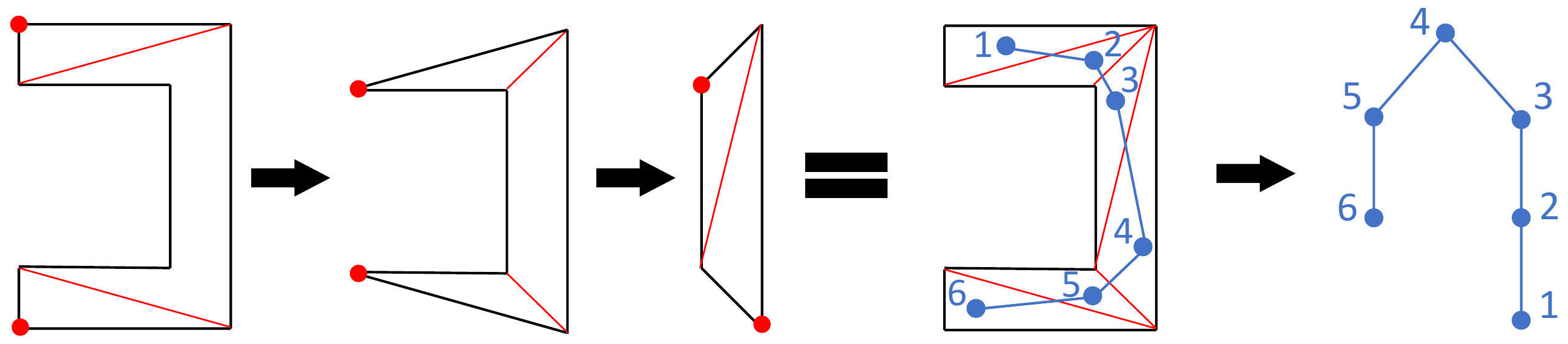}
\caption{Triangulation of a non-convex obstacle using the Ear Clipping Method. The original polygon is guaranteed to have at least two ears (red dots) by the Two Ears Theorem, which induce triangles that can be removed from the polygon. By repeating this process, we get the final triangulation and its dual graph, which is guaranteed to be a tree. This tree can be restructured by setting the root to be the triangle of maximal surface area, to yield the order of purging transformations in descending depth; in this particular example this order is $1\rightarrow2\rightarrow6\rightarrow3\rightarrow5\rightarrow4$.} \label{fig:triangulation_example}
\end{figure}

Briefly, an {\it ear} of a simple polygon is a vertex of the polygon such that the line segment between the two neighbors of the vertex lies entirely in the interior of the polygon. The Two Ears Theorem guarantees that every simple polygon has at least two such ears, and the Ear Clipping Method uses this result to efficiently construct polygon triangulations in $\mathcal{O}(n^2)$ time. Namely, an ear and its two neighbors form a triangle that is not crossed by any other part of the polygon and can be, therefore, safely removed. Removing a triangle of this type produces a polygon with one less vertex than the original polygon; we can repeat the process to eventually get a single triangle and complete the triangulation. An example is shown in Fig. \ref{fig:triangulation_example}.

Except for its utility in constructing triangulations, the Two Ears Theorem guarantees that the dual graph of the triangulation of a simple polygon with no holes constructed with the Ear Clipping Method (i.e., a graph with one vertex per triangle and one edge per pair of adjacent triangles) is in fact a tree \cite{orourke_1987}. 

Therefore, in order to construct a tree of triangles $\triangletree{\knownobstacledilated_i}:=(\trianglevertices{\knownobstacledilated_i},\triangleedges{\knownobstacledilated_i})$ corresponding to a polygon $\knownobstacledilated_i$, with $\trianglevertices{\knownobstacledilated_i}$ a set of vertices identified with triangles (i.e., vertices of the dual of the formal triangulation) and $\triangleedges{\knownobstacledilated_i}$ a set of edges encoding triangle adjacency, we can triangulate $\knownobstacledilated_i$ using the Ear Clipping Method, pick any triangle as root, and construct $\triangletree{\knownobstacledilated_i}$ based on the adjacency properties induced by the dual graph of the triangulation, as shown in Fig. \ref{fig:triangulation_example}. If $\knownobstacledilated_i \in \knownobstaclesetdilatedmappeddisk$, we pick as root the triangle with the largest surface area, whereas if $\knownobstacledilated_i \in \knownobstaclesetdilatedmappedintrusion$, we pick as root a triangle adjacent to $\partial \enclosingfreespace$. This will give us a {\it tree-of-triangles} for $\knownobstacledilated_i$ in a notion similar to \cite{rimon1992}. Our goal is then to successively ``purge'' this tree, triangle by triangle, in order of descending depth, until we reach the root triangle. Then, we can use a diffeomorphism similar to \cite{vasilopoulos_koditschek_WAFR2018} to map the exterior and boundary of the root triangle onto the exterior and boundary of a topologically equivalent disk if $\knownobstacledilated_i \in \knownobstaclesetdilatedmappeddisk$, or merge the root triangle into $\partial \enclosingfreespace$ if $\knownobstacledilated_i \in \knownobstaclesetdilatedmappedintrusion$. These operations are all performed online; we provide some computational performance metrics with our experiments in Section \ref{sec:experiments}. 

We describe the algorithm for each purging transformation of the leaf nodes in Section \ref{subsec:leaf_purging} and the (final) root triangle purging transformation in Section \ref{subsec:root_purging}. Finally, Section \ref{subsec:diffeo_description} defines the diffeomorphism between the mapped and model spaces, along with associated qualitative properties.

\subsection{Intermediate Spaces Related by Leaf Purging Transformations}
\label{subsec:leaf_purging}
In this Section, we describe the purging transformation that maps the boundary of a leaf triangle $j_i \in \trianglevertices{\knownobstacledilated_i}$ onto the boundary of its parent $p(j_i) \in \trianglevertices{\knownobstacledilated_i}$, as shown in Fig. \ref{fig:purging}-(1a), (2a). This gives rise to a composition of transformations between a succession of {\it intermediate spaces}, each including the triangle $j_i$, and $\freespacemappedpurging{p(j_i)}$, where $j_i$ has been mapped onto the boundary of its parent. Each of these transformations is in principle similar and performing a role analogous to the corresponding purging transformation in \cite{rimon1992}, with two important differences. First, it deforms space only ``locally'' around the triangle, without taking into consideration other triangles or polygons, affording better numerical stability since only one triangle is considered at a time. Second, the method presented in \cite{rimon1992} is limited to parent-child pairs that strongly overlap instead of just being adjacent, which makes the method impractical for the arbitrary polygonal shapes and meshes involved in this work. 

\subsubsection{Center of the Transformation and Surrounding Polygonal Collars}
\begin{figure}[H]
\centering
\includegraphics[width=0.8\textwidth]{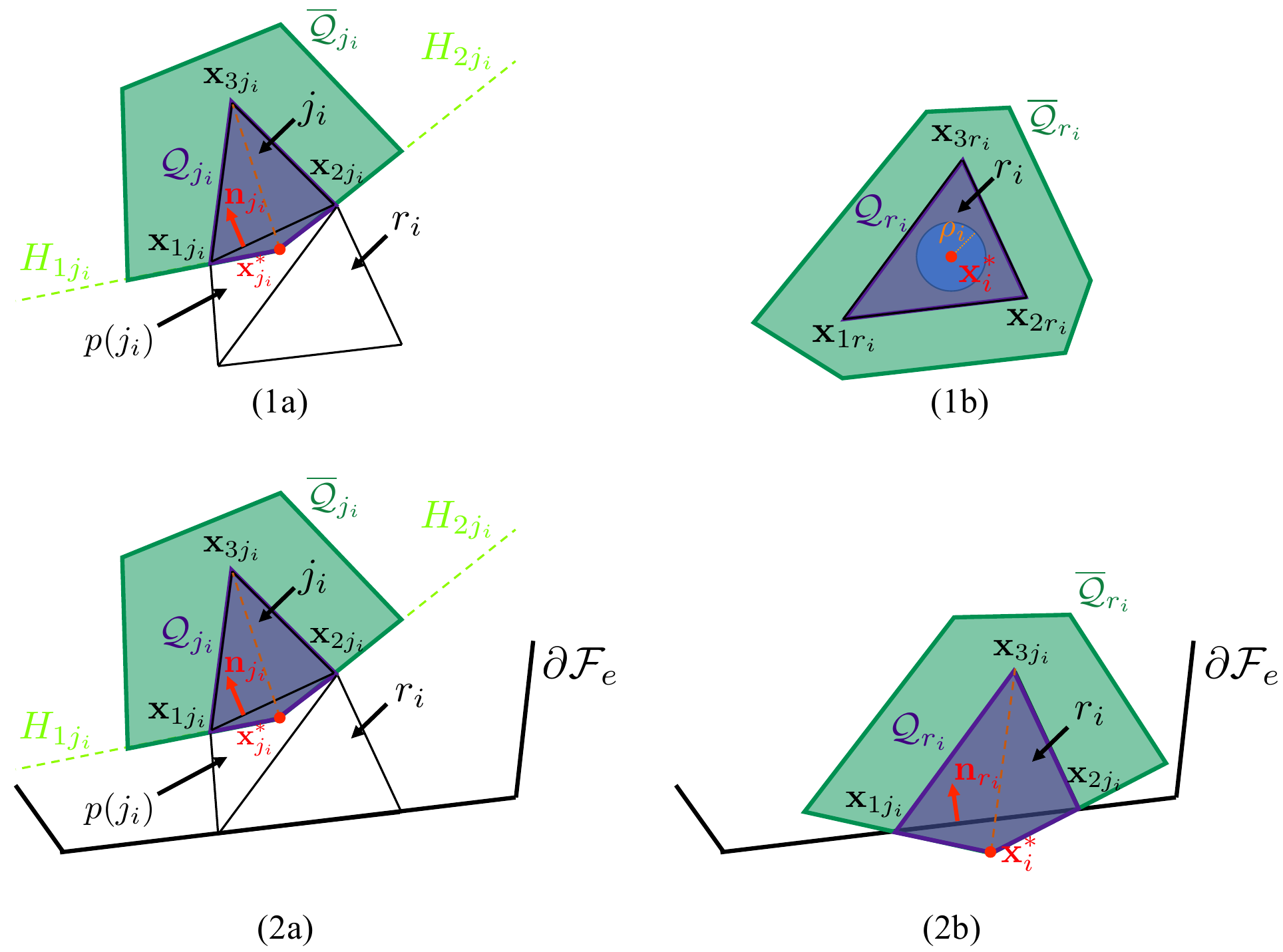}
\caption{Illustration of features used in the transformation of - Top: (1a) a leaf triangle $j_i$ onto its parent $p(j_i)$, and (1b) a root triangle $r_i$ onto a disk centered at $\diffeocenter{i}$ with radius $\diffeoradius{i}$ for an obstacle in $\knownobstaclesetdilatedmappeddisk$, Bottom: (2a) a leaf triangle $j_i$ onto its parent $p(j_i)$, and (2b) a root triangle $r_i$ onto $\partial \enclosingfreespace$ for an obstacle in $\knownobstaclesetdilatedmappedintrusion$.} \label{fig:purging}
\end{figure}

Let the vertices of the triangle $j_i \in \trianglevertices{\knownobstacledilated_i}$ be $\robotposition_{1j_i}$, $\robotposition_{2j_i}$ and $\robotposition_{3j_i}$ in counterclockwise order, with $\robotposition_{1j_i}\robotposition_{2j_i}$ the common edge between $j_i$ and $p(j_i)$. 
\begin{definition}
\label{definition:center}
An admissible center for the purging transformation of the leaf triangle $j_i \in \trianglevertices{\knownobstacledilated_i}$, denoted by $\diffeocenter{j_i}$, is a point along the triangle median from $\robotposition_{3j_i}$ and in the interior of its parent $p(j_i) \in \trianglevertices{\knownobstacledilated_i}$.
\end{definition}
Such a point is always possible to be found, since the two triangles share a common edge. By picking an admissible center, we ensure that the quadrilateral $\innerpolygon{j_i}$ with boundary $\robotposition_{3j_i}\robotposition_{1j_i}\diffeocenter{j_i}\robotposition_{2j_i}\robotposition_{3j_i}$ is convex, as shown in Fig. \ref{fig:purging}-(1a),(2a).

\begin{definition}
\label{definition:collars}
An admissible polygonal collar for the purging transformation of the leaf triangle $j_i$ is a convex polygon $\outerpolygon{j_i}$ such that:
\begin{enumerate}
    \item the edges $\robotposition_{1j_i}\diffeocenter{j_i}$ and $\diffeocenter{j_i}\robotposition_{2j_i}$ are edges of $\outerpolygon{j_i}$.
    \item $\outerpolygon{j_i}$ does not intersect the interior of any triangle $k \in \trianglevertices{\knownobstacledilated}$ with $k \neq j_i, k \neq p(j_i), \forall \knownobstacledilated$.
    \item $\outerpolygon{j_i}$ does not intersect any obstacle $\unknownobstacledilated \in \unknownobstaclesetdilatedmapped$.
    \item $\outerpolygon{j_i}$ does not intersect the boundary of $\freespacemappedpurging{j_i}$.
    \item $\innerpolygon{j_i} \subset \outerpolygon{j_i}$.
\end{enumerate}
\end{definition}

Examples of such polygons are shown in Fig. \ref{fig:purging}-(1a),(2a). This polygon is responsible for limiting the effect of the purging transformation in its interior, while keeping its value equal to the identity everywhere else. Intuitively, the requirements in Definition \ref{definition:collars} will limit the effect of the purging transformation in a region that encloses the triangle $j_i$ and is away from the boundary of any other obstacle. The importance of enforcing the edges $\robotposition_{1j_i}\diffeocenter{j_i}$ and $\diffeocenter{j_i}\robotposition_{2j_i}$ to be edges of $\outerpolygon{j_i}$ will become evident in the construction of the provable properties of the diffeomorphism, summarized below in Proposition \ref{proposition:diffeo_purging}. We provide more details about the construction of admissible collars in Appendix \ref{appendix:computational_geometry}.

For the following, we also construct implicit functions $\innerpolygonimplicit{j_i}(\robotposition)$ and $\outerpolygonimplicit{j_i}(\robotposition)$ corresponding to the leaf triangle $j_i \in \trianglevertices{\knownobstacledilated_i}$, as described in Appendix \ref{appendix:implicit}, such that
\begin{align}
    \innerpolygon{j_i} = \{\robotposition \in \mathbb{R}^2 \, | \, \innerpolygonimplicit{j_i}(\robotposition) \leq 0 \} \label{eq:implicit_gamma_ji} \\
    \outerpolygon{j_i} = \{\robotposition \in \mathbb{R}^2 \, | \, \outerpolygonimplicit{j_i}(\robotposition) \geq 0 \} \label{eq:implicit_delta_ji}
\end{align}

\subsubsection{Description of the $C^\infty$ switches}
In order to simplify the diffeomorphism construction, we depart from the construction of analytic switches \cite{Rimon_Koditschek_1989} and rely instead on the $C^\infty$ function $\zeta_\mu:\mathbb{R} \rightarrow \mathbb{R}$ \cite{hirsch_1976} described by
\begin{equation}
\label{eq:zeta}
\zeta_\mu(\chi) = \left\{ \begin{matrix}
e^{-\mu/\chi}, & \quad \chi>0 \\
0,  & \quad \chi \leq 0
\end{matrix}\right.
\end{equation}
and parametrized by $\mu > 0$, that has derivative
\begin{equation} \label{eq:zeta_derivative}
\zeta_\mu'(\chi) = \left\{ \begin{matrix}
\frac{\mu \, \zeta_\mu(\chi)}{\chi^{2}}, & \quad \chi>0 \\
0,  & \quad \chi \leq 0
\end{matrix}\right.
\end{equation}
Based on that function, we can then define the auxiliary $C^\infty$ switches
\begin{align}
    \innerpolygonsigma{j_i}(\robotposition) & := \eta_{\innerpolygontune{j_i},\innerpolygondistance{j_i}} \circ \innerpolygonimplicit{j_i}(\robotposition) \label{eq:sigma_gamma_ji} \\ 
    \outerpolygonsigma{j_i}(\robotposition) & := \zeta_{\outerpolygontune{j_i}} \circ \frac{\outerpolygonimplicit{j_i}(\robotposition)}{||\robotposition-\diffeocenter{j_i}||} \label{eq:sigma_delta_ji}
\end{align}
with $\eta_{\mu,\epsilon}(\chi) := \zeta_\mu(\epsilon - \chi)/\zeta_\mu(\epsilon)$, and $\innerpolygontune{j_i}, \outerpolygontune{j_i}, \innerpolygondistance{j_i} > 0$ tunable parameters. Notice that $\innerpolygonsigma{j_i}$ is exactly equal to 1 on the boundary of $\innerpolygon{j_i}$ and equal to 0 when $\innerpolygonimplicit{j_i}(\robotposition) \geq \innerpolygondistance{j_i}$, whereas $\outerpolygonsigma{j_i}$ is 0 outside $\outerpolygon{j_i}$. The parameters $\innerpolygontune{j_i}$ and $\outerpolygontune{j_i}$ are used to tune the ``slope'' of $\innerpolygonsigma{j_i}$ on the boundary of $\innerpolygon{j_i}$ and how fast $\outerpolygonsigma{j_i}$ approaches 1 in the interior of $\outerpolygon{j_i}$ respectively.

Based on the above, we define the {\it $C^\infty$ switch of the purging transformation for the leaf triangle $j_i \in \trianglevertices{\knownobstacledilated_i}$} as a function $\switch{j_i} : \freespacemappedpurging{j_i} \rightarrow \mathbb{R}$, defined by
\begin{equation}
    \switch{j_i}(\robotposition):= \left\{ \begin{matrix} \frac{\innerpolygonsigma{j_i}(\robotposition)\outerpolygonsigma{j_i}(\robotposition)}{\innerpolygonsigma{j_i}(\robotposition)\outerpolygonsigma{j_i}(\robotposition) + \left(1-\innerpolygonsigma{j_i}(\robotposition)\right)}, & \robotposition \neq \robotposition_{1j_i}, \robotposition_{2j_i} \\ 1, & \robotposition = \robotposition_{1j_i},\robotposition_{2j_i} \end{matrix} \right. \label{eq:sigma_ji}
\end{equation}
In this way, we see that $\switch{j_i}(\robotposition) = 0$ when $\innerpolygonsigma{j_i}(\robotposition) = 0$ or $\outerpolygonsigma{j_i}(\robotposition) = 0$ (i.e., when $\innerpolygonimplicit{j_i}(\robotposition) \geq \innerpolygondistance{j_i}$ or outside $\outerpolygon{j_i}$), $\switch{j_i}(\robotposition) = 1$ when $\innerpolygonsigma{j_i}(\robotposition) = 1$ (i.e., on the boundary of $\innerpolygon{j_i}$) and $\switch{j_i}$ varies between 0 and 1 everywhere, since $\innerpolygonsigma{j_i}$ and $\outerpolygonsigma{j_i}$ also vary between 0 and 1. Based on Definitions \ref{definition:center} and \ref{definition:collars}, it is straightforward to show the following lemma.

\begin{lemma}
\label{lemma:singular_leaf}
The function $\switch{j_i} : \freespacemappedpurging{j_i} \rightarrow \mathbb{R}$ is smooth away from the triangle vertices $\robotposition_{1j_i},\robotposition_{2j_i},\robotposition_{3j_i}$, none of which lies in the interior of $\freespacemappedpurging{j_i}$.
\end{lemma}
\begin{proof}
Included in Appendix \ref{appendix:proofs_diffeo}.
\end{proof}

\subsubsection{Description of the Deforming Factors}
The {\it deforming factors} are the functions $\deformingfactor{j_i}:\freespacemappedpurging{j_i} \rightarrow \mathbb{R}$, responsible for mapping the boundary of the leaf triangle $j_i \in \trianglevertices{\knownobstacledilated_i}$ onto the boundary of its parent $p(j_i)$. Based on Definitions \ref{definition:center} and \ref{definition:collars} and as shown in Fig. \ref{fig:purging}-(1a),(2a), this implies that the functions $\deformingfactor{j_i}$ are responsible for mapping the polygonal chain $\robotposition_{2j_i}\robotposition_{3j_i}\robotposition_{1j_i}$ onto the shared edge $\robotposition_{2j_i}\robotposition_{1j_i}$ between $j_i$ and $p(j_i)$. For this reason, we construct $\deformingfactor{j_i}$ as follows
\begin{equation}
    \deformingfactor{j_i}(\robotposition):= \frac{\left(\robotposition_{1j_i} - \diffeocenter{j_i} \right)^\top \sharednormal{j_i}}{\left(\robotposition - \diffeocenter{j_i} \right)^\top \sharednormal{j_i}} \label{eq:deforming_factor_purging}
\end{equation}
with
\begin{equation}
    \sharednormal{j_i} := \mathbf{R}_{\frac{\pi}{2}} \frac{\robotposition_{2j_i}-\robotposition_{1j_i}}{||\robotposition_{2j_i}-\robotposition_{1j_i}||}, \quad \mathbf{R}_{\frac{\pi}{2}}:= \begin{bmatrix} 0 & -1 \\ 1 & 0 \end{bmatrix}
\end{equation}
the normal vector corresponding to the shared edge between $j_i$ and $p(j_i)$.

\subsubsection{The Map Between $\freespacemappedpurging{j_i}$ and $\freespacemappedpurging{p(j_i)}$}
Based on the above, we then construct the map between $\freespacemappedpurging{j_i}$ and $\freespacemappedpurging{p(j_i)}$ with the $j_i$-th leaf triangle of $\knownobstacledilated_i$ purged, as
\begin{equation}
    \diffeopurging{j_i}(\robotposition) := \switch{j_i}(\robotposition) \left( \diffeocenter{j_i} + \deformingfactor{j_i}(\robotposition)(\robotposition-\diffeocenter{j_i}) \right) + \left(1-\switch{j_i}(\robotposition) \right) \robotposition \label{eq:map_purging}
\end{equation}

\subsubsection{Qualitative Properties of the Map Between $\freespacemappedpurging{j_i}$ and $\freespacemappedpurging{p(j_i)}$}
We first verify that the construction is a smooth change of coordinates between the intermediate mapped spaces.

\begin{lemma}
\label{lemma:purging_smooth}
The map $\diffeopurging{j_i}:\freespacemappedpurging{j_i} \rightarrow \freespacemappedpurging{p(j_i)}$ is smooth away from the triangle vertices $\robotposition_{1j_i},\robotposition_{2j_i},\robotposition_{3j_i}$, none of which lies in the interior of $\freespacemappedpurging{j_i}$.
\end{lemma}
\begin{proof}
Included in Appendix \ref{appendix:proofs_diffeo}.
\end{proof}

\begin{proposition}
\label{proposition:diffeo_purging}
The map $\diffeopurging{j_i}$ is a $C^\infty$ diffeomorphism between $\freespacemappedpurging{j_i}$ and $\freespacemappedpurging{p(j_i)}$ away from the triangle vertices $\robotposition_{1j_i},\robotposition_{2j_i},\robotposition_{3j_i}$, none of which lies in the interior of $\freespacemappedpurging{j_i}$.
\end{proposition}
\begin{proof}
Included in Appendix \ref{appendix:proofs_diffeo}.
\end{proof}

\subsubsection{Composition of Leaf Purging Transformations}
\label{subsubsec:leaf_purging_composition}
The application of the purging transformation described above will result in a tree for $\knownobstacledilated_i$ with one less vertex and one less edge. Therefore, similarly to \cite{rimon1992}, we can keep applying such purging transformations by composition, during execution time, for all leaf triangles of all obstacles $\knownobstacledilated$ in $\knownobstaclesetdilatedmappedintrusion$ and $\knownobstaclesetdilatedmappeddisk$ (in any order), until we reach their root triangles. We denote by $\freespacemappedhat$ this final intermediate space, where all obstacles in $\freespacemapped$ have been deformed to their root triangles $\{r_i\}$, and by $\diffeocomposition:\freespacemapped \rightarrow \freespacemappedhat$ the map between $\freespacemapped$ and $\freespacemappedhat$, arising from this composition of purging transformations. Since $\diffeocomposition$ is a composition of diffeomorphisms, we immediately get the following result.

\begin{corollary}
    The map $\diffeocomposition:\freespacemapped \rightarrow \freespacemappedhat$ is a $C^\infty$ diffeomorphism between $\freespacemapped$ and $\freespacemappedhat$ away from sharp corners, none of which lie in the interior of freespace for any of the intermediate or final spaces.
\end{corollary}

\subsection{Purging of Root Triangles}
\label{subsec:root_purging}
After the successive application of the leaf purging transformations presented in Section \ref{subsec:leaf_purging}, familiar obstacles in $\freespacemapped$ are reduced to triangles in $\freespacemappedhat$. These triangles either are homeomorphic to a disk in the interior of $\freespacemappedhat$ if they correspond to obstacles in $\knownobstaclesetdilatedmappeddisk$, or have a common edge with $\partial \enclosingfreespace$ if they correspond to obstacles in $\knownobstaclesetdilatedmappedintrusion$. Therefore, the final step to generate the model space $\freespacemodel$ is to transform each of the root triangles corresponding to obstacles in $\knownobstaclesetdilatedmappeddisk$ to disks, following a procedure similar to \cite{vasilopoulos_koditschek_WAFR2018}, and merge the root triangles corresponding to obstacles in $\knownobstaclesetdilatedmappedintrusion$ to $\partial \enclosingfreespace$.

\subsubsection{Center of the Transformation and Surrounding Polygonal Collars for Obstacles in $\knownobstaclesetdilatedmappeddisk$}
Here we assume that $\knownobstacledilated_i \in \knownobstaclesetdilatedmappeddisk$. Let the vertices of the root triangle $r_i \in \trianglevertices{\knownobstacledilated_i}$ be $\robotposition_{1r_i}$, $\robotposition_{2r_i}$ and $\robotposition_{3r_i}$ in counterclockwise order, as shown in Fig. \ref{fig:purging}-(1b). 

\begin{definition}
\label{definition:centers_root_disk}
An admissible center for the transformation of the root triangle $r_i$, corresponding to a polygon $\knownobstacledilated_i \in \knownobstaclesetdilatedmappeddisk$, is a point $\diffeocenter{i}$ in the interior of $r_i$.
\end{definition}

Without loss of generality, we pick $\diffeocenter{i}$ to be the barycenter of $r_i$, and the {\it radius of the transformation} to be a number $\diffeoradius{i} < d(\diffeocenter{i}, \partial r_i)$, following the admissibility assumptions made in \cite{rimon1992}. We also set $\innerpolygon{r_i}$ to be the closure of $r_i$ itself, and define a polygonal collar $\outerpolygon{r_i}$ for the root triangle transformation as follows.
\begin{definition}
\label{definition:collars_root_disk}
An admissible polygonal collar for the transformation of the root triangle $r_i$, corresponding to a polygon $\knownobstacledilated_i \in \knownobstaclesetdilatedmappeddisk$, is a convex polygon $\outerpolygon{r_i}$ such that:
\begin{enumerate}
    \item $\outerpolygon{r_i} \cap \outerpolygon{r_j} = \varnothing$, with $i \neq j$.
    \item $\outerpolygon{r_i}$ does not intersect any obstacle $\unknownobstacledilated \in \unknownobstaclesetdilatedmapped$.
    \item $\outerpolygon{r_i}$ does not intersect $\partial \freespacemappedhat$.
    \item $\innerpolygon{r_i} \subset \outerpolygon{r_i}$.
\end{enumerate}
\end{definition}
An example of such a polygon is shown in Fig. \ref{fig:purging}-(1b). Again, this polygon is responsible for limiting the effect of the transformation in its interior, while keeping it equal to the identity map everywhere else. Similarly to Section \ref{subsec:leaf_purging}, we also construct implicit functions $\innerpolygonimplicit{r_i}(\robotposition)$ and $\outerpolygonimplicit{r_i}(\robotposition)$ for each root triangle $r_i$, such that
\begin{align}
    \innerpolygon{r_i} = \{\robotposition \in \mathbb{R}^2 \, | \, \innerpolygonimplicit{r_i}(\robotposition) \leq 0 \} \label{eq:implicit_gamma_ri}\\
    \outerpolygon{r_i} = \{\robotposition \in \mathbb{R}^2 \, | \, \outerpolygonimplicit{r_i}(\robotposition) \geq 0 \label{eq:implicit_delta_ri} \}
\end{align}

\subsubsection{Description of the $C^\infty$ switches for Obstacles in $\knownobstaclesetdilatedmappeddisk$}
Following the notation of Section \ref{subsec:leaf_purging}, we can define the auxiliary $C^\infty$ switches
\begin{align}
    \innerpolygonsigma{r_i}(\robotposition) & := \eta_{\innerpolygontune{r_i},\innerpolygondistance{r_i}} \circ \innerpolygonimplicit{r_i}(\robotposition) \label{eq:sigma_gamma_ri} \\
    \outerpolygonsigma{r_i}(\robotposition) & := \zeta_{\outerpolygontune{r_i}} \circ \frac{\outerpolygonimplicit{r_i}(\robotposition)}{||\robotposition-\diffeocenter{i}||} \label{eq:sigma_delta_ri}
\end{align}
with $\eta_{\mu,\epsilon}(\chi) := \zeta_\mu(\epsilon - \chi)/\zeta_\mu(\epsilon)$, $\zeta$ defined as in \eqref{eq:zeta} and $\innerpolygontune{r_i}, \outerpolygontune{r_i}, \innerpolygondistance{r_i} > 0$ tunable parameters.

Based on the above, we then define the {\it $C^\infty$ switch of the transformation of the root triangle $r_i$} as the function $\switch{r_i}:\freespacemappedhat \rightarrow \freespacemapped$ given by
\begin{equation}
    \switch{r_i}(\robotposition):= \frac{\innerpolygonsigma{r_i}(\robotposition) \outerpolygonsigma{r_i}(\robotposition)}{\innerpolygonsigma{r_i}(\robotposition) \outerpolygonsigma{r_i}(\robotposition) + \left(1-\innerpolygonsigma{r_i}(\robotposition)\right)} \label{eq:sigma_ri}
\end{equation}
It can be seen that the function $\switch{r_i}$ will be 0 outside $\outerpolygon{r_i}$, exactly equal to 1 on the boundary of $\innerpolygon{r_i}$ (i.e., on the boundary of the root triangle $r_i$) and varies smoothly between 0 and 1 everywhere else.

We can easily show the following lemma, as the function $\zeta$ eliminates all the singular points of $\outerpolygonimplicit{r_i}$ that correspond to the vertices of $\outerpolygon{r_i}$.
\begin{lemma}
\label{lemma:singular_root}
The switch $\switch{r_i} : \freespacemappedhat \rightarrow \mathbb{R}$ is smooth away from the triangle vertices $\robotposition_{1r_i},\robotposition_{2r_i},\robotposition_{3r_i}$, none of which lies in the interior of $\freespacemappedhat$.
\end{lemma}

\subsubsection{Description of the Deforming Factors for Obstacles in $\knownobstaclesetdilatedmappeddisk$}
Here, the deforming factors are the functions $\deformingfactor{r_i}:\freespacemappedhat \rightarrow \mathbb{R}$, responsible for transforming each root triangle corresponding to an obstacle in $\knownobstaclesetdilatedmappeddisk$ to a disk in $\mathbb{R}^2$. The deforming factors we use are inspired by those in \cite{rimon1992}, but do not depend on the values of the implicit functions $\innerpolygonimplicit{r_i}$. Namely, the deforming factors are given based on the desired final radii $\rho_i$ as
\begin{equation}
    \deformingfactor{r_i}(\robotposition) := \frac{\diffeoradius{i}}{||\robotposition-\diffeocenter{i}||} \label{eq:deforming_factor_disk}
\end{equation}

\subsubsection{Center of the Transformation and Surrounding Polygonal Collars for Obstacles in $\knownobstaclesetdilatedmappedintrusion$}
Next we focus on obstacles in $\knownobstaclesetdilatedmappedintrusion$. The procedure here is slightly different, since we want to merge the root triangle $r_i$ to the boundary of the enclosing freespace $\enclosingfreespace$. Namely, we assume that the vertices of the triangle $r_i$ are $\robotposition_{1r_i}, \robotposition_{2r_i}, \robotposition_{3r_i}$ in counterclockwise order, with $\robotposition_{1r_i}\robotposition_{2r_i}$ the common edge between $r_i$ and $\partial \enclosingfreespace$, as shown in Fig. \ref{fig:purging}-(2b)\footnote{If $r_i$ and $\partial \enclosingfreespace$ share two common edges, we just pick one of them at random.}. Then, we pick an admissible center similarly to Definition \ref{definition:center}, as follows.

\begin{definition}
\label{definition:centers_root_boundary}
An admissible center for the transformation of the root triangle $r_i$, corresponding to a polygon $\knownobstacledilated_i \in \knownobstaclesetdilatedmappedintrusion$, denoted by $\diffeocenter{i}$, is a point along the triangle median from $\robotposition_{3r_i}$, such that $\diffeocenter{i} \in \mathbb{R}^2 \backslash \enclosingfreespace$.
\end{definition}

We also define $\innerpolygon{r_i}$ to be the convex quadrilateral with boundary $\robotposition_{3r_i}\robotposition_{1r_i}\diffeocenter{i}\robotposition_{2r_i}\robotposition_{3r_i}$. The collars used are defined similarly to Definition \ref{definition:collars}, as the transformation itself is designed to be quite similar with the purging transformation.
\begin{definition}
\label{definition:collars_root_boundary}
An admissible polygonal collar for the transformation of the root triangle $r_i$, corresponding to a polygon $\knownobstacledilated_i \in \knownobstaclesetdilatedmappedintrusion$, is a convex polygon $\outerpolygon{r_i}$ such that:
\begin{enumerate}
    \item the edges $\robotposition_{1r_i}\diffeocenter{i}$ and $\diffeocenter{i}\robotposition_{2r_i}$ are edges of $\outerpolygon{r_i}$.
    \item $\outerpolygon{r_i} \cap \outerpolygon{r_j} = \varnothing$, with $i \neq j$.
    \item $\outerpolygon{r_i}$ does not intersect any obstacle $\unknownobstacledilated \in \unknownobstaclesetdilatedmapped$.
    \item $\innerpolygon{r_i} \subset \outerpolygon{r_i}$.
\end{enumerate}
\end{definition}

\subsubsection{Description of the $C^\infty$ switches for Obstacles in $\knownobstaclesetdilatedmappedintrusion$}
With the definition of $\innerpolygon{r_i}$ and $\outerpolygon{r_i}$ as described above, we associate implicit functions $\innerpolygonimplicit{r_i}(\robotposition)$ and $\outerpolygonimplicit{r_i}(\robotposition)$ as in \eqref{eq:implicit_gamma_ri}, \eqref{eq:implicit_delta_ri}, auxiliary switches $\innerpolygonsigma{r_i}$ and $\outerpolygonsigma{r_i}$ as in \eqref{eq:sigma_gamma_ri} and \eqref{eq:sigma_delta_ri}, and overall {\it $C^\infty$ switch of the transformation of the root triangle $r_i$} as the function $\switch{r_i}:\freespacemappedhat \rightarrow \freespacemapped$ given in \eqref{eq:sigma_ri}.

\subsubsection{Description of the Deforming Factors for Obstacles in $\knownobstaclesetdilatedmappedintrusion$}
Finally, in order to merge the root triangle into the boundary $\partial \enclosingfreespace$, we define the deforming factors similarly to \eqref{eq:deforming_factor_purging}, as the functions $\deformingfactor{r_i}:\freespacemappedhat \rightarrow \mathbb{R}$, given by
\begin{equation}
    \deformingfactor{r_i}(\robotposition):= \frac{\left(\robotposition_{1r_i} - \diffeocenter{i} \right)^\top \sharednormal{r_i}}{\left(\robotposition - \diffeocenter{i} \right)^\top \sharednormal{r_i}} \label{eq:deforming_factor_root_boundary}
\end{equation}
with
\begin{equation}
    \sharednormal{r_i} := \mathbf{R}_{\frac{\pi}{2}} \frac{\robotposition_{2r_i}-\robotposition_{1r_i}}{||\robotposition_{2r_i}-\robotposition_{1r_i}||}, \quad \mathbf{R}_{\frac{\pi}{2}}:= \begin{bmatrix} 0 & -1 \\ 1 & 0 \end{bmatrix}
\end{equation}
the normal vector corresponding to the shared edge between $r_i$ and $\partial \enclosingfreespace$.

\subsubsection{The Map Between $\freespacemappedhat$ and $\freespacemodel$}
\label{subsubsec:root_purging}
First of all, we define
\begin{equation}
    \switchresidual(\robotposition) := 1-\sum_{i \in \knownobstaclesetdilatedmappedintrusionindex \cup \knownobstaclesetdilatedmappeddiskindex} \switch{r_i}(\robotposition)
\end{equation}
Using the above constructions and Definitions \ref{definition:centers_root_disk}, \ref{definition:collars_root_disk}, \ref{definition:centers_root_boundary} and \ref{definition:collars_root_boundary} we are led to the following results.
\begin{lemma} \label{lemma:switches_nonzero}
At any point $\robotposition \in \freespacemappedhat$, at most one of the switches $\{ \switch{r_i} \}_{i \in \knownobstaclesetdilatedmappedintrusionindex \cup \knownobstaclesetdilatedmappeddiskindex}$ can be nonzero.
\end{lemma}

\begin{corollary}
    The set $\{ \switch{r_i} \}_{i \in \knownobstaclesetdilatedmappedintrusionindex \cup \knownobstaclesetdilatedmappeddiskindex} \cup \{\switchresidual\}$ defines a partition of unity over $\freespacemappedhat$.
\end{corollary}

With the construction of $\switch{r_i}$ (as in \eqref{eq:sigma_ri}) and $\deformingfactor{r_i}$ (as in either \eqref{eq:deforming_factor_disk} or \eqref{eq:deforming_factor_root_boundary}, depending on whether $i$ belongs to $\knownobstaclesetdilatedmappeddiskindex$ or $\knownobstaclesetdilatedmappedintrusionindex$ respectively) for each root triangle, we can now construct the map $\diffeoroot:\freespacemappedhat \rightarrow \freespacemodel$ given by
\begin{equation}
    \diffeoroot(\robotposition):=\sum_{i \in \knownobstaclesetdilatedmappedintrusionindex \cup \knownobstaclesetdilatedmappeddiskindex} \switch{r_i}(\robotposition) \left[ \diffeocenter{i} + \deformingfactor{r_i}(\robotposition)(\robotposition-\diffeocenter{i}) \right] + \switchresidual(\robotposition) \robotposition \label{eq:map_root}
\end{equation}

\subsubsection{Qualitative Properties of the Map Between $\freespacemappedhat$ and $\freespacemodel$}
We can again verify that the construction is a smooth change of coordinates between $\freespacemappedhat$ and $\freespacemodel$. Using Lemma \ref{lemma:singular_root} and the fact that the deforming factors $\deformingfactor{r_i}$ are smooth in $\freespacemappedhat$ (because the centers $\diffeocenter{i}$ do not belong in $\freespacemappedhat$) for all $i$, we get the following result.
\begin{lemma} \label{lemma:root_smooth}
The map $\diffeoroot : \freespacemappedhat \rightarrow \freespacemodel$ is smooth away from any sharp corners, none of which lie in the interior of $\freespacemappedhat$.
\end{lemma}

\begin{proposition}
\label{proposition:diffeo_root}
The map $\diffeoroot$ is a $C^\infty$ diffeomorphism between $\freespacemappedhat$ and $\freespacemodel$ away from any sharp corners, none of which lie in the interior of $\freespacemappedhat$.
\end{proposition}
\begin{proof}
Included in Appendix \ref{appendix:proofs_diffeo}.
\end{proof}

\begin{figure}[H]
\centering
\includegraphics[width=0.6\textwidth]{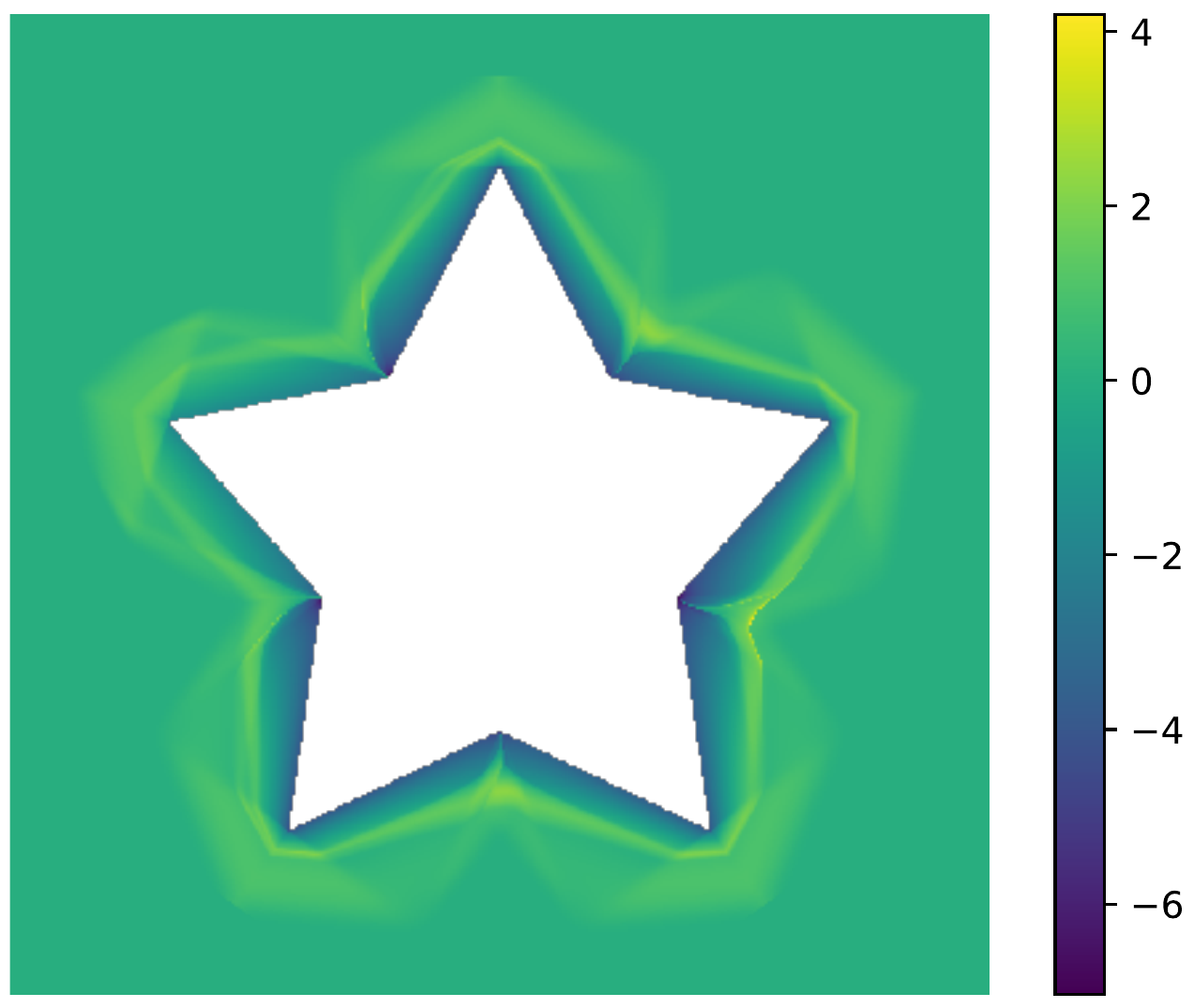}
\caption{Values of $\text{det}(D_\robotposition \diffeo)$ for a single polygon in logarithmic scale, showing the local nature of the diffeomorphism ($\diffeo$ becomes equal to the identity transform away from the polygon) and the fact that $\diffeo$ is smooth away from sharp corners, that do not lie in the interior of the freespace.} \label{fig:heatmap}
\end{figure}

\subsection{The Map Between the Mapped Space and the Model Space}

\label{subsec:diffeo_description}
Based on the construction of $\diffeocomposition:\freespacemapped \rightarrow \freespacemappedhat$ in Section \ref{subsubsec:leaf_purging_composition} and $\diffeoroot:\freespacemappedhat \rightarrow \freespacemodel$ in Section \ref{subsubsec:root_purging}, we can finally write the map between the mapped space and the model space as the function $\diffeo : \freespacemapped \rightarrow \freespacemodel$ given by 
\begin{equation}
    \diffeo(\robotposition) = \diffeoroot \circ \diffeocomposition(\robotposition) \label{eq:diffeo_final}
\end{equation}
It is straightforward to get the following result, since both $\diffeocomposition$ and $\diffeoroot$ are $C^\infty$ diffeomorphisms away from sharp corners.
\begin{corollary} \label{corollary:diffeo_full}
    The map $\diffeo$ is a $C^\infty$ diffeomorphism between $\freespacemapped$ and $\freespacemodel$ away from any sharp corners, none of which lie in the interior of $\freespacemapped$.
\end{corollary}

An illustration of the behavior of the map $\diffeo$ through a visualization of the values of $\text{det}(D_\robotposition \diffeo)$ for a specific example with a single polygon is included in Fig. \ref{fig:heatmap}.
\section{Reactive Controller}
\label{sec:controller}

\begin{table}[htbp]
    \centering
    \begin{tabular}{l l}
        \toprule
        \toprule
        $\hybridmode \in 2^{\knownobstaclesetindex}$ & Mode of the hybrid system, Sections \ref{subsec:physical_space}, \ref{subsec:control_hybrid} \\
        $\hybridfreespacemode, \hybridfreespacemodesemantic, \hybridfreespacemodemapped, \hybridfreespacemodemodel \in \mathbb{R}^2$ & Physical, semantic, mapped, model freespace for $\hybridmode$, \\
        & Sections \ref{sec:environment_representation}, \ref{subsec:control_hybrid} \\
        $\diffeo : \hybridfreespacemodemapped \rightarrow \hybridfreespacemodemodel$ & Map between $\hybridfreespacemodemapped$ and $\hybridfreespacemodemodel$ for mode $\hybridmode$, \\
        & Section \ref{sec:diffeomorphism} \\
        $\hybridgraph \subset 2^{\knownobstaclesetindex} \times 2^{\knownobstaclesetindex}$ & Directed graph of discrete transitions over the modes, \\
        & Section \ref{subsec:control_hybrid} \\
        $\hybriddomain := \bigsqcup_{\hybridmode \in 2^{\knownobstaclesetindex}} \hybridfreespacemode$ & Collection of domains for a fully actuated robot, \\
        & Section \ref{subsec:control_hybrid} \\
        $\hybridguard:=\bigsqcup_{(\hybridmode,\hybridmode') \in \hybridgraph} \hybridguardrestriction^{\hybridmode,\hybridmode'}$ & Collection of guards for a fully actuated robot \eqref{eq:restriction_guard} \\
        $\hybridreset:\hybridguard \rightarrow \hybriddomain$ & Continuous reset map for a fully actuated robot \eqref{eq:restriction_reset} \\
        $\hybridfield : \hybriddomain \rightarrow T\hybriddomain$ & Hybrid vector field for a fully actuated robot \eqref{eq:control_fullyactuated_hybrid} \\
        $\hybridsystem := \left(2^{\knownobstaclesetindex}, \hybridgraph, \hybriddomain, \hybridfield, \hybridguard, \hybridreset \right)$ & Hybrid system for a fully actuated robot, Theorem \ref{theorem:hybrid_fullyactuated} \\
        $\hybriddomainunicycle := \bigsqcup_{\hybridmode \in 2^{\knownobstaclesetindex}} (\hybridfreespacemode \times S^1)$ & Collection of domains for a differential drive robot, \\
        & Section \ref{subsec:control_hybrid} \\
        $\hybridguardunicycle:=\bigsqcup_{(\hybridmode,\hybridmode') \in \hybridgraph} \hybridguardrestrictionunicycle^{\hybridmode,\hybridmode'}$ & Collection of guards for a differential drive robot \eqref{eq:restriction_guard_unicycle} \\
        $\hybridresetunicycle:\hybridguardunicycle \rightarrow \hybriddomainunicycle$ & Continuous reset map for a differential drive robot \eqref{eq:restriction_reset_unicycle} \\
        $\hybridfieldunicycle : \hybriddomainunicycle \rightarrow T\hybriddomainunicycle$ & Hybrid vector field for a differential drive robot \eqref{eq:control_unicycle_hybrid} \\
        $\hybridsystemunicycle := \left(2^{\knownobstaclesetindex}, \hybridgraph, \hybriddomainunicycle, \hybridfieldunicycle, \hybridguardunicycle, \hybridresetunicycle \right)$ & Hybrid system for a differential drive robot, Theorem \ref{theorem:hybrid_unicycle} \\
        \hline \\
        $\robotposition \in \freespacemapped$ & Fully actuated robot position in $\freespacemapped$, Section \ref{subsec:control_modes} \\
        $\robotpositionmodel:=\diffeo(\robotposition) \in \freespacemodel$ & Fully actuated robot position in $\freespacemodel$, Section \ref{subsec:control_modes} \\
        $\goalpositionmodel := \diffeo (\goalposition)$ & Goal position in $\freespacemodel$, Section \ref{subsec:control_modes} \\
        $\localfreespace{\robotpositionmodel} \subset \freespacemodel$ & Local freespace at $\robotpositionmodel \in \freespacemodel$ \eqref{eq:local_freespace} \\
        $\controlfullyactuatedmodel^\hybridmode : \freespacemodel \rightarrow T\freespacemodel$ & Vector field controller for a fully actuated robot \\
        & in $\freespacemodel$ \eqref{eq:control_fullyactuated_model} \\
        $\controlfullyactuated^\hybridmode : \freespacemapped \rightarrow T \freespacemapped$ & Vector field controller for a fully actuated robot \\
        & in $\freespacemapped$ \eqref{eq:control_fullyactuated} \\
        $\diffeounicycle : \freespacemapped \times S^1 \rightarrow \freespacemodel \times S^1$ & Diffeomorphism between $\freespacemapped \times S^1$ and \\
        & $\freespacemodel \times S^1$ \eqref{eq:diffeo_unicycle} \\
        $\robotpositionunicycle := (\robotposition, \psi) \in \freespacemapped \times S^1$ & Differential drive robot state in $\freespacemapped \times S^1$, \\
        & Section \ref{subsec:control_modes} \\
        $\robotpositionunicyclemodel := (\robotpositionmodel, \robotorientationmodel) = \diffeounicycle(\robotpositionunicycle) \in \freespacemodel \times S^1$ & Differential drive robot state in $\freespacemodel \times S^1$, \\
        & Section \ref{subsec:control_modes} \\
        $\angletransform:S^1 \rightarrow S^1$ & Angle transformation between $\freespacemapped \times S^1$ and \\
        & $\freespacemodel \times S^1$ \eqref{eq:phi} \\
        $\localgoallinear, \localgoalangular \in \freespacemodel$ & Linear \eqref{eq:linearlocalgoal} and angular \eqref{eq:angularlocalgoal} local goals for \\
        & $\robotpositionunicyclemodel \in \freespacemodel \times S^1$ \\
        $\controlunicyclemodel^\hybridmode := (\linearinputmodel^\hybridmode,\angularinputmodel^\hybridmode) \in \mathbb{R}^2$ & Linear and angular inputs for a unicycle robot in \\
        & $\freespacemodel \times S^1$ \eqref{eq:control_unicycle_reference} \\
        $\controlunicycle^\hybridmode := (\linearinput^\hybridmode,\angularinput^\hybridmode) \in \mathbb{R}^2$ & Linear and angular inputs for a unicycle robot in \\
        & $\freespacemapped \times S^1$ \eqref{eq:control_unicycle} \\
        \hline
    \end{tabular}
    \caption{Key symbols related to the hybrid systems formulation (top - Section \ref{subsec:control_hybrid}) and the reactive controller construction in each mode of the hybrid system (bottom - Section \ref{subsec:control_modes}) for both a fully actuated robot and a differential drive robot.}
    \label{table:notation_controller}
\end{table}

The preceding analysis in Section \ref{sec:diffeomorphism} describes the diffeomorphism construction between $\freespacemapped$ and $\freespacemodel$ for a given index set $\hybridmode$ of instantiated familiar obstacles. However, the onboard sensor might discover new obstacles and, subsequently, incorporate them in the semantic map, updating the set $\hybridmode$. Therefore, we have to provide a hybrid systems description of our reactive controller, where each mode is defined by an index set $\hybridmode \in 2^{\knownobstaclesetindex}$ of familiar obstacles stored in the semantic map, the guards describe the sensor trigger events where a previously ``unexplored'' obstacle is discovered and incorporated in the semantic map (thereby changing $\knownobstaclesetdilatedmapped$, along with $\knownobstaclesetdilatedmappeddisk$, $\knownobstaclesetdilatedmappedintrusion$), and the resets describe transitions to new modes that might result in discrete ``jumps'' of the robot position in the model space. We then need to show that the resulting hybrid controller, for both the fully actuated robot and the differential drive robot, must succeed in the navigation task. 

In the following, Section \ref{subsec:control_hybrid} provides the hybrid systems description, Section \ref{subsec:control_modes} describes the reactive controller applied in each mode of the hybrid system, Section \ref{subsec:hybrid_qualitative} summarizes the qualitative properties of our hybrid controller, and Section \ref{subsec:bounded_inputs} describes our method of generating bounded inputs, each time for both the fully actuated and the differential drive robot. Table \ref{table:notation_controller} summarizes associated notation used throughout this Section.

\subsection{Hybrid Systems Description of Navigation Framework}
\label{subsec:control_hybrid}

\subsubsection{Fully Actuated Robots}
First, we consider a fully actuated particle with state $\robotposition \in \freespace$, and dynamics
\begin{equation}
    \dot{\robotposition} = \controlfullyactuated \label{eq:dynamics_fullyactuated}
\end{equation}
Since different subsets of instantiated obstacles in $\knownobstacleset$, indexed by $\hybridmode$, result in different consolidated polygonal obstacles stored in $\knownobstaclesetdilatedmapped$, it is natural to index the {\it modes of the hybrid controller} according to elements $\hybridmode$ of the power set $2^{\knownobstaclesetindex}$. Every execution, from any initial state, is required to start in the {\it initial mode}, indexed by $\hybridmode = \varnothing$. We also define a {\it terminal mode} as follows.
\begin{definition}
\label{definition:terminal_mode}
    The terminal mode of the hybrid system is indexed by the improper subset, $\hybridmode = \knownobstaclesetindex$, where all familiar obstacles in the workspace have been instantiated in the set $\knownobstaclesetdilatedsemantic$, in the sense of Definition \ref{definition:instantiation}.
\end{definition}

We denote the freespace in the semantic, mapped and model spaces, associated with a unique subset $\hybridmode$ of $\knownobstaclesetindex$, by $\hybridfreespacemodesemantic, \hybridfreespacemodemapped, \hybridfreespacemodemodel$ respectively, as in Section \ref{sec:environment_representation}. We also denote the corresponding perceived physical freespace by $\hybridfreespacemode$, with $\hybridfreespacemode := \hybridfreespacemodemapped$, since the dilation of obstacles by $\robotradius$ in the passage from the physical to the semantic space and the obstacle merging in the passage from the semantic to the mapped space do not alter the freespace description. The domain $\hybriddomain$ of our hybrid system is then defined as the collection $\hybriddomain := \bigsqcup_{\hybridmode \in 2^{\knownobstaclesetindex}} \hybridfreespacemode$.

Following the notation in \cite{Johnson_Burden_Koditschek_2016}, we can then denote by $\hybridgraph \subset 2^{\knownobstaclesetindex} \times 2^{\knownobstaclesetindex}$ the {\it set of discrete transitions} for the hybrid system, forming a directed graph structure over the set of modes $2^{\knownobstaclesetindex}$. The collection of {\it guards} associated with $\hybridgraph$ can be described as $\hybridguard:=\bigsqcup_{(\hybridmode,\hybridmode') \in \hybridgraph} \hybridguardrestriction^{\hybridmode,\hybridmode'}$, with $\hybridguardrestriction^{\hybridmode,\hybridmode'} \subset \hybridfreespacemode$ given by
\begin{align}
    \hybridguardrestriction^{\hybridmode,\hybridmode'} := & \{ \robotposition \in \hybridfreespacemode \, | \, \hybridmode' = \hybridmode \cup \hybridmode_u \nonumber \\
    & \hybridmode_u \neq \varnothing, \hybridmode_u \cap \hybridmode = \varnothing, \nonumber \\ 
    & \ballclosure{\robotposition}{\sensorrange} \cap \knownobstacle_i \neq \varnothing \, \, \text{for all} \, \, i \in \hybridmode_u, \nonumber \\ 
    & \ballclosure{\robotposition}{\sensorrange} \cap \knownobstacleset_{\knownobstaclesetindex \backslash (\hybridmode \cup \hybridmode_u)}  = \varnothing \} \label{eq:restriction_guard}
\end{align}
with $\knownobstacleset_{\knownobstaclesetindex \backslash (\hybridmode \cup \hybridmode_u)} := \{\knownobstacle_i\}_{i \in \knownobstaclesetindex \backslash (\hybridmode \cup \hybridmode_u)}$.

Also, the {\it reset} $\hybridreset:\hybridguard \rightarrow \hybriddomain$ is the continuous map that restricts simply as $\hybridresetrestriction^{\hybridmode,\hybridmode'} := \hybridreset|_{\hybridguardrestriction^{\hybridmode,\hybridmode'}} : \hybridguardrestriction^{\hybridmode,\hybridmode'} \rightarrow \hybridfreespacemodeprime$, with
\begin{equation}
\hybridresetrestriction^{\hybridmode,\hybridmode'}(\robotposition) = \robotposition \label{eq:restriction_reset}
\end{equation}
the identity map. Note, however, that although the robot cannot experience discrete jumps in the physical space, the model space $\hybridfreespacemodemodelprime$ is likely to be a discontinuously different space from $\hybridfreespacemodemodel$ (i.e., there is no guaranteed inclusion from $\hybridfreespacemodemodelprime$ into $\hybridfreespacemodemodel$), hence the model position in the new space bears no obvious relationship to that in the prior. Namely, the position of the robot in the model space after a transition from mode $\hybridmode$ to mode $\hybridmode'$ will be given by $\diffeogeneric^{\hybridmode'} \circ (\diffeogeneric^{\hybridmode})^{-1}(\robotpositionmodel)$, with $\robotpositionmodel \in \hybridfreespacemodemodel$.

Finally, we can construct the {\it hybrid vector field} $\hybridfield : \hybriddomain \rightarrow T\hybriddomain$ that restricts to a vector field $\hybridfieldrestriction^{\hybridmode}:=\hybridfield|_{\hybridfreespacemode} : \hybridfreespacemode \rightarrow T\hybridfreespacemode$, that can be written as
\begin{equation}
    \hybridfieldrestriction^{\hybridmode} (\robotposition) := \controlfullyactuated^\hybridmode (\robotposition) \label{eq:control_fullyactuated_hybrid}
\end{equation}
with $\controlfullyactuated^\hybridmode$ given in \eqref{eq:control_fullyactuated} and described in the next Section.

Based on the above definitions, we define the {\it navigational hybrid system for fully actuated robots} as the tuple $\hybridsystem := \left(2^{\knownobstaclesetindex}, \hybridgraph, \hybriddomain, \hybridfield, \hybridguard, \hybridreset \right)$ describing the modes, discrete transitions, domains, associated vector fields, guards and resets.

\subsubsection{Differential Drive Robots}
Next, we focus on a differential drive robot, whose state is $\robotpositionunicycle:=(\robotposition,\robotorientation) \in \freespace \times S^1 \subset SE(2)$, and its dynamics are given by\footnote{We use the ordered set notation $(*,*,\ldots)$ and the matrix notation $\begin{bmatrix} * & * & \ldots \end{bmatrix}^\top$ for vectors interchangeably.}
\begin{equation}
    \dot{\robotpositionunicycle} = \mathbf{B}(\robotorientation) \controlunicycle \label{eq:unicycle_dynamics}
\end{equation}
with $\mathbf{B}(\robotorientation):=\begin{bmatrix} \cos\robotorientation & \sin\robotorientation & 0 \\ 0 & 0 & 1 \end{bmatrix}^\top$ and $\controlunicycle:=(\linearinput,\angularinput)$, with $\linearinput,\angularinput \in \mathbb{R}$ the linear and angular input respectively.

The analysis here is fairly similar; the modes and discrete transitions are identical. However, the robot operates on a subset of $SE(2)$ and, therefore, the domains must be described as $\hybriddomainunicycle:=\bigsqcup_{\hybridmode \in 2^{\knownobstaclesetindex}} (\hybridfreespacemode \times S^1)$. Consequently, the collection of {\it guards} that will result in transitions between different modes according to $\hybridgraph$ are described as $\hybridguardunicycle:=\bigsqcup_{(\hybridmode,\hybridmode') \in \hybridgraph} \hybridguardrestrictionunicycle^{\hybridmode,\hybridmode'}$, with $\hybridguardrestrictionunicycle^{\hybridmode,\hybridmode'} \subset (\hybridfreespacemode \times S^1)$ given by
\begin{align}
    \hybridguardrestrictionunicycle^{\hybridmode,\hybridmode'} := & \{ \robotpositionunicycle = (\robotposition,\robotorientation) \in \hybridfreespacemode \times S^1 \, | \, \hybridmode' = \hybridmode \cup \hybridmode_u \nonumber \\
    & \hybridmode_u \neq \varnothing, \hybridmode_u \cap \hybridmode = \varnothing, \nonumber \\
    & \ballclosure{\robotposition}{\sensorrange} \cap \knownobstacle_i \neq \varnothing \, \, \text{for all} \, \, i \in \hybridmode_u, \nonumber \\ 
    & \ballclosure{\robotposition}{\sensorrange} \cap \knownobstacleset_{\knownobstaclesetindex \backslash (\hybridmode \cup \hybridmode_u)}  = \varnothing  \} \label{eq:restriction_guard_unicycle}
\end{align}

Also, the {\it reset} $\hybridresetunicycle:\hybridguardunicycle \rightarrow \hybriddomainunicycle$ is the continuous map that restricts simply as $\hybridresetrestrictionunicycle^{\hybridmode,\hybridmode'} := \hybridresetunicycle|_{\hybridguardrestrictionunicycle^{\hybridmode,\hybridmode'}} : \hybridguardrestrictionunicycle^{\hybridmode,\hybridmode'} \rightarrow (\hybridfreespacemodeprime \times S^1)$, with
\begin{equation}
\hybridresetrestrictionunicycle^{\hybridmode,\hybridmode'}(\robotpositionunicycle) = \robotpositionunicycle \label{eq:restriction_reset_unicycle}
\end{equation}
the identity map.

Finally, the fact that the robot operates in $SE(2)$ gives rise to a new {\it hybrid vector field} $\hybridfieldunicycle : \hybriddomainunicycle \rightarrow T\hybriddomainunicycle$, that restricts to a vector field $\hybridfieldrestrictionunicycle^{\hybridmode} := \hybridfieldunicycle|_{\hybridfreespacemode \times S^1} : \hybridfreespacemode \times S^1 \rightarrow T(\hybridfreespacemode \times S^1)$, that can be written as
\begin{equation}
    \hybridfieldrestrictionunicycle^{\hybridmode} (\robotpositionunicycle) := \mathbf{B}(\robotorientation) \controlunicycle^\hybridmode \label{eq:control_unicycle_hybrid}
\end{equation}
with the inputs $\controlunicycle^\hybridmode = (\linearinput^\hybridmode, \angularinput^\hybridmode)$ given as in \eqref{eq:control_unicycle} and described in the next Section.

Based on the above definitions, we define the {\it navigational hybrid system for differential drive robots} as the tuple $\hybridsystemunicycle := \left(2^{\knownobstaclesetindex}, \hybridgraph, \hybriddomainunicycle, \hybridfieldunicycle, \hybridguardunicycle, \hybridresetunicycle \right)$ describing the modes, discrete transitions, domains, associated vector fields, guards and resets.

\subsection{Reactive Controller in Each Hybrid Mode}
\label{subsec:control_modes}
The preceding analysis of the hybrid system allows us to now describe the constituent controllers in each mode $\hybridmode$ of the hybrid system, for both the fully actuated and the differential drive robot. For the results pertaining to each separate mode, we are going to assume that $\hybridmode$ describes the terminal mode of the hybrid system, in the notion of Definition \ref{definition:terminal_mode}.

With this assumption, we can arrive to Theorems \ref{theorem:control_fullyactuated} and \ref{theorem:control_se2}, that allow us to establish the main results about our hybrid controller in Theorems \ref{theorem:hybrid_fullyactuated} and \ref{theorem:hybrid_unicycle}. We assume that the robot operates in $\hybridfreespacemodemapped$\footnote{This is afforded by the fact that the perceived physical freespace $\hybridfreespacemode$ was explicitly constructed in Section \ref{subsec:control_hybrid} to be equal to $\hybridfreespacemodemapped$. To be accurate, one must write the identity map from $\hybridfreespacemode$ into $\hybridfreespacemodemapped$ as $\iota: \hybridfreespacemode \rightarrow \hybridfreespacemodemapped$ and, subsequently, define the control in the physical space as $[D_{\robotposition}\iota]^{-1} \controlfullyactuated^{\hybridmode}$, with $\controlfullyactuated^{\hybridmode} : \hybridfreespacemodemapped \rightarrow T\hybridfreespacemodemapped$ the control strategy in the mapped space, described next. However, since $[D_{\robotposition}\iota]$ resolves to the identity matrix, this construction reduces to direct application of $\controlfullyactuated^{\hybridmode}$ on the physical space.}, and the set of consolidated obstacles $\knownobstaclesetdilatedmapped$ in mode $\hybridmode$ has been identified.

\subsubsection{Fully Actuated Robots}
The dynamics of the fully actuated particle in $\hybridfreespacemodemodel$ with state $\robotpositionmodel = \diffeo(\robotposition) \in \hybridfreespacemodemodel$ can be described by $\dot{\robotpositionmodel} = \controlfullyactuatedmodel^\hybridmode(\robotpositionmodel)$ with the input $\controlfullyactuatedmodel^\hybridmode(\robotpositionmodel)$ given in \cite{arslan_kod_WAFR2016} as\footnote{Here $\projection{C}{\mathbf{q}}$ denotes the metric projection of $\mathbf{q}$ on a convex set $C$.}
\begin{equation}
    \controlfullyactuatedmodel^\hybridmode(\robotpositionmodel) = - \left(\robotpositionmodel - \projection{\localfreespace{\robotpositionmodel}}{\goalpositionmodel} \right) \label{eq:control_fullyactuated_model}
\end{equation}
with $\goalpositionmodel = \diffeo(\goalposition)$, and the convex {\it local freespace} for $\robotpositionmodel$, $\localfreespace{\robotpositionmodel}$, defined as the Voronoi cell in \cite[Eqns. (7), (24)]{arslan_kod_WAFR2016}:
\begin{equation}
    \localfreespace{\robotpositionmodel}:= \left\{ \mathbf{q} \in \hybridfreespacemodemodel \, | \, || \mathbf{q}-\robotpositionmodel|| \leq ||\mathbf{q}-\projection{\overline{\obstacledilated}_i}{\robotpositionmodel} || , \forall i \right\} \cap \ballclosure{\robotpositionmodel}{\tfrac{\sensorrange_{model}}{2}} \label{eq:local_freespace}
\end{equation}
Here, $i$ spans the obstacles in both $\knownobstaclesetdilatedmappeddisk$ (represented in $\freespacemodel$ by $\ballclosure{\diffeocenter{i}}{\diffeoradius{i}}$) and $\unknownobstaclesetdilatedmapped$ (transferred to $\freespacemodel$ with an identity map), and $\sensorrange_{model}$ is the range of the virtual sensor used for obstacle detection in $\hybridfreespacemodemodel$. Similarly to \cite{vasilopoulos_koditschek_WAFR2018}, using the diffeomorphism construction in \eqref{eq:diffeo_final}, we construct our controller as the vector field $\controlfullyactuated^\hybridmode=:\freespacemapped \rightarrow T\freespacemapped$ given by
\begin{equation}
    \controlfullyactuated^\hybridmode(\robotposition) = k \left[ D_\robotposition \diffeo \right]^{-1} \cdot \left(\controlfullyactuatedmodel^\hybridmode \circ \diffeo(\robotposition) \right) \label{eq:control_fullyactuated}
\end{equation}
with $k > 0$. Note here that the strategy employed never requires the explicit computation of $(\diffeo)^{-1}$, which would make our numerical realization quite difficult; instead, it merely requires inversion of $[D_\robotposition \diffeo]$.

We notice that if the range of the virtual sensor $\sensorrange_{model}$ used to construct $\localfreespace{\robotpositionmodel}$ in the model space is smaller than the range of our sensor $\sensorrange$, the vector field $\controlfullyactuated^\hybridmode$ is Lipschitz continuous since $\controlfullyactuatedmodel^\hybridmode(\robotpositionmodel)$ is shown to be Lipschitz continuous in \cite{arslan_kod_WAFR2016}, $\robotpositionmodel = \diffeo(\robotposition)$ is a smooth change of coordinates away from sharp corners, and the robot discovers obstacles before actually using them for navigation, because $\sensorrange_{model} < R$. We are led to the following result.

\begin{corollary}
\label{corollary:flow_mode}
    With $\hybridmode$ the terminal mode of the hybrid controller, the vector field $\controlfullyactuated^\hybridmode:\freespacemapped \rightarrow T\freespacemapped$ generates a unique continuously differentiable partial flow.
\end{corollary}

To ensure completeness (i.e., absence of finite time escape through boundaries in $\freespacemapped$) we must verify that the robot never collides with any obstacle in the environment, i.e., leaves its freespace positively invariant. However, this property follows almost directly from the fact that the vector field $\controlfullyactuated^\hybridmode$ on $\freespacemapped$ is the pushforward of the complete vector field $\controlfullyactuatedmodel^\hybridmode$ through $(\diffeo)^{-1}$, guaranteed to insure that $\hybridfreespacemodemodel$ remain positively invariant under its flow as shown in \cite{arslan_kod_WAFR2016}, away from sharp corners on the boundary of $\freespacemapped$. Therefore, we immediately get the following result.

\begin{proposition} \label{proposition:positive_invariance}
    With $\hybridmode = \knownobstaclesetindex$ the terminal mode of the hybrid controller, the freespace interior $\freespacemapped$ is positively invariant under the law \eqref{eq:control_fullyactuated}.
\end{proposition}

Next, we focus on the stationary points of $\controlfullyactuated^\hybridmode$.
\begin{lemma} \label{lemma:stationary_points}
    With $\hybridmode = \knownobstaclesetindex$ the terminal mode of the hybrid controller:
    \begin{enumerate}
        \item The set of stationary points of control law \eqref{eq:control_fullyactuated} is given as
        \begin{equation}
            \{ \goalposition \} \bigcup \{ (\diffeo)^{-1}(\mathbf{s}_i)\}_{i \in \knownobstaclesetdilatedmappeddiskindex} \bigcup \{\mathcal{G}_k\}_{k \in \unknownobstaclesetdilatedsemanticindex}
        \end{equation}
        where
        \begin{subequations} \label{eq:saddles}
            \begin{eqnarray}
                & \mathbf{s}_i = \diffeocenter{i} - \diffeoradius{i} \dfrac{\diffeo(\goalposition)-\diffeocenter{i}}{|| \diffeo(\goalposition)-\diffeocenter{i} ||} \label{eq:saddles_disks} \\
                & \mathcal{G}_k = \left\{ \mathbf{q} \in \freespacemapped \Big | d(\mathbf{q},\unknownobstacledilated_k)=\robotradius, \kappa(\mathbf{q}) = 1 \right\} \label{eq:saddles_convex}
            \end{eqnarray}
        \end{subequations}
        with
        \begin{equation*}
            \kappa(\mathbf{q}):=\dfrac{(\mathbf{q}-\projection{\overline{\unknownobstacledilated}_k}{\mathbf{q}})^\top(\mathbf{q}-\diffeo(\goalposition))}{||\mathbf{q}-\projection{\overline{\unknownobstacledilated}_k}{\mathbf{q}}|| \cdot ||\mathbf{q}-\diffeo(\goalposition)||}
        \end{equation*}
        \item The goal $\goalposition$ is the only locally stable equilibrium of control law \eqref{eq:control_fullyactuated} and all the other stationary points $\{ (\diffeo)^{-1}(\mathbf{s}_i)\}_{i \in \knownobstaclesetdilatedmappeddiskindex} \bigcup \{\mathcal{G}_k\}_{k \in \unknownobstaclesetdilatedsemanticindex}$, each associated with an obstacle, are nondegenerate saddles.
    \end{enumerate}
\end{lemma}
\begin{proof}
    Included in Appendix \ref{appendix:proofs_control}.
\end{proof}
Note that there is a slight complication here; each stationary point $\mathbf{s}_i, i \in \knownobstaclesetdilatedmappeddiskindex$ lies on the boundary of the corresponding ball $\ballclosure{\diffeocenter{i}}{\diffeoradius{i}}$ in the model space and thus, by construction of the diffeomorphism $\diffeo$, it might not lie in the domain of $(\diffeo)^{-1}$ because it could correspond to a sharp corner (i.e., a polygon vertex) in the mapped space. Although such problems can only occur for a thin subset of obstacle placements, we explicitly impose the following assumption to facilitate our formal results.

\begin{assumption}
\label{assumption:saddles}
The stationary points of control law \eqref{eq:control_fullyactuated_model} in the model space lie in the domain of the map $(\diffeo)^{-1}$ between $\freespacemodel$ and $\freespacemapped$.
\end{assumption}

Since such pathological cases can only occur for a ``thin'' (empty interior) subset of obstacle placements and the considered stationary points are shown to be nondegenerate saddles, it should be highlighted that Assumption \ref{assumption:saddles} has only theoretical and no practical implications, and does not affect the controller's performance in any way.

Then, using Lemma \ref{lemma:stationary_points}, we arrive at the following result, that establishes (almost) global convergence to the goal $\goalposition$.

\begin{proposition} \label{proposition:stability_fullyactuated}
    With $\hybridmode$ the terminal mode of the hybrid controller, the goal $\goalposition$ is an asymptotically stable equilibrium of \eqref{eq:control_fullyactuated}, whose region of attraction includes the freespace $\freespacemapped$ except a set of measure zero.
\end{proposition}
\begin{proof}
    Included in Appendix \ref{appendix:proofs_control}.
\end{proof}

We can now immediately conclude the following central summary statement.

\begin{theorem}
    \label{theorem:control_fullyactuated}
    With $\hybridmode$ the terminal mode of the hybrid controller, the reactive controller in \eqref{eq:control_fullyactuated} leaves the freespace $\freespacemapped$ positively invariant, and its unique continuously differentiable flow, starting at almost any robot placement $\robotposition \in \freespacemapped$, asymptotically reaches the goal location $\goalposition$, while strictly decreasing $||\diffeo(\robotposition) - \diffeo(\goalposition)||$ along the way.
\end{theorem}

\begin{figure}[H]
\centering
\includegraphics[width=0.5\textwidth]{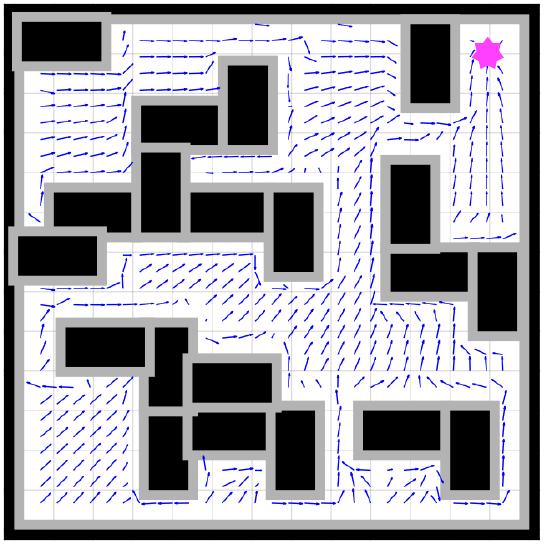}
\caption{Depiction of the vector field in \eqref{eq:control_fullyactuated} for the terminal mode $\hybridmode$ from one of our numerical examples presented in Section \ref{sec:numerical_results} with several overlapping obstacles. Notice how the vector field guarantees safety around each obstacle, with the goal in purple attracting globally.} \label{fig:vector_field}
\end{figure}

A depiction of the vector field in \eqref{eq:control_fullyactuated} for the terminal mode $\hybridmode$ (Definition \ref{definition:terminal_mode}) from one of our numerical examples presented in Section \ref{sec:numerical_results} is included in Fig. \ref{fig:vector_field}.

\subsubsection{Differential Drive Robots}
Since the robot operates in $SE(2)$ instead of $\mathbb{R}^2$, we first need to come up with a smooth diffeomorphism $\diffeounicycle:\freespacemapped \times S^1 \rightarrow \hybridfreespacemodemodel \times S^1$ away from sharp corners on the boundary of $\freespacemapped \times S^1$, and then establish the results about our controller.

Following our previous work \cite{vasilopoulos_koditschek_WAFR2018}, we construct our map $\diffeounicycle$ from $\freespacemapped \times S^1$ to $\hybridfreespacemodemodel \times S^1$ as
\begin{equation}
    \robotpositionunicyclemodel = (\robotpositionmodel,\robotorientationmodel) = \diffeounicycle(\robotpositionunicycle):=(\diffeo(\robotposition),\angletransform(\robotpositionunicycle)) \label{eq:diffeo_unicycle}
\end{equation}
with $\robotpositionunicycle=(\robotposition,\robotorientation) \in \freespacemapped \times S^1$, $\robotpositionunicyclemodel:=(\robotpositionmodel,\robotorientationmodel) \in \hybridfreespacemodemodel \times S^1$ and
\begin{equation}
    \robotorientationmodel=\angletransform(\robotpositionunicycle) := \angle(\directionvector(\robotpositionunicycle)) \label{eq:phi}
\end{equation}
Here, $\angle \directionvector:=\text{atan2}(e_2,e_1)$ and
\begin{equation}
    \directionvector(\robotpositionunicycle) = \mathrm{\Pi}_\robotpositionmodel \cdot D_{\robotpositionunicycle} \diffeounicycle \cdot \mathbf{B}(\robotorientation) \cdot \begin{bmatrix} 1 \\ 0 \end{bmatrix} = D_\robotposition \diffeo \begin{bmatrix} \cos\robotorientation \\ \sin\robotorientation \end{bmatrix}
\end{equation}
with $\mathrm{\Pi}_\robotpositionmodel$ denoting the projection onto the first two components. The reason for choosing $\robotorientationmodel$ as in \eqref{eq:phi} will become evident later, in our effort to control the equivalent differential drive robot dynamics in $\hybridfreespacemodemodel$.

\begin{proposition} \label{proposition:diffeo_unicycle}
    The map $\diffeounicycle$ in \eqref{eq:diffeo_unicycle} is a $C^\infty$ diffeomorphism from $\freespacemapped \times S^1$ to $\freespacemodel \times S^1$ away from sharp corners, none of which lie in the interior of $\freespacemapped \times S^1$.
\end{proposition}
\begin{proof}
    Included in Appendix \ref{appendix:proofs_control}.
\end{proof}

Then, using \eqref{eq:diffeo_unicycle}, we can find the pushforward of the differential drive robot dynamics in \eqref{eq:unicycle_dynamics} as
\begin{align}
    \dot{\robotpositionunicyclemodel} = & \frac{d}{dt} \begin{bmatrix} \diffeo(\robotposition) \\ \angletransform(\robotpositionunicycle) \end{bmatrix} \nonumber \\
    = & \left[ D_{\robotpositionunicycle} \diffeounicycle \circ (\diffeounicycle)^{-1} \right] \cdot \left( \mathbf{B} \circ (\diffeounicycle)^{-1}(\robotpositionunicyclemodel) \right) \cdot \controlunicycle^\hybridmode \label{eq:se2_pushforward}
\end{align}
Based on the above, we can then write
\begin{equation}
    \dot{\robotpositionunicyclemodel} = \begin{bmatrix} \dot{\robotpositionmodel} \\ \dot{\robotorientationmodel} \end{bmatrix} = \frac{d}{dt} \begin{bmatrix} \diffeo(\robotposition) \\ \angletransform(\robotpositionunicycle) \end{bmatrix} = \mathbf{B}(\robotorientationmodel) \controlunicyclemodel^\hybridmode \label{eq:unicycle_dynamics_se2}
\end{equation}
with $\controlunicyclemodel^\hybridmode = (\linearinputmodel^\hybridmode,\angularinputmodel^\hybridmode)$, and the inputs $(\linearinputmodel^\hybridmode,\angularinputmodel^\hybridmode)$ related to $(\linearinput^\hybridmode,\angularinput^\hybridmode)$ through
\begin{align}
& \linearinputmodel^\hybridmode = ||\directionvector(\robotpositionunicycle)|| \, \linearinput^\hybridmode \label{eq:reference_input_1}\\
& \angularinputmodel^\hybridmode = \linearinput^\hybridmode D_\robotposition\angletransform \begin{bmatrix}
\cos\robotorientation \\ \sin\robotorientation
\end{bmatrix} + \dfrac{\partial \angletransform}{\partial \robotorientation} \angularinput^\hybridmode \label{eq:reference_input_2}
\end{align}
with $D_\robotposition\angletransform = \begin{bmatrix}
\frac{\partial \angletransform}{\partial x} & \frac{\partial \angletransform}{\partial y}
\end{bmatrix}$. Here, we can calculate
\begin{equation}
    D_\robotposition \angletransform \begin{bmatrix} \cos \robotorientation \\ \sin \robotorientation \end{bmatrix} = \frac{\alpha_1(\robotpositionunicycle) \alpha_3(\robotpositionunicycle) + \alpha_2(\robotpositionunicycle) \alpha_4(\robotpositionunicycle)}{|| \directionvector(\robotpositionunicycle) ||^2} \label{eq:dksidx}
\end{equation}
with the auxiliary terms $\alpha_1, \alpha_2, \alpha_3, \alpha_4$ defined as
\begin{align}
    \alpha_1(\robotpositionunicycle) := & -\left([D_\robotposition\diffeo]_{21}\cos\robotorientation + [D_\robotposition\diffeo]_{22}\sin\robotorientation \right) \\
    \alpha_2(\robotpositionunicycle) := & [D_\robotposition\diffeo]_{11}\cos\robotorientation + [D_\robotposition\diffeo]_{12}\sin\robotorientation \\
    \alpha_3(\robotpositionunicycle) := & \frac{\partial [D_\robotposition\diffeo]_{11}}{\partial [\robotposition]_1} \cos^2\robotorientation + \frac{\partial [D_\robotposition\diffeo]_{12}}{\partial [\robotposition]_2} \sin^2\robotorientation \nonumber \\
    & + \left(\frac{\partial [D_\robotposition\diffeo]_{11}}{\partial [\robotposition]_2} + \frac{\partial [D_\robotposition\diffeo]_{12}}{\partial [\robotposition]_1} \right) \sin\robotorientation \cos\robotorientation \\
    \alpha_4(\robotpositionunicycle) := & \frac{\partial [D_\robotposition\diffeo]_{21}}{\partial [\robotposition]_1} \cos^2\robotorientation + \frac{\partial [D_\robotposition\diffeo]_{22}}{\partial [\robotposition]_2} \sin^2\robotorientation \nonumber \\
    & + \left(\frac{\partial [D_\robotposition\diffeo]_{21}}{\partial [\robotposition]_2} + \frac{\partial [D_\robotposition\diffeo]_{22}}{\partial [\robotposition]_1} \right) \sin\robotorientation \cos\robotorientation
\end{align}
We provide more details about the calculation of partial derivatives for elements of $D_\robotposition\diffeo$ used above in Section \ref{sec:online_algorithms} and in Appendix \ref{appendix:calculation_jacobian}.

Hence, we have found equivalent differential drive robot dynamics, defined on $\hybridfreespacemodemodel \times S^1$. The idea now is to use the control strategy in \cite{arslan_kod_WAFR2016} for the dynamical system in \eqref{eq:unicycle_dynamics_se2} to find inputs $\linearinputmodel^\hybridmode,\angularinputmodel^\hybridmode$ in $\freespacemodel \times S^1$, and then use \eqref{eq:reference_input_1}, \eqref{eq:reference_input_2} to find the actual inputs $\linearinput^\hybridmode,\angularinput^\hybridmode$ in $\freespacemapped \times S^1$ that achieve $\linearinputmodel^\hybridmode,\angularinputmodel^\hybridmode$ as
\begin{subequations} \label{eq:control_unicycle}
\begin{eqnarray}
& \linearinput^\hybridmode =\dfrac{k_v \, \linearinputmodel^\hybridmode}{||\directionvector(\robotpositionunicycle)||} \\
& \angularinput^\hybridmode = \left(\dfrac{\partial \angletransform}{\partial \robotorientation}\right)^{-1} \left(k_\omega \, \angularinputmodel^\hybridmode-\linearinput^\hybridmode D_\robotposition\angletransform \begin{bmatrix}
\cos\robotorientation \\ \sin\robotorientation
\end{bmatrix} \right)
\end{eqnarray}
\end{subequations}
with $k_v,k_\omega>0$ fixed gains.

Namely, inspired by \cite{arslan_kod_WAFR2016,astolfi_1999}, we design our inputs $\linearinputmodel^\hybridmode$ and $\angularinputmodel^\hybridmode$ as\footnote{In \eqref{eq:diffeo_unicycle}, we construct a diffeomorphism $\diffeounicycle$ between $\freespacemapped \times S^1$ and $\hybridfreespacemodemodel \times S^1$. However, for practical purposes, we deal only with one specific chart of $S^1$ in our control structure, described by the angles $(-\pi,\pi]$. As shown in \cite{astolfi_1999}, the discontinuity at $\pm \pi$ does not induce a discontinuity in our controller due to the use of the $\text{atan}$ function in \eqref{eq:reference_control_omega}. On the contrary, with the use of \eqref{eq:reference_control_omega} as in \cite{astolfi_1999,arslan_kod_WAFR2016}, the robot never changes heading in $\hybridfreespacemodemodel$, which implies that the generated trajectories both in $\hybridfreespacemodemodel$ and (by the properties of the diffeomorphism $\diffeounicycle$) in $\freespacemapped$ have no cusps, even though the robot might change heading in $\freespacemapped$ because of the more complicated nature of the function $\angletransform$ in \eqref{eq:phi}.}
\begin{subequations} \label{eq:control_unicycle_reference}
\begin{eqnarray}
& \linearinputmodel^\hybridmode = - \begin{bmatrix}
\cos\robotorientationmodel \\ \sin\robotorientationmodel
\end{bmatrix}^\top\left(\robotpositionmodel-\localgoallinear \right) \label{eq:reference_control_v} \\
& \angularinputmodel^\hybridmode = \text{atan} \left( \dfrac{\begin{bmatrix}
-\sin\robotorientationmodel \\ \cos\robotorientationmodel
\end{bmatrix}^\top \left( \robotpositionmodel - \localgoalangular \right)}{\begin{bmatrix}
\cos\robotorientationmodel \\ \sin\robotorientationmodel
\end{bmatrix}^\top \left( \robotpositionmodel - \localgoalangular \right)} \right) \label{eq:reference_control_omega}
\end{eqnarray}
\end{subequations}
with $\localfreespace{\robotpositionmodel} \subset \hybridfreespacemodemodel$ the convex polygon defining the local freespace at $\robotpositionmodel = \diffeo(\robotposition)$, and linear and angular local goals $\localgoallinear, \localgoalangular$ given by
\begin{align}
    \localgoallinear & := \projection{\localfreespace{\robotpositionmodel} \cap H_\parallel}{\goalpositionmodel} \label{eq:linearlocalgoal} \\
    \localgoalangular & := \dfrac{\projection{\localfreespace{\robotpositionmodel} \cap H_G}{\goalpositionmodel}+\projection{\localfreespace{\robotpositionmodel}}{\goalpositionmodel}}{2} \label{eq:angularlocalgoal}
\end{align}
with $H_\parallel$ and $H_G$ the lines defined in \cite{arslan_kod_WAFR2016} as 
\begin{align}
H_\parallel = & \left\{ \mathbf{z} \in \hybridfreespacemodemodel \, \Big | \, \begin{bmatrix} -\sin\robotorientationmodel \\ \cos\robotorientationmodel \end{bmatrix}^\top (\mathbf{z}-\robotpositionmodel) = 0 \right\} \\
H_G = & \left\{ \alpha \robotpositionmodel + (1-\alpha) \goalpositionmodel \in \hybridfreespacemodemodel \, | \, \alpha \in \mathbb{R} \right\}
\end{align}

The properties of the differential drive robot control law given in \eqref{eq:control_unicycle} can be summarized in the following theorem.

\begin{theorem} \label{theorem:control_se2}
With $\hybridmode$ the terminal mode of the hybrid controller, the reactive controller for differential drive robots, given in \eqref{eq:control_unicycle}, leaves the freespace $\freespacemapped \times S^1$ positively invariant, and its unique continuously differentiable flow, starting at almost any robot configuration $(\robotposition, \robotorientation) \in \freespacemapped \times S^1$, asymptotically steers the robot to the goal location $\goalposition$, without increasing $|| \diffeo(\robotposition)-\diffeo(\goalposition)||$ along the way.
\end{theorem}
\begin{proof}
    Included in Appendix \ref{appendix:proofs_control}.
\end{proof}

\subsection{Qualitative Properties of the Hybrid Controller}
\label{subsec:hybrid_qualitative}

\subsubsection{Fully Actuated Robots}
First, we show that the navigational hybrid system $\hybridsystem$ inherits the fundamental consistency properties outlined in \cite[Theorems 5-9]{Johnson_Burden_Koditschek_2016}, in order to establish that the hybrid system is well-behaved in the sense of being both deterministic and non-blocking (i.e., generating executions defined for all future times).

\begin{lemma}
    \label{lemma:disjoint_guards}
    The hybrid system $\hybridsystem$ has disjoint guards.
\end{lemma}
\begin{proof}
    Included in Appendix \ref{appendix:proofs_control}.
\end{proof}

An immediate result following Lemma \ref{lemma:disjoint_guards}, that does not allow a robot state $\robotposition \in \hybridfreespacemode$ to be contained in more than one guard $\hybridguardrestriction^{\hybridmode,\hybridmode'}$, and the nice properties of the flow in each separate mode, summarized in Corollary \ref{corollary:flow_mode}, is the following important consistency property.
\begin{corollary}
    \label{corollary:determinism}
    The hybrid system $\hybridsystem$ is deterministic.
\end{corollary}

Next, we focus on the non-blocking property. As stated in \cite{Johnson_Burden_Koditschek_2016}, a hybrid execution might be blocked either by conventional finite escape through the boundary of the hybrid domain at a point in the complement of all the guards, by escape through a point in the guard whose reset lies outside of the hybrid domain, or by hybrid ambiguity, i.e., by arriving at a point through the continuous flow that lies in the complement of the guard $\hybridguard$ and yet still on the boundary of $\hybridguard$. We eliminate all cases in the proof of the following result.
\begin{lemma}
    \label{lemma:non_blocking}
    The hybrid system $\hybridsystem$ is non-blocking.
\end{lemma}
\begin{proof}
    Included in Appendix \ref{appendix:proofs_control}.
\end{proof}

Finally, using the last part of the proof of Lemma \ref{lemma:non_blocking} which shows that the (identity) reset from a given mode cannot lie in the guard of the next mode, we arrive at the following result about the discrete transitions of the hybrid system $\hybridsystem$.
\begin{corollary}
    \label{corollary:discrete_transitions}
    An execution of the hybrid system $\hybridsystem$ undergoes no more than one hybrid transition at a single time $t$.
\end{corollary}

Based on the above, the central result about the hybrid controller for a fully actuated robot can be summarized in the following Theorem.

\begin{theorem}
\label{theorem:hybrid_fullyactuated}
    For a fully actuated robot with dynamics defined in \eqref{eq:dynamics_fullyactuated}, the deterministic, non-blocking navigational hybrid system $\hybridsystem := \left(2^{\knownobstaclesetindex}, \hybridgraph, \hybriddomain, \hybridfield, \hybridguard, \hybridreset \right)$, with the restrictions of guards $\hybridguard$, resets $\hybridreset$ and vector fields $\hybridfield$ defined as in \eqref{eq:restriction_guard}, \eqref{eq:restriction_reset} and \eqref{eq:control_fullyactuated_hybrid} respectively, leaves the free space $\freespace$ positively invariant under the Lipschitz continuous, piecewise smooth flow associated with each of its hybrid domains, and, starting at almost any robot placement $\robotposition \in \freespace$ at time $t_0$ with an initial mode $\hybridmode = \varnothing$, asymptotically reaches a designated goal location $\goalposition \in \freespace$, in a previously unexplored environment satisfying Assumptions \ref{assumption:convex} - \ref{assumption:saddles}, with a uniquely defined (in both state and mode) execution for all $t > t_0$.
\end{theorem}
\begin{proof}
    Included in Appendix \ref{appendix:proofs_control}.
\end{proof}

\subsubsection{Differential Drive Robots}
We can then follow exactly the same procedure to prove the following statement for the hybrid controller for differential drive robots.
\begin{theorem}
\label{theorem:hybrid_unicycle}
    For a differential drive robot with dynamics defined in \eqref{eq:unicycle_dynamics}, the deterministic, non-blocking navigational hybrid system $\hybridsystemunicycle := \left(2^{\knownobstaclesetindex}, \hybridgraph, \hybriddomainunicycle, \hybridfieldunicycle, \hybridguardunicycle, \hybridresetunicycle \right)$, with the restrictions of guards $\hybridguardunicycle$, resets $\hybridresetunicycle$ and vector fields $\hybridfieldunicycle$ defined as in \eqref{eq:restriction_guard_unicycle}, \eqref{eq:restriction_reset_unicycle} and \eqref{eq:control_unicycle_hybrid} respectively, leaves the free space $\freespace \times S^1$ positively invariant under the Lipschitz continuous, piecewise smooth flow associated with each of its hybrid domains, and, starting at almost any robot placement $\robotpositionunicycle \in \freespace \times S^1$ at time $t_0$ with an initial mode $\hybridmode = \varnothing$, asymptotically reaches a designated goal location $\goalposition \in \freespace$, in a previously unexplored environment satisfying Assumptions \ref{assumption:convex} - \ref{assumption:saddles}, with a uniquely defined (in both state and mode) execution for all $t > t_0$.
\end{theorem}

\subsection{Generating Bounded Inputs}
\label{subsec:bounded_inputs}
Although the control inputs for both a fully actuated robot and a differential drive robot, described in \eqref{eq:control_fullyactuated} and \eqref{eq:control_unicycle} respectively, can be used in the hybrid systems description of the controller (see \eqref{eq:control_fullyactuated_hybrid} and \eqref{eq:control_unicycle_hybrid}) to yield the desired results of Theorems \ref{theorem:hybrid_fullyactuated} and \ref{theorem:hybrid_unicycle}, we have so far implicitly assumed that there is no bound in the magnitude of $\controlfullyactuated^\hybridmode$ in \eqref{eq:control_fullyactuated} or the magnitudes of $\linearinput^\hybridmode,\angularinput^\hybridmode$ in \eqref{eq:control_unicycle} for each separate mode $\hybridmode$. In this Section, we show how to generate bounded inputs without affecting the results of Theorems \ref{theorem:hybrid_fullyactuated} and \ref{theorem:hybrid_unicycle}.

\subsubsection{Fully Actuated Robots}
We focus on fully actuated robots first. Let $\controlfullyactuated_{nom}^\hybridmode : \hybridfreespacemodemapped \rightarrow T \hybridfreespacemodemapped$ denote the nominal input for mode $\hybridmode$, defined using \eqref{eq:control_fullyactuated} as
\begin{equation}
    \controlfullyactuated_{nom}^\hybridmode(\robotposition) := \left[ D_\robotposition \diffeo \right]^{-1} \cdot \left(\controlfullyactuatedmodel^\hybridmode \circ \diffeo(\robotposition) \right) \label{eq:control_fullyactuated_nominal}
\end{equation}

We can then easily satisfy the requirement $||\controlfullyactuated^\hybridmode|| \leq u_{\max}$ by picking a gain $k$ such that $ 0 < k \leq u_{\max}$ and defining our controller as
\begin{equation}
    \controlfullyactuated^\hybridmode(\robotposition) := k \frac{\controlfullyactuated_{nom}^\hybridmode(\robotposition)}{||\controlfullyactuated_{nom}^\hybridmode(\robotposition)|| + \epsilon_\controlfullyactuated} \label{eq:control_fullyactuated_bounded}
\end{equation}
with $\epsilon_\controlfullyactuated > 0$ a small number. The modified (bounded) controller in \eqref{eq:control_fullyactuated_bounded} does not affect the results of Theorem \ref{theorem:hybrid_fullyactuated}, since it maintains the heading direction of the original (unbounded) controller in \eqref{eq:control_fullyactuated}, and just limits its magnitude.

\subsubsection{Differential Drive Robots}
The analysis is slightly more complicated for differential drive robots, since we have to respect the fact that the actual inputs $\linearinput^\hybridmode$, $\angularinput^\hybridmode$ are related to the inputs $\linearinputmodel^\hybridmode$, $\angularinputmodel^\hybridmode$ through \eqref{eq:control_unicycle}. However, an important observation, deriving from the proof of Theorem \ref{theorem:control_se2}, is that the choice of gains $k_v, k_\omega>0$ in \eqref{eq:control_unicycle} does not affect the positive invariance or convergence properties of the controller, which rely entirely on $\linearinputmodel^\hybridmode$, $\angularinputmodel^\hybridmode$, given in \eqref{eq:control_unicycle_reference}.

Therefore, the main idea is to adaptively change the gains online, in order to satisfy the constraints $|\linearinput^\hybridmode| \leq \linearinput_{\max}$, $|\angularinput^\hybridmode| \leq \angularinput_{\max}$. Namely, using \eqref{eq:control_unicycle}, we look for gains $k_v(\robotpositionunicycle), k_\omega(\robotpositionunicycle)$ such that
\begin{subequations}
\begin{eqnarray}
& k_v(\robotpositionunicycle) \dfrac{|\linearinputmodel^\hybridmode(\robotpositionunicycle)|}{||\directionvector(\robotpositionunicycle)||} \leq \linearinput_{\max} \nonumber\\
& \left|k_\omega(\robotpositionunicycle) \, \angularinputmodel^\hybridmode(\robotpositionunicycle) - k_v(\robotpositionunicycle)\dfrac{\linearinputmodel^\hybridmode(\robotpositionunicycle)}{||\directionvector(\robotpositionunicycle)||} \vartheta(\robotorientation) \right| \leq \dfrac{\partial \angletransform}{\partial \robotorientation}\angularinput_{\max} \nonumber
\end{eqnarray}
\end{subequations}
with $\vartheta(\robotorientation) : = D_\robotposition\angletransform \begin{bmatrix}
\cos\robotorientation & \sin\robotorientation
\end{bmatrix}^\top$, since $\frac{\partial \angletransform}{\partial \robotorientation} > 0$, as shown in the proof of Proposition \ref{proposition:diffeo_unicycle}. A conservative selection of gains that satisfies the above constraints can then be extracted using the triangle inequality as follows
\begin{align}
k_v(\robotpositionunicycle) = & \min \left( k_{v,nom}, \dfrac{ ||\directionvector(\robotpositionunicycle)||}{|\linearinputmodel^\hybridmode(\robotpositionunicycle)|} v_{\max}, \right. \nonumber \\
& \left. \lambda \dfrac{\partial \angletransform}{\partial \robotorientation} \dfrac{ ||\directionvector(\robotpositionunicycle)||}{|\linearinputmodel^\hybridmode(\robotpositionunicycle)| \, |\vartheta(\robotorientation)|}\omega_{\max} \right) \label{eq:bounded_gain_v} \\
k_\omega(\robotpositionunicycle) = & \min \left(k_{\omega,nom}, (1-\lambda) \dfrac{\partial \angletransform}{\partial \robotorientation} \dfrac{\omega_{\max}}{|\angularinputmodel^\hybridmode(\robotpositionunicycle)|} \right) \label{eq:bounded_gain_omega}
\end{align}
with $k_{v,nom}, k_{\omega,nom} >0$ initially provided nominal gains and $\lambda \in (0,1)$ a tuning parameter. It can be seen that $k_v(\robotpositionunicycle), k_\omega(\robotpositionunicycle)$ are always positive since $||\directionvector(\robotpositionunicycle)||, \frac{\partial \angletransform}{\partial \robotorientation}$ are always positive.

\section{Online Reactive Planning Algorithms}
\label{sec:online_algorithms}

With the description of the diffeomorphism construction and the overall hybrid controller, we are now ready to describe the algorithm we use during execution time to generate our control inputs. As shown in Fig. \ref{fig:algorithm} that summarizes the whole architecture, we divide the main algorithm that communicates with the semantic mapping and the perception pipelines\footnote{For our numerical studies, we simply assume an idealized sensor of fixed range that can instantly recognize and localize obstacles within its range. For our hardware implementation, the semantic mapping and perception pipelines rely mostly on prior work and are briefly described in Section \ref{sec:experimental_setup}.} in two distinct components. First, the {\it mapped space recovery} component, described in Section \ref{subsec:mapped_space_recovery}, is responsible for keeping track of all encountered objects, and extracting the sets of obstacles $\knownobstaclesetdilatedmappeddisk$, $\knownobstaclesetdilatedmappedintrusion$. Next, the {\it reactive planning} component, described in Section \ref{subsec:reactive_planning}, uses the input from the mapped space recovery component to generate the diffeomorphism $\diffeo$ (described in Section \ref{sec:diffeomorphism}) between the mapped space and the model space during execution time, and provide the commands for the robot according to the hybrid controller (described in Section \ref{sec:controller}).

\begin{algorithm}
\begin{algorithmic}
\Function{MappedSpaceRecovery}{$\knownobstaclesetphysical$}
\State $\knownobstaclesetdilatedmapped \gets \texttt{Union}(\texttt{dilate}(\knownobstaclesetphysical,\robotradius))$
\Do
\State $\knownobstacledilated \gets \texttt{pop}(\knownobstaclesetdilatedmapped)$ \Comment{Pop next component}
\If{$\knownobstacledilated \cap \partial \enclosingfreespace \neq \varnothing$}
\State $\knownobstacledilatedmappedintrusion.\texttt{geometry} \gets \knownobstacledilated \cap \enclosingfreespace$
\State $\knownobstacledilatedmappedintrusion.\texttt{tree} \gets \texttt{EarClipping}(\knownobstacledilatedmappedintrusion.\texttt{geometry})$
\State Find root of tree $\knownobstacledilatedmappedintrusion.\texttt{root}$ as in Section \ref{subsec:obstacle_representation}
\State Restructure $\knownobstacledilatedmappedintrusion.\texttt{tree}$ around $\knownobstacledilatedmappedintrusion.\texttt{root}$
\For{$j \in \knownobstacledilatedmappedintrusion.\texttt{tree}.\texttt{vertices}$} \Comment{Dfns. \ref{definition:center}-\ref{definition:collars_root_boundary}}
\State $\knownobstacledilatedmappedintrusion.\texttt{tree}.\texttt{vertices}(j).\texttt{append}(\diffeocenter{j})$
\State $\knownobstacledilatedmappedintrusion.\texttt{tree}.\texttt{vertices}(j).\texttt{append}(\outerpolygon{j})$
\EndFor
\State $\knownobstaclesetdilatedmappedintrusion.\texttt{append}(\knownobstacledilatedmappedintrusion)$
\Else
\State $\knownobstacledilatedmappeddisk.\texttt{geometry} \gets \knownobstacledilated$
\State $\knownobstacledilatedmappeddisk.\texttt{tree} \gets \texttt{EarClipping}(\knownobstacledilatedmappeddisk.\texttt{geometry})$
\State Find root of tree $\knownobstacledilatedmappeddisk.\texttt{root}$ as in Section \ref{subsec:obstacle_representation}
\State Restructure $\knownobstacledilatedmappeddisk.\texttt{tree}$ around $\knownobstacledilatedmappeddisk.\texttt{root}$
\For{$j \in \knownobstacledilatedmappeddisk.\texttt{tree}.\texttt{vertices}$} \Comment{Dfns. \ref{definition:center}-\ref{definition:collars_root_boundary}}
\State $\knownobstacledilatedmappeddisk.\texttt{tree}.\texttt{vertices}(j).\texttt{append}(\diffeocenter{j})$
\State $\knownobstacledilatedmappeddisk.\texttt{tree}.\texttt{vertices}(j).\texttt{append}(\outerpolygon{j})$
\EndFor
\State Find $\diffeoradius{} = \knownobstacledilatedmappeddisk.\texttt{radius}$ as in Section \ref{subsec:root_purging}
\State $\knownobstaclesetdilatedmappeddisk.\texttt{append}(\knownobstacledilatedmappeddisk)$
\EndIf
\doWhile{$\knownobstaclesetdilatedmapped \neq \varnothing$}
\State $\textbf{return} \quad \knownobstaclesetdilatedmappeddisk, \knownobstaclesetdilatedmappedintrusion$
\EndFunction
\end{algorithmic}
\caption{Derivation of the sets of obstacles $\knownobstaclesetdilatedmappeddisk, \knownobstaclesetdilatedmappedintrusion$, used in the diffeomorphism construction, and their associated properties, from the aggregated list of known obstacles in the physical space $\knownobstaclesetphysical$.} \label{algorithm:mapped_space_recovery}
\end{algorithm}

\subsection{Mapped Space Recovery}
\label{subsec:mapped_space_recovery}
Given as input the aggregated set of localized, recognized familiar obstacles $\knownobstaclesetphysical$, we first dilate all these elements of $\knownobstaclesetphysical$ by the robot radius $\robotradius$, to form the components of $\knownobstaclesetdilatedsemantic$, and consolidate the connected components resulting from their union into a new set of merged obstacles to form $\knownobstaclesetdilatedmapped$. Then, for each connected component $\knownobstacledilated$ of $\knownobstaclesetdilatedmapped$ that intersects the boundary of the enclosing freespace $\enclosingfreespace$, we take $\knownobstacledilatedmappedintrusion = \knownobstacledilated \cap \enclosingfreespace$, as described in Section \ref{subsec:mapped_space}, and include $\knownobstacledilatedmappedintrusion$ in the list of obstacles to be merged into $\partial \enclosingfreespace$, $\knownobstaclesetdilatedmappedintrusion$; the rest of the components of $\knownobstaclesetdilatedmapped$ are included in the list of obstacles to be deformed into disks, $\knownobstaclesetdilatedmappeddisk$\footnote{Note in consequence of these consolidations that the cardinality of the index $\hybridmode$ denoting the subset of familiar objects discovered and localized in the semantic space, $\knownobstaclesetdilatedsemantic$, will in general be larger than the cardinality of connected components in $\knownobstaclesetdilatedmapped$, whose cardinality in turn will generally be larger than that of the connected components in $\knownobstaclesetdilatedmappeddisk$. Nevertheless, under the assumption of fixed obstacles, these cardinalities are also fixed functions of $\hybridmode$, constant over some fixed subset of robot placements, hence these subsets of semantically identified and localized familiar obstacles, $\hybridmode \subseteq \knownobstaclesetindex$, comprise the appropriate indices for the modes (i.e., they label the vertices of the graph $\hybridgraph$) of the hybrid system just analyzed in Section \ref{subsec:control_hybrid}. This situation is illustrated in Section \ref{fig:illustration_merging}.}. All these computational steps rely on underlying polygon operations (unions, intersections, differences); the development of such algorithms has been heavily explored in the computational geometry literature \cite{egenhofer_1991,clementini_1993,douglas_1973}, and here we rely on their efficient implementations, either in the open-source C++ Boost library \cite{boost}, or in the open-source Shapely package \cite{shapely} in Python.

The next step is to triangulate every obstacle $\knownobstacledilated_i$ in both $\knownobstaclesetdilatedmappeddisk$ and $\knownobstaclesetdilatedmappedintrusion$ using the Ear Clipping Method, find its root triangle $r_i$ and extract the corresponding tree of triangles $\triangletree{\knownobstacledilated_i} := (\trianglevertices{\knownobstacledilated_i}, \triangleedges{\knownobstacledilated_i})$, as described in Section \ref{subsec:obstacle_representation}. For the implementation of the Ear Clipping Method, we use either the open-source Boost library \cite{boost}, for our C++ implementation, or the open-source \texttt{tripy} package \cite{tripy}, for our Python implementation. 

The final operation of the mapped space recovery algorithm is to extract the admissible centers of transformation, $\diffeocenter{j}$, according to Definitions \ref{definition:center} - \ref{definition:collars_root_boundary}, the corresponding radius of transformation, $\diffeoradius{i}$ (if $\knownobstacledilated_i \in \knownobstaclesetdilatedmappeddisk$), and the admissible polygonal collars, $\outerpolygon{j}$ for all triangles $j \in \trianglevertices{\knownobstacledilated_i}$ and polygons $\knownobstacledilated_i$ in $\knownobstaclesetdilatedmappeddisk$ and $\knownobstaclesetdilatedmappedintrusion$. There is not a unique method of performing this operation, and we provide our implemented method along with other details in Appendix \ref{appendix:computational_geometry}. The mapped space recovery algorithm is summarized in Algorithm \ref{algorithm:mapped_space_recovery}.

\subsection{Reactive Planning Component}
\label{subsec:reactive_planning}
The mapped space recovery algorithm described above just informs the robot about its surroundings, by post-processing aggregated information from the semantic mapping pipeline. In this Section, we describe the algorithm for generating actual robot inputs, that closes our control loop.

\begin{algorithm}[H]
\begin{algorithmic}
\Function{ReactivePlanning}{$\texttt{State}$, $\texttt{LIDAR}$, $\knownobstaclesetdilatedmappeddisk, \knownobstaclesetdilatedmappedintrusion$}
\If{$\texttt{RobotType}$ is $\texttt{FullyActuated}$}
\State $\robotposition \gets \texttt{State}$
\State $\robotpositionmodel \gets \diffeo(\robotposition)$ \Comment{Sec. \ref{sec:diffeomorphism}, Appendix \ref{appendix:calculation_jacobian}}
\State Compute $D_\robotposition \diffeo$ \Comment{Appendix \ref{appendix:calculation_jacobian}}
\State Populate $\freespacemodel$ using $\texttt{LIDAR},\knownobstaclesetdilatedmappeddisk, \knownobstaclesetdilatedmappedintrusion$
\State Construct $\localfreespace{\robotpositionmodel}$ \Comment{\eqref{eq:local_freespace}}
\State Compute input $\controlfullyactuatedmodel^\hybridmode$ \Comment{\eqref{eq:control_fullyactuated_model}}
\State Compute input $\controlfullyactuated^\hybridmode$ \Comment{\eqref{eq:control_fullyactuated_bounded}}
\State $\texttt{RobotInput} \gets \controlfullyactuated^\hybridmode$
\ElsIf{$\texttt{RobotType}$ is $\texttt{DiffDrive}$}
\State $\robotpositionunicycle \gets \texttt{State}$
\State $\robotpositionunicyclemodel \gets \diffeounicycle(\robotpositionunicycle)$ \Comment{Sec. \ref{subsec:control_modes}, \eqref{eq:diffeo_unicycle}, Appendix \ref{appendix:calculation_jacobian}}
\State Compute $D_\robotposition \diffeo, \frac{\partial [D_\robotposition \diffeo]_{ij}}{\partial [\robotposition]_k}$ \Comment{Appendix \ref{appendix:calculation_jacobian}}
\State Populate $\freespacemodel$ using $\texttt{LIDAR},\knownobstaclesetdilatedmappeddisk, \knownobstaclesetdilatedmappedintrusion$
\State Construct $\localfreespace{\robotpositionmodel}$ \Comment{\eqref{eq:local_freespace}}
\State Compute inputs $\controlunicyclemodel^\hybridmode$ \Comment{\eqref{eq:control_unicycle_reference}}
\State Compute input $\controlunicycle^\hybridmode$ \Comment{\eqref{eq:control_unicycle} using \eqref{eq:bounded_gain_v},\eqref{eq:bounded_gain_omega}}
\State $\texttt{RobotInput} \gets \controlunicycle^\hybridmode$
\EndIf
\State $\textbf{return} \quad \texttt{RobotInput}$
\EndFunction
\end{algorithmic}
\caption{Description of the online reactive planning module that uses the state of the robot, LIDAR input, and $\knownobstaclesetdilatedmappeddisk, \knownobstaclesetdilatedmappedintrusion$.} \label{algorithm:reactive_planning}
\end{algorithm}

Given the robot state in the mapped space, ($\robotposition$ for a fully actuated robot or $\robotpositionunicycle$ for a differential drive robot), and the list of obstacles in the mapped space, $\knownobstaclesetdilatedmappeddisk, \knownobstaclesetdilatedmappedintrusion$, along with their associated triangulation trees and their properties as computed with Algorithm \ref{algorithm:mapped_space_recovery}, the first step of the reactive planning algorithm is to compute the state of the robot in the model space ($\diffeo(\robotposition)$ for a fully actuated robot \eqref{eq:diffeo_final} or $\diffeounicycle(\robotpositionunicycle)$ for a differential drive robot \eqref{eq:diffeo_unicycle}), the diffeomorphism jacobian $D_\robotposition \diffeo$, and partial derivatives of the terms of the jacobian $\frac{\partial [D_\robotposition \diffeo]_{ij}}{\partial [\robotposition]_k}$ (needed in \eqref{eq:control_unicycle} for a differential drive robot), following the methods outlined in Section \ref{sec:diffeomorphism}. We show in Appendix \ref{appendix:calculation_jacobian} how to perform this operation inductively, given the general form of $\diffeo$ in \eqref{eq:diffeo_final}.

Next, we need to properly populate the model space with obstacles, in order to compute the input \eqref{eq:control_fullyactuated_bounded} for a fully actuated robot, or the inputs \eqref{eq:control_unicycle} (using \eqref{eq:bounded_gain_v},\eqref{eq:bounded_gain_omega}) for a differential drive robot. This procedure is straightforward for familiar obstacles; obstacles in $\knownobstaclesetdilatedmappedintrusion$ are not taken into account in the model space, since they are merged into the boundary $\partial \enclosingfreespace$, and obstacles in $\knownobstaclesetdilatedmappeddisk$ are represented in the model space as disks with radius $\diffeoradius{i}$ centered at $\diffeocenter{i}$, with $i$ spanning the elements of $\knownobstaclesetdilatedmappeddisk$. For unknown obstacles in $\unknownobstaclesetdilatedmapped$, we use the LIDAR measurements (see Fig. \ref{fig:algorithm}). Namely, we first pre-process the 2D LIDAR pointcloud by disregarding points that correspond to obstacles in $\knownobstaclesetdilatedmappeddisk$ or $\knownobstaclesetdilatedmappedintrusion$, since those have already been considered. The pointcloud with the remaining points is then transferred with an identity transform to the model space; this is allowed because, by construction, the diffeomorphism $\diffeo$ between $\freespacemapped$ and $\freespacemodel$ defaults to the identity transform (i.e., $\diffeo(\robotposition) = \robotposition$) on the boundary of any unknown obstacle, provided that this obstacle is sufficienty separated from any obstacle in $\knownobstaclesetdilatedmapped$ (see Assumption \ref{assumption:beta}).

With the (``virtual'') model space constructed, we can then construct the local freespace \eqref{eq:local_freespace}, as in \cite[Eqn. (24)]{arslan_kod_WAFR2016}, and, subsequently, compute the input $\controlfullyactuatedmodel^\hybridmode$ \eqref{eq:control_fullyactuated_model} for a fully actuated robot or the inputs $\controlunicyclemodel^\hybridmode$ \eqref{eq:control_unicycle_reference} for a differential drive robot. The final step is to compute the ``pull-backs'' of these inputs in the physical space and enforce bounds, by using \eqref{eq:control_fullyactuated_bounded} for a fully actuated robot, or \eqref{eq:control_unicycle} along with \eqref{eq:bounded_gain_v}, \eqref{eq:bounded_gain_omega} to adaptively modify the input gains for a differential drive robot. The reactive planning module functionality is summarized in Algorithm \ref{algorithm:reactive_planning}.

It must be highlighted that the presented reactive planning pipeline (summarized in Fig. \ref{fig:algorithm}) runs at 10Hz online and onboard our physical robots' Nvidia Jetson TX2 modules, during execution time.
\section{Numerical Results}
\label{sec:numerical_results}
In this Section, we present numerical simulations that illustrate our formal results. Our simulations are run in MATLAB using \texttt{ode45}, and $p = 20$ for the R-function construction, as described in Appendix \ref{appendix:implicit}. Our mapped space recovery (Section \ref{subsec:mapped_space_recovery}) and reactive planning (Section \ref{subsec:reactive_planning}) algorithms are implemented in Python and communicate with MATLAB using the standard MATLAB-Python interface. For our numerical results, we assume perfect robot state estimation and localization of obstacles, using a fixed range sensor that can instantly identify and localize either the entirety of familiar obstacles that intersect its footprint, or the corresponding fragments of unknown obstacles within its range.

\subsection{Comparison with Original Doubly Reactive Algorithm}
\label{subsec:comparison_numerical}

\begin{figure}[H]
\centering
\includegraphics[width=0.4\textwidth]{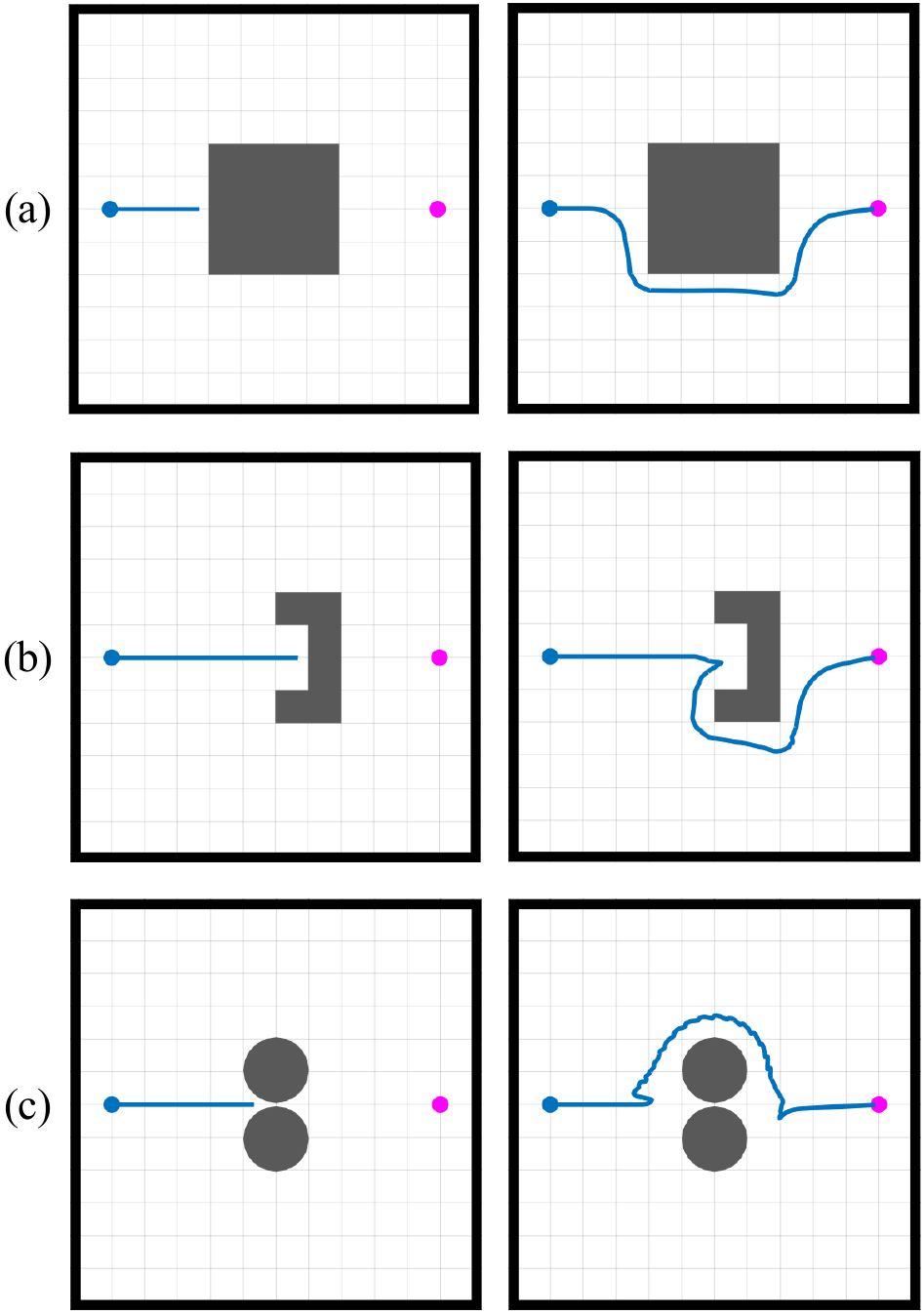}
\caption{Comparison with original doubly reactive algorithm for a fully actuated robot (blue) navigating towards a goal (purple). (a) Convex obstacle with flat surfaces, (b) Non-convex obstacle, (c) Convex obstacles violating the separation assumptions of \cite{arslan_kod_WAFR2016}. Left column: Original doubly reactive algorithm \cite{arslan_kod_WAFR2016}, Right column: Our algorithm.} \label{fig:comparison}
\vspace{-8pt}
\end{figure}

We begin with a comparison of our algorithm performance with the original version of the doubly reactive algorithm in \cite{arslan_kod_WAFR2016}, that we use in the model space computed at each instant from the perceptual inputs as depicted in Fig. \ref{fig:algorithm}-(e) and described in Section \ref{subsec:reactive_planning}. Fig. \ref{fig:comparison} demonstrates the well understood limitations of this algorithm (limitations of all online \cite{borenstein_koren_TRA1991} or offline \cite{filippidis_kyriakopoulos_2012} reactive schemes we are aware of). Namely, in the presence of a flat surface or a non-convex obstacle, or when separation assumptions are violated, the robot gets stuck in undesired local minima, which are locally stable and trap a set of initial conditions whose area becomes arbitrarily large as their ``shadows'' (i.e., the corresponding basins of attraction) grow -- see, e.g., \cite[Fig. 6-(b)]{vasilopoulos_pavlakos_bowman_caporale_daniilidis_pappas_koditschek_2020}. Absent the new methods introduced in this paper, handling these spurious basins of attraction would require complex dynamic replanning algorithms \cite{Reverdy_Ilhan_Koditschek_2015,Revzen_Ilhan_Koditschek_2012}, whose presentation falls beyond the scope of the present paper. In contrast, our algorithm overcomes this limitation, by recourse to the robot's ability to recognize obstacles at hand (documented empirically in Section \ref{sec:experiments}) and transform them appropriately (as detailed in Section \ref{sec:diffeomorphism}) for both a fully actuated and a differential drive robot. The robot radius used in our simulation studies is 0.2m, the control gains are $k = k_v = k_\omega = 0.4$, and the values of $\innerpolygontune{j_i} = \innerpolygontune{r_i}$, $\outerpolygontune{j_i} = \outerpolygontune{r_i}$ and $\innerpolygondistance{j_i} = \innerpolygondistance{r_i}$ used in the diffeomorphism construction are 4.0, 0.05 and 2.0 respectively. Finally, the maximum input $u_{\max}$ for the fully actuated robot as well as the maximum linear and angular inputs $v_{\max},\omega_{\max}$ for the differential drive robot are limited to 0.4.

\subsection{Navigation in a Cluttered Environment with Obstacle Merging}

\begin{figure}[H]
\centering
\includegraphics[width=1.0\textwidth]{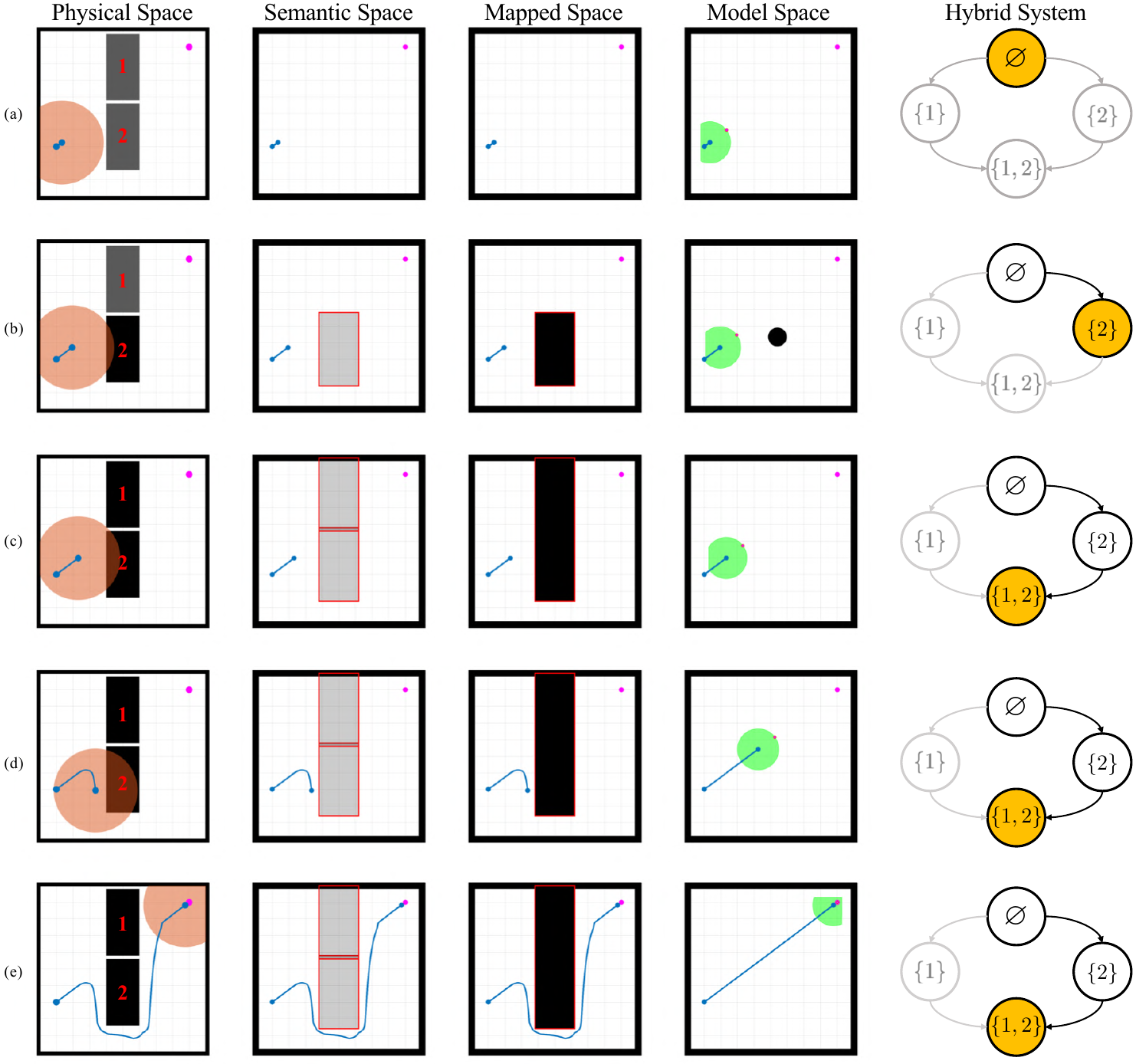}
\caption{Illustration of the algorithm with successive snapshots of a single simulation run in the presence of two familiar obstacles with \`a-priori unknown pose. (a) The robot starts navigating towards the goal with no prior information about its environment. The initial mode of the hybrid controller is $\hybridmode = \varnothing$. (b) The robot discovers the first familiar obstacle (labeled 2 as shown in the physical space), driving the hybrid dynamical system (Section \ref{sec:controller}) into mode $\hybridmode = \{2\}$, wherein it makes an (incorrect) hypothesis about the topological state of the workspace (shown in the mapped space). The robot now computes according to \cite{arslan_kod_WAFR2016} the model control input in the topological model space (shown in the fourth column). (c) The robot discovers the second familiar obstacle (labeled 1 in the physical space), driving the hybrid dynamical system into the terminal (Definition \ref{definition:terminal_mode}) mode $\hybridmode = \{1,2\}$, wherein it corrects the initial hypothesis by merging the union of the two obstacles to the boundary. (d) The reactive field pushing the robot along a direct path to the goal in the unobstructed model space is deformed to generate a sharp correction of course in the geometrically accurate mapped space until, finally, (e) safely navigates to the goal. The deformation of space that aligns the geometrically informed mapped space with its topologically equivalent model space can be visualized by comparing the direct path to the goal the planner generates in the model space with its diffeomorphic image, the curved path connecting the robot's starting point to the goal in the mapped space. Note that the robot has no prior information about the structure of the hybrid system (depicted in the right-most column with unexplored modes in grey): it is driven around the hybrid graph, $\hybridgraph$, by its online perceptual experiences as it accumulates more information about its surroundings. Note, as well, that the cardinality of topological obstacles (the number of punctures in the model space) is independent of the number of semantically localized objects, $|\hybridmode|$.} \label{fig:illustration_merging}
\end{figure}

\begin{figure}[H]
\centering
\includegraphics[width=0.6\textwidth]{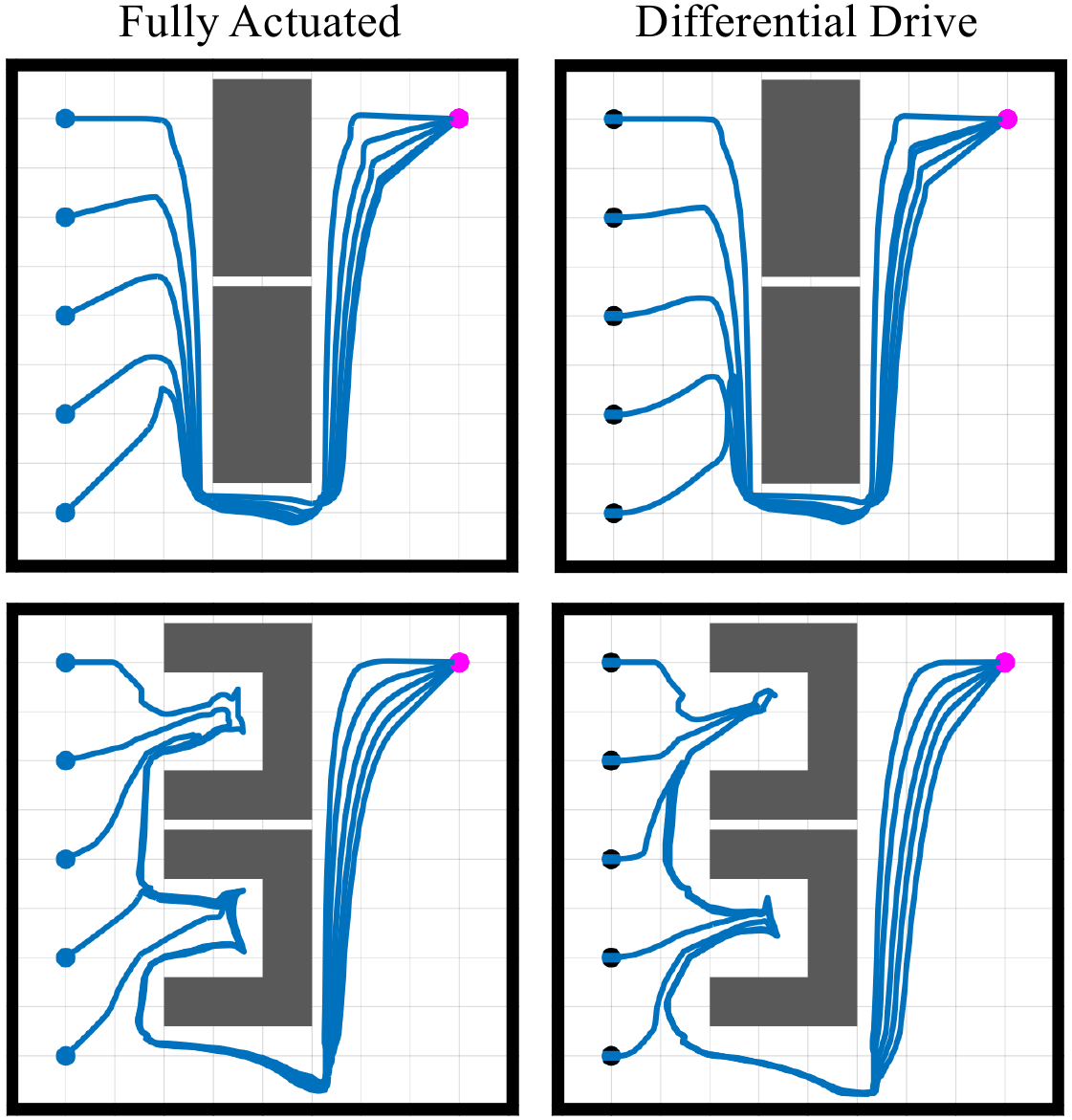}
\caption{Numerically simulated illustrations of the navigation planner's behavior from multiple initial conditions for both a fully actuated and a differential drive robot, in the presence of two familiar obstacles with \`a-priori completely unknown placement in the workspace. Top: Obstacles with rectangular shape, Bottom: U-shaped obstacles. The hybrid systems theorems presented in Section \ref{sec:controller} guarantee the robot will safely navigate to the goal with no collisions along the way.} \label{fig:merging_simple}
\end{figure}

For the next set of numerical studies, we focus on environments cluttered with several instances of the same familiar obstacle, in different, \`a-priori unknown poses. We illustrate the concept in Fig. \ref{fig:illustration_merging}. The robot abstracts away the familiar geometry to explore the unknown topology of the workspace online during execution time. In this particular example, the robot first adopts the hypothesis that an ``opening'' exists above the initially observed obstacle. With the observation and instantiation of the second obstacle in the semantic map, it is then capable of correcting this hypothesis by merging the obstacle to the boundary of $\enclosingfreespace$. The properties of the hybrid controller presented in Section \ref{sec:controller} guarantee convergence to the goal for both the fully actuated and the differential drive robot, as shown in Fig. \ref{fig:merging_simple}. 

\begin{figure}
\centering
\includegraphics[width=0.7\textwidth]{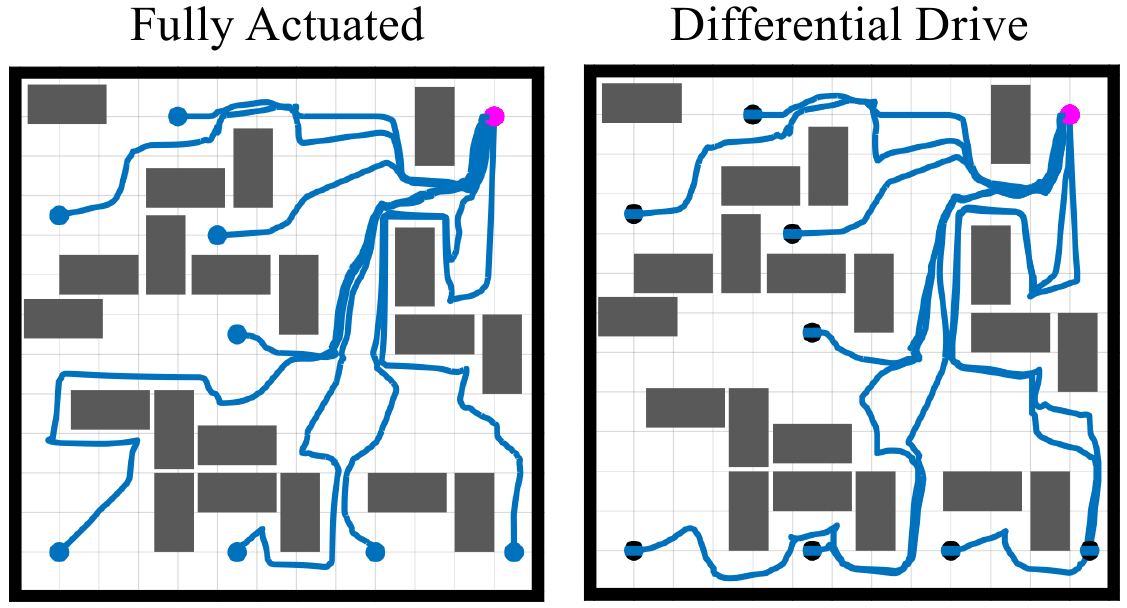}
\caption{Simulated trajectories from multiple initial conditions for both a fully actuated and a differential drive robot, in the presence of many instances of the same familiar obstacle with \`a-priori unknown pose. The robot explores the geometry and topology of the workspace online during execution time, and the guarantees of the hybrid controller in Section \ref{sec:controller} allow it to safely navigate to the goal, without converging to local minima arising from the complicated geometry of the workspace.} \label{fig:environment_cluttered}
\end{figure}

We further illustrate the scope of formal results by presenting numerical simulations where the constellation of fixed obstacles incurs the need for multiple mergings between obstacles or between obstacles and the boundary of the enclosing freespace $\enclosingfreespace$. As guaranteed, both the fully actuated (Theorem \ref{theorem:hybrid_fullyactuated}) and the differential drive (Theorem \ref{theorem:hybrid_unicycle}) robots converge to the desired goal from a variety of initial conditions (all but a set of measure zero must converge), as shown in Fig. \ref{fig:environment_cluttered}. The robot radius used in our simulations is 0.25m, the control gains are $k = k_v = k_\omega = 0.4$, and the values of $\innerpolygontune{j_i} = \innerpolygontune{r_i}$, $\outerpolygontune{j_i} = \outerpolygontune{r_i}$ and $\innerpolygondistance{j_i} = \innerpolygondistance{r_i}$ used in the diffeomorphism construction are 2.0, 0.05 and 1.0 respectively. The maximum input $u_{\max}$ for the fully actuated robot as well as the maximum linear and angular inputs $v_{\max},\omega_{\max}$ for the differential drive robot are limited to 0.4.

\subsection{Navigation Among Mixed Known and Unknown Obstacles}

\begin{figure}[H]
\centering
\includegraphics[width=0.8\textwidth]{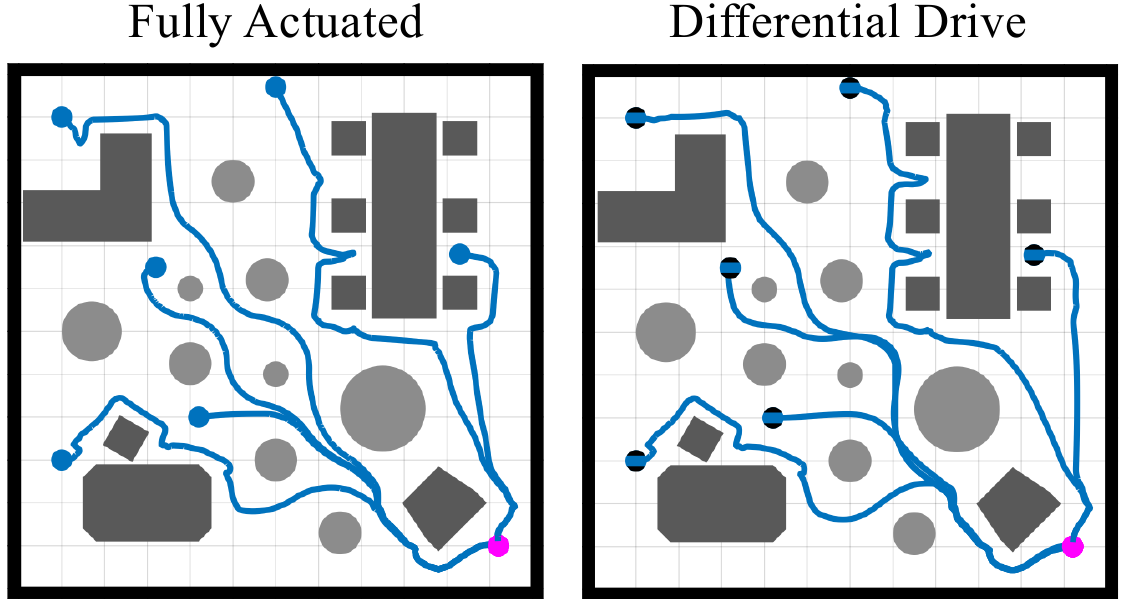}
\caption{Simulated trajectories from multiple initial conditions for both a fully actuated robot and a differential drive robot, in the presence of both familiar obstacles with \`a-priori unknown pose (dark grey) and completely unknown obstacles (light grey). The guarantees of the hybrid controller in Section \ref{sec:controller} allow the robot to always safely navigate to the goal.} \label{fig:mixed}
\end{figure}

Finally, Fig. \ref{fig:mixed} illustrates the convergence guarantees for both a fully actuated as well as a differential drive robot when confronted both by familiar obstacles (with \`a-priori unknown pose) as well as completely unknown obstacles (presumed to satisfy the convexity and separation assumptions of \cite{arslan_kod_WAFR2016}), as outlined in Section \ref{sec:problem_formulation}. The robot radius used in our simulations is 0.25m, the control gains are $k = k_v = k_\omega = 0.4$, and the values of $\innerpolygontune{j_i} = \innerpolygontune{r_i}$, $\outerpolygontune{j_i} = \outerpolygontune{r_i}$ and $\innerpolygondistance{j_i} = \innerpolygondistance{r_i}$ used in the diffeomorphism construction are 1.6, 0.05 and 0.8 respectively. The maximum input $u_{\max}$ for the fully actuated robot as well as the maximum linear and angular inputs $v_{\max},\omega_{\max}$ for the differential drive robot are limited to 0.4.
\section{Experimental Setup}
\label{sec:experimental_setup}

Because the reactive planners introduced in this paper take the form of first order vector fields (i.e., issuing velocity commands at each state), we use a quasi-static platform, the Turtlebot robot \cite{turtlebot}, for the bulk of physical experiments reported next. With the aim of merely suggesting the robustness of these feedback controllers, we also repeat two of those experiments using the highly dynamic Minitaur robot \cite{ghostminitaur}, whose rough approximation to the quasi-static differential drive motion model is adequate to yield nearly indistinguishable navigation behavior. In the conclusion, we will further motivate the importance of this legged implementation by sketching a longer term agenda emerging from recent results in \cite{Vasilopoulos_Topping_Vega-Brown_Roy_Koditschek_2018} for integrating this planner into a more complicated system that achieves mobile manipulation tasks with legged robots in dynamic environments.

\begin{figure}[H]
\centering
\includegraphics[width=0.7\textwidth]{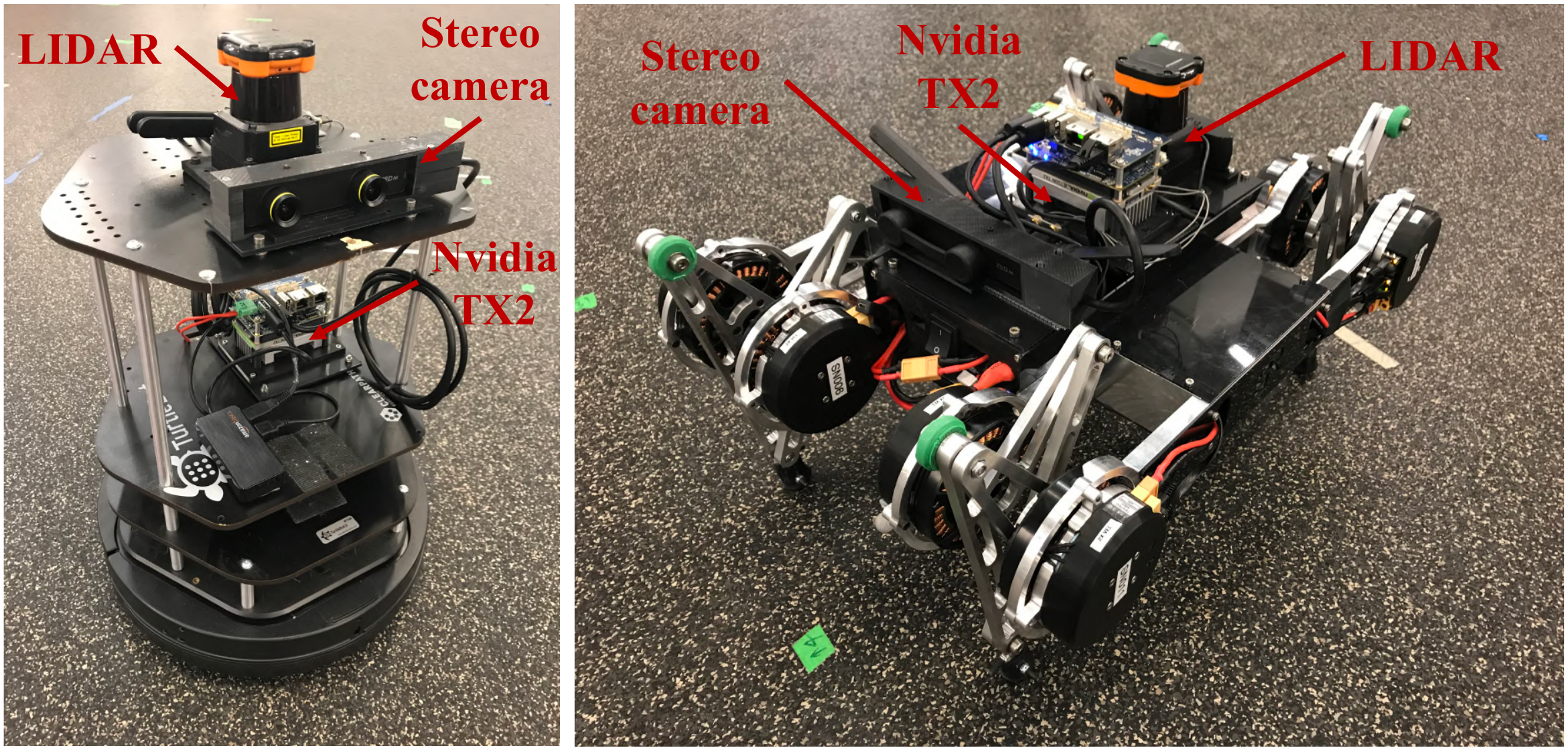}
\caption{The platforms used in our experiments: (Left) Turtlebot, (Right) Minitaur, equipped with a Hokuyo LIDAR for avoidance of unknown obstacles, a stereo camera for object recognition and visual odometry, and an NVIDIA TX2 GPU module as the main onboard computer.} \label{fig:robots}
\end{figure}

The experimental setups for our robots are depicted in Fig. \ref{fig:robots}. In both cases, the main computer is an Nvidia TX2 GPU unit \cite{nvidiatx2}, responsible for running our mapped space recovery and reactive planning algorithms online, during execution time, according to Fig. \ref{fig:algorithm}. The GPU unit communicates with a Hokuyo LIDAR \cite{hokuyolidar}, used to detect unknown obstacles, and a ZED Mini stereo camera \cite{zedmini}, used for visual-inertial state estimation and for detecting familiar obstacles. As shown in Fig. \ref{fig:algorithm}, we choose to run our perception and semantic mapping pipelines described next either {\it onboard} (using the same Nvidia TX2 GPU unit) or {\it offboard} (on a desktop computer with an Nvidia GeForce RTX 2080 GPU), for faster inference and improved performance. We also assume that the differential drive robot model, presented in \eqref{eq:unicycle_dynamics}, is the most suitable motion model for both robots. This is indeed the case for Turtlebot, and an extensive discussion on the empirical anchoring \cite{full-koditschek-1999} of the unicycle template on Minitaur is included in \cite{vasilopoulos2017}.

Since the main focus of this paper is not the development of new perception or state estimation algorithms, but rather the development of a provably correct planning architecture for partially known environments, we rely to as great an extent as possible on off-the-shelf perception algorithms, implemented in ROS \cite{ros}, and couple them with our motion planner for the hardware experiments. We are further motivated by the intent for our accompanying software to be modular and easily integrated to existing perception pipelines for future users. We briefly describe the perception and semantic mapping algorithms employed in this paper in the Sections below, and refer the reader once more to the summary illustration of the whole navigation stack in Fig. \ref{fig:algorithm}.

\subsection{Object Detection and Keypoint Localization}
\label{subsec:object_recognition}

The pipeline we use to detect the objects in the scene and extract the geometric properties needed in order to estimate their 3D pose relies on \cite{Pavlakos2017}. The two components involved in this procedure are:
\begin{itemize}
    \item Object detection, which returns 2D bounding boxes for each object.
    \item Keypoint localization, which estimates the 2D locations for a set of predefined keypoints for the specific object instance and class.
\end{itemize}
The algorithm is described in detail in \cite{Pavlakos2017}, but here we give a brief overview of each step in Sections \ref{subsubsec:object_detection} and \ref{subsubsec:keypoint_localization}, and provide training details for our neural networks in Section \ref{subsubsec:training_details}.

\subsubsection{Object Detection}
\label{subsubsec:object_detection}
For the task of object detection, we only require the estimation of a 2D bounding box for each object that is visible on the image. We use the YOLOv3 detector~\cite{yolov3} which offers a good trade-off between detection accuracy and inference speed. Given a single RGB image as input, the output of the detector is a 2D bounding box for each object instance, along with the estimated class for this bounding box.

\subsubsection{Keypoint Localization}
\label{subsubsec:keypoint_localization}
%Given the list of 2D bounding boxes from the detection network, we need to estimate their 3D pose with respect to the camera. This is a two-step procedure, which involves the detection of a predefined set of 2D keypoints on the object and then it requires the lifting of these keypoints to a reasonable 3D pose.
For the keypoint localization task, we use a Convolutional Neural Network to accurately estimate the 2D location of the keypoints within the object's bounding box. 
The keypoints are defined on the 3D model of the object and are selected in advance for each object instance.
The keypoint localization network uses as input an RGB image of a specific object, which is cropped using the bounding box information from the detection step.
The output of the network is a set of 2D heatmaps.
Assuming we select $k$ keypoints for an object, each heatmap is responsible for the localization of the corresponding keypoint.
To estimate the locations $\mathbf{W} \in \mathbb{R}^{2 \times k}$ of the $k$ keypoints on the image, we use the 2D heatmap location with the maximum activation for each heatmap as the detected location for the corresponding keypoint. 
We also consider the value of the activation at this location as the detection confidence $d_i$ for keypoint $i$.
The architecture for this network follows the Stacked Hourglass design~\cite{stacked-hourglass}.
In practice, we train a single network for all objects of interest and at test time we use only the heatmaps for the specific class, which is already known from the detection step.

\subsubsection{Training Details}
\label{subsubsec:training_details}
\begin{figure}[t]
\centering
\includegraphics[width=1.0\textwidth]{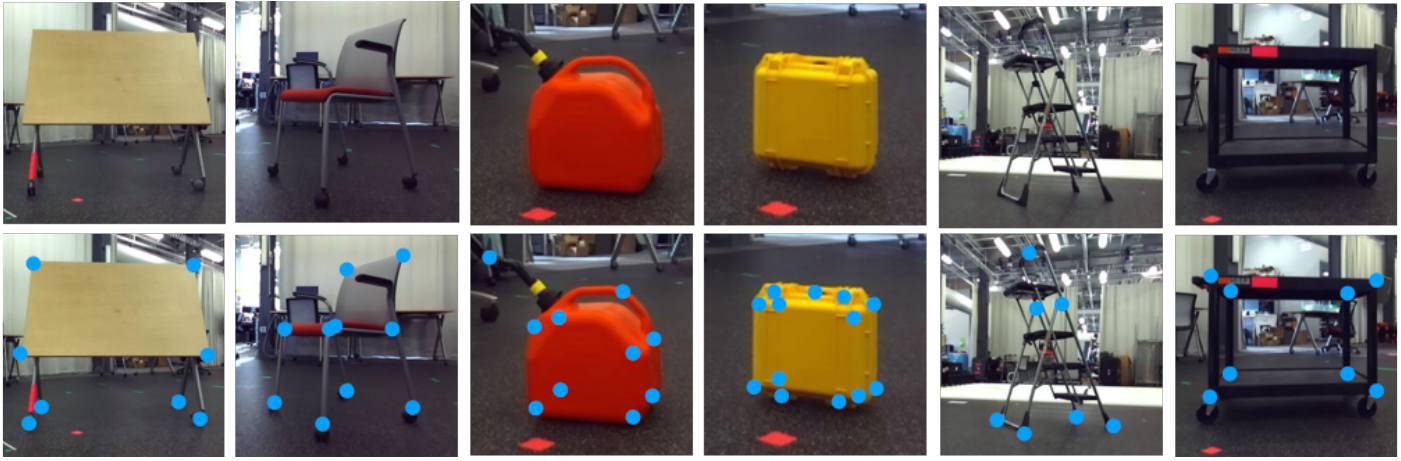}
\caption{Top row: Objects used in our experimental setup: table, chair, gascan, pelican case, ladder, cart. Bottom row: Visualization of the semantic keypoints for each object class.} \label{fig:objects}
\end{figure}

The aforementioned neural networks are trained to detect a predefined set of object instances visualized in Figure~\ref{fig:objects}.
The object classes represented for our experiments are chair, table, ladder, cart, gascan and pelican case.
Our goal is to include a variety of instances in terms of the size, shape and visual appearance, in an attempt to simulate the variety of objects that can be encountered in a partially familiar environment.
The training data for the particular instances of interest are collected with a semi-automatic procedure, similarly to \cite{Pavlakos2017}.
Given the bounding box and keypoint annotations for each image, the two networks were trained with their default configurations until convergence.

% Describe assumptions made about:
% \begin{enumerate}
%     \item objects used
%     \item acceptable object poses
%     \item object overlaps in the image
%     \item highlight that we use keypoints to accelerate the online process instead of dealing with meshes
% \end{enumerate}

% + more details about the experimental setting
% e.g., we assumed x specific object instances

\subsection{Semantic Mapping}
\label{subsec:semantic_mapping}

Our semantic mapping infrastructure relies on the algorithm presented in \cite{Bowman2017}, and implemented in C++ using GTSAM \cite{dellaert-gtsam} and its iSAM2 implementation \cite{isam2} as the optimization back-end. Briefly, this algorithm fuses inertial information (here simply provided by the position tracking implementation from StereoLabs on the ZED Mini stereo camera \cite{state-estimation-zedmini}), and semantic information (i.e., the detected keypoints and the associated object labels as described in Section \ref{subsec:object_recognition}) to provide a posterior estimate for both the robot state and the associated poses for all tracked objects, by simultaneously solving the data association problem arising when several objects of the same class exist in the map. As described in \cite{Bowman2017}, except for providing an estimate for all poses tracked in the environment, this algorithm facilitates loop closure recognition based on viewpoint-independent semantic information (i.e., tracked objects), rather than low-level geometric features such as points, lines, or planes.

For a single frame detection, the 3D pose of each object with respect to the camera is recovered using the estimated 2D locations of the associated object keypoints. By denoting with $\mathbf{S} \in \mathbb{R}^{3 \times k}$ the 3D locations of the keypoints in the canonical pose of an object instance with $k$ keypoints, the goal is to estimate the rotation $\mathbf{R} \in \mathbb{R}^{3 \times 3}$ and translation $\mathbf{T} \in \mathbb{R}^{3 \times 1}$ of the object, such that the distance of the projected 3D keypoints from their corresponding detected 2D locations is minimized. To incorporate the detection confidence for each keypoint in the optimization, we define the matrix $\mathbf{D} \in \mathbb{R}^{k \times k}$. This is a diagonal matrix that features the detection confidences $d_i$ for each keypoint $i$ in its diagonal. The optimization problem is then formulated as:
\begin{align}\label{eq:cost}
    \min_{\mathbf{R}, \mathbf{T}} ~~ & \frac{1}{2} \left\| (\tilde{\mathbf{W}}\mathbf{Z}- {\mathbf{R}} \mathbf{S} - {\mathbf{T}}\mathbf{I}^\top)\mathbf{D}^{\frac{1}{2}} \right\|_F^2,
\end{align}
where $\tilde{\mathbf{W}} \in \mathbb{R}^{3 \times k}$ represents the normalized homogeneous coordinates of the 2D keypoints and $\mathbf{Z} \in \mathbb{R}^{k \times k}$ is a diagonal matrix, that features the depths $z_i$ for each keypoint $i$ in its diagonal. 

\begin{figure}[t]
\centering
\includegraphics[width=0.6\textwidth]{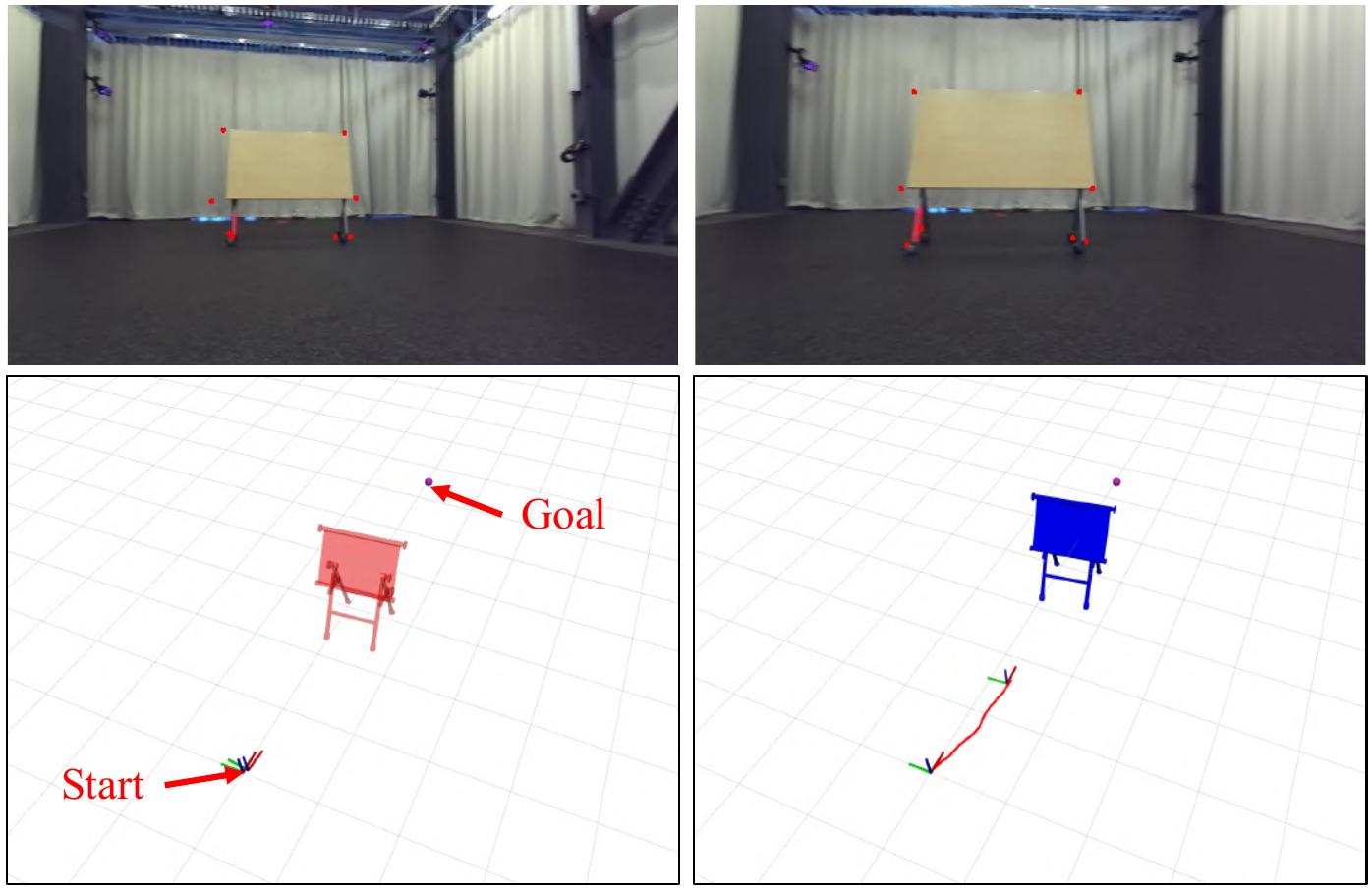}
\caption{Illustration of the object localization process using the semantic mapping pipeline from \cite{Bowman2017}. Left: The robot starts navigating toward its goal and discovers a familiar obstacle (table). The obstacle is temporarily included in the semantic map, after its 3D pose is estimated using a single frame measurement \eqref{eq:cost} (red). Right: Once a sufficient number of frame measurements has been incorporated and the 3D pose has been accordingly updated, the object is permanently localized and included in the semantic map (blue).} \label{fig:localization_example}
\end{figure}

After the estimation of the object's 3D pose from a {\it single frame measurement} as described above, the 3D positions of its corresponding semantic keypoints are then independently tracked and the object's pose is appropriately updated, as more frame measurements are added. Once a sufficient number of frame measurements\footnote{This number depends on the needed camera motion between successive measurements, in order to establish a good baseline for triangulation \cite{hartley_triangulation_1997}; in this work, we found that 5 measurements for the offboard experiments and 3 measurements for the onboard experiments yielded reasonably fast keypoint localization.} has been incorporated so that the 3D keypoint positions can be triangulated, the object is considered to be {\it localized} and is permanently added to the map. The reader is referred to \cite{Bowman2017} for more details. Fig. \ref{fig:localization_example} shows an example of this localization process. It should be noted that for our onboard implementation, where inference using the object detection and keypoint estimation neural networks is slower, we include in the semantic map both the localized objects, after several frame measurements, and objects resulting from a single frame measurement pose estimation, to allow for faster response to sensory input.

As shown in Fig. \ref{fig:algorithm}, the meshes of the objects in the semantic map, defined by the corresponding keypoint adjacency properties and the extracted 3D pose, are projected on the robot's plane of motion to provide the aggregated list of known obstacles in the physical space $\knownobstacleset_{\hybridmode}$, forwarded to our mapped space recovery module (described in Section \ref{subsec:mapped_space_recovery}). On the other hand, the posterior estimate of the robot pose on the plane, extracted by the semantic mapping module, is forwarded to our reactive planning module (described in Section \ref{subsec:reactive_planning}).
\section{Experimental Results}
\label{sec:experiments}

In this Section, we provide our experimental results using both the Turtlebot and the Minitaur robot, and the setup described in Section \ref{sec:experimental_setup}. We begin with experiments run using Turtlebot and offboard (Section \ref{subsec:turtlebot_offboard}) or onboard (Section \ref{subsec:turtlebot_onboard}) perception, and continue with Minitaur experiments using offboard perception (Section \ref{subsec:minitaur}), to demonstrate the robustness of our method on a more dynamic legged platform. It should be noted that although the perception algorithms, described in Section \ref{sec:experimental_setup}, are run either offboard or onboard, our mapped space recovery and reactive planning modules, described in Algorithms \ref{algorithm:mapped_space_recovery} and \ref{algorithm:reactive_planning} respectively, are always run onboard each robot's Nvidia TX2 module. The control gains used in our experiments are $k_v = k_\omega = 0.4$, and the maximum linear and angular inputs $v_{\max},\omega_{\max}$ are set to 0.4.

\begin{figure}[H]
\centering
\includegraphics[width=0.7\textwidth]{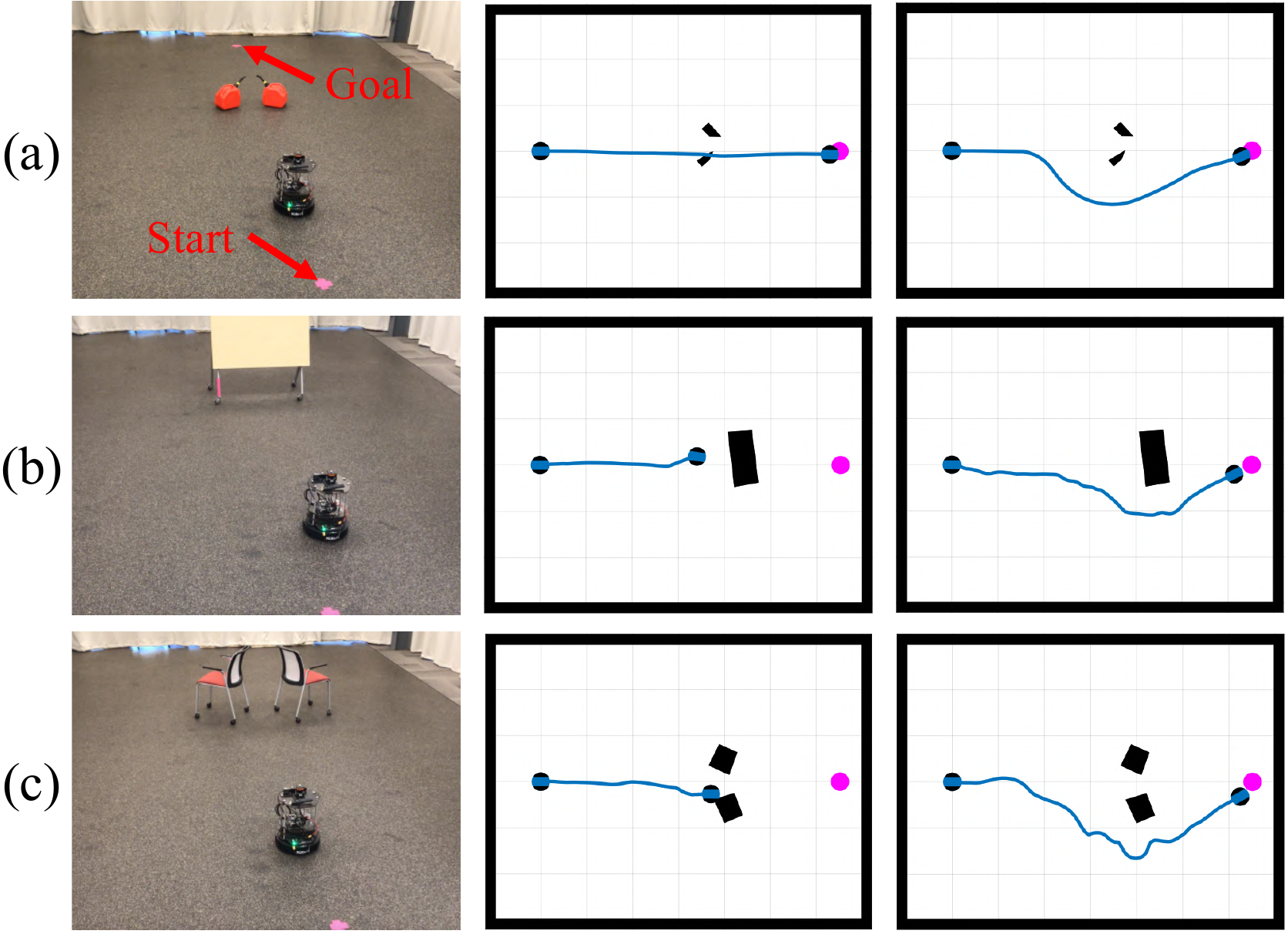}
\caption{Physical experiments akin to the numerical simulations depicted in Fig. \ref{fig:comparison}, comparing the original doubly reactive algorithm \cite{arslan_kod_WAFR2016} (middle column) with our algorithm (right column) in different physical settings (left column), using Turtlebot and offboard perception. (a) Two gascans forming a non-convex trap, (b) Table used as a flat obstacle, (c) Two chairs violating the separation assumptions of \cite{arslan_kod_WAFR2016}.} \label{fig:experimental_comparison}
\end{figure}

\subsubsection{Comparison with Original Doubly Reactive Algorithm}
In this Section, we demonstrate experiments similar to the simulations reported in Section \ref{subsec:comparison_numerical}. We first illustrate various well understood failures of the original version of the doubly reactive algorithm in \cite{arslan_kod_WAFR2016}. Collisions result from the presence of short obstacles that cannot be detected by the 2D LIDAR (Fig. \ref{fig:experimental_comparison}-(a)). Confronted by obstacles with flat surfaces (Fig. \ref{fig:experimental_comparison}-(b)), or when separation assumptions are violated (Fig. \ref{fig:experimental_comparison}-(c)), the original algorithm gets stuck in undesired local minima (Fig. \ref{fig:experimental_comparison}-(b),(c)). In contrast, our new algorithm guarantees safe convergence to the goal in all these cases: short but familiar obstacles (in this case the gascan in column 3 of Fig. \ref{fig:objects}) are recognized by the camera system and localized; once localized, these known geometries can then be appropriately abstracted into the model space (Section \ref{sec:diffeomorphism}) which is topologically equivalent but geometrically simplified to meet the requirements of \cite{arslan_kod_WAFR2016}. Fig. \ref{fig:experimental_comparison} shows the groundtruth trajectory of the robot, recorded using Vicon, along with 2D projections on the horizontal plane of the obstacles' keypoint meshes, that were used for the construction of the semantic space (Section \ref{subsec:semantic_space}). The values of $\innerpolygontune{j_i} = \innerpolygontune{r_i}$, $\outerpolygontune{j_i} = \outerpolygontune{r_i}$ and $\innerpolygondistance{j_i} = \innerpolygondistance{r_i}$ used in the diffeomorphism construction are 4.0, 0.05 and 2.0 respectively.

\begin{figure}[H]
\centering
\includegraphics[width=1.0\textwidth]{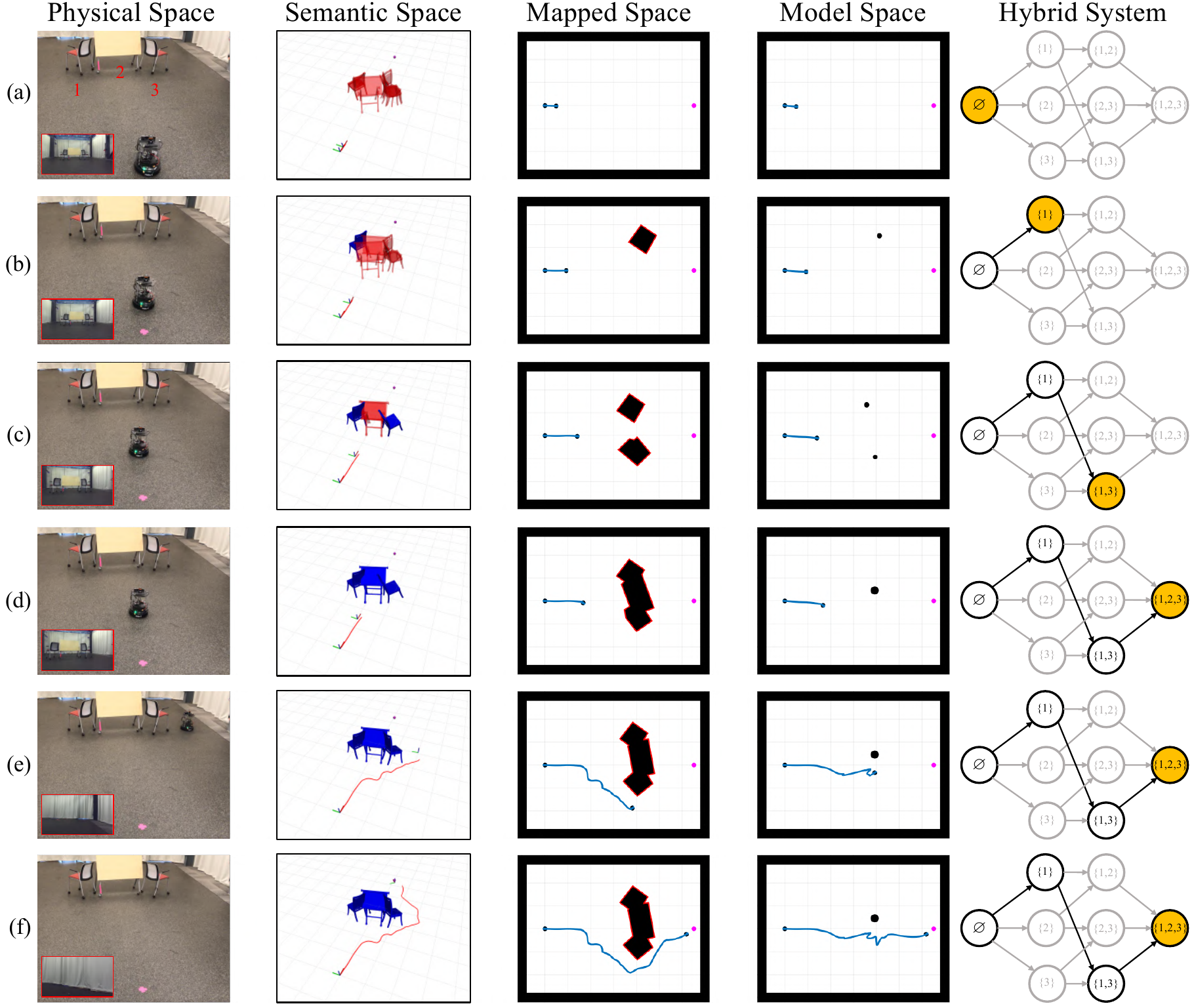}
\caption{Illustration of the empirically implemented complete navigation scheme (akin to the numerical simulation depicted in Fig. \ref{fig:illustration_merging}) in a physical setting where three familiar obstacles (two chairs and a table) form a non-convex trap. (a) The robot starts navigating toward its designated target in a previously unknown environment, and detects familiar obstacles. The initial mode of the hybrid system is $\hybridmode = \varnothing$. (b)-(d) The robot keeps localizing familiar obstacles, and changes its belief about the topological state of the workspace (as evident in the column showing the corresponding model space). (e) Using the information in the semantic space and now being in the terminal (Definition \ref{definition:terminal_mode}) mode $\hybridmode = \{1,2,3\}$, wherein it has encountered and localized all the environment's familiar obstacles, the robot is driven by the mapped space transformation (Section \ref{sec:diffeomorphism}) of the model space vector field \cite{arslan_kod_WAFR2016} to avoid the obstacles, until (f) it converges to the designated goal as guaranteed by the results of Section \ref{sec:controller}. The right column shows how the robot experiences transitions in the (previously unknown) hybrid system (modes that are never experienced are shown in grey).} \label{fig:experimental_hybrid}
\end{figure}

\subsection{Experiments with Turtlebot and Offboard Perception}
\label{subsec:turtlebot_offboard}

\subsubsection{Navigation in a Cluttered Environment with Obstacle Merging}

We begin the second set of experiments by demonstrating the merging process and the properties of the hybrid controller, reported in Section \ref{sec:controller}, in a physical setting. As shown in Fig. \ref{fig:experimental_hybrid}, the robot starts navigating toward its target and localizing obstacles in front of it, until it converges to its target; at the same time, by incorporating more information in its semantic map, it experiences transitions to different modes of the (previously unknown) hybrid system. The values of $\innerpolygontune{j_i} = \innerpolygontune{r_i}$, $\outerpolygontune{j_i} = \outerpolygontune{r_i}$ and $\innerpolygondistance{j_i} = \innerpolygondistance{r_i}$ used in this experiment are 4.0, 0.05 and 2.0 respectively.

\begin{figure}
\centering
\includegraphics[width=1.0\textwidth]{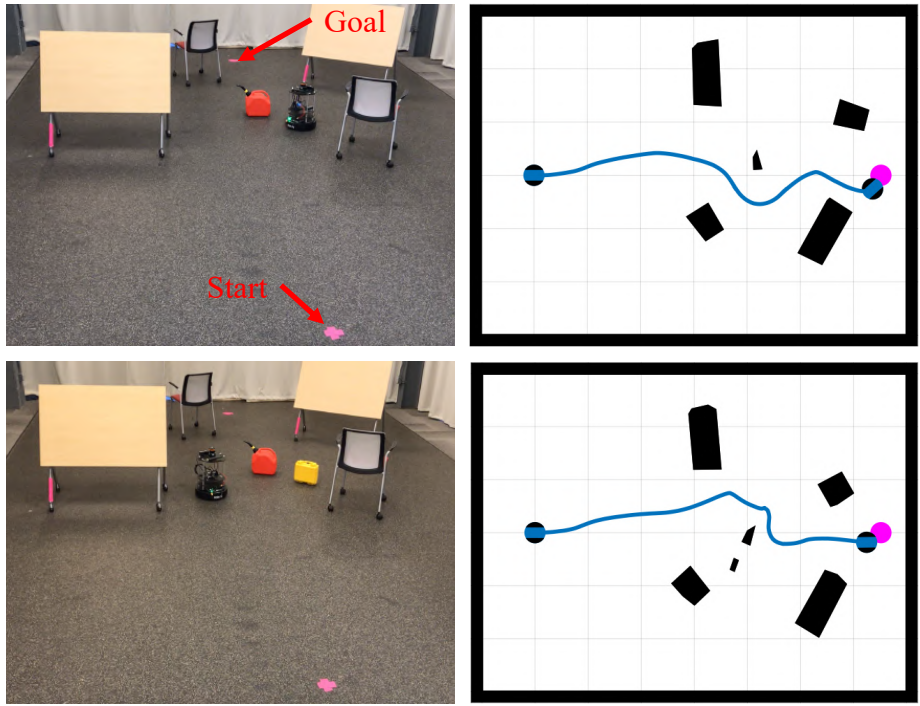}
\caption{Navigation among multiple familiar obstacles, using Turtlebot and offboard perception. Top: The robot exploits the gap between the gascan and the chair to safely navigate to the goal. Bottom: When we block this gap by another familiar obstacle (pelican case), the robot reactively chooses to follow another safe and convergent trajectory, by consolidating the semantic triad \{gascan, pelican case, chair\} into a single, ``mapped'' obstacle in $\knownobstaclesetdilatedmappeddisk$.} \label{fig:experimental_multiple}
\end{figure}

Finally, Fig. \ref{fig:experimental_multiple} demonstrates navigation in environments cluttered with multiple familiar obstacles. In the first illustration, the robot reactively chooses to navigate through a gap between the gascan and a chair. Despite the blockage of this gap by another familiar obstacle (pelican case) in the second illustration, the robot reactively chooses to follow another safe and convergent trajectory (as guaranteed by the theorems of Section \ref{sec:controller}), by merging the set gascan - pelican case - chair, and considering them as a single obstacle. The values of $\innerpolygontune{j_i} = \innerpolygontune{r_i}$, $\outerpolygontune{j_i} = \outerpolygontune{r_i}$ and $\innerpolygondistance{j_i} = \innerpolygondistance{r_i}$ used in this experiment are 1.6, 0.05 and 0.8 respectively.

\subsubsection{Navigation Among Mixed Known and Unknown Obstacles}

\begin{figure}
\centering
\includegraphics[width=1.0\textwidth]{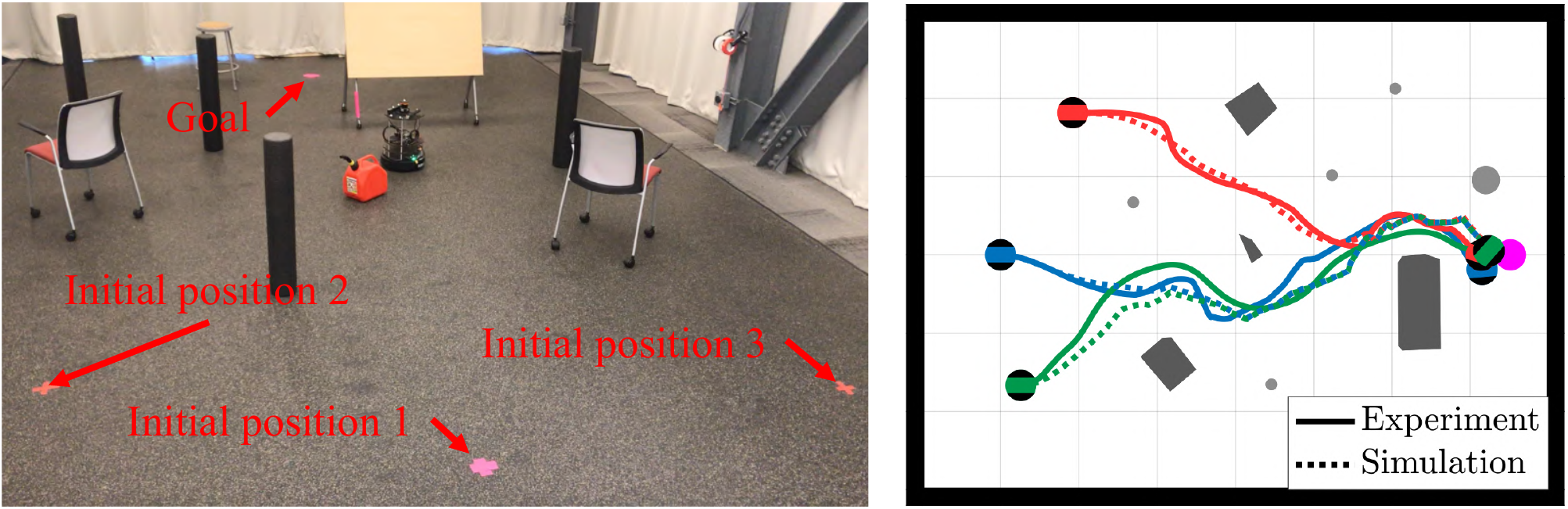}
\caption{Navigation among familiar and unknown obstacles, using Turtlebot and offboard perception, from three different initial conditions. Left: A snapshot of the physical workspace. Right: A ``bird's-eye'' view of the workspace, with 2D projections of the localized familiar obstacles (dark grey) and unknown obstacles (light grey - groundtruth locations recorded using Vicon), along with groundtruth trajectories from the physical experiments and overlaid numerical simulations in MATLAB.} \label{fig:experimental_known_unknown}
\end{figure}

In the next set of experiments, we consider navigation among multiple familiar and unknown obstacles. Fig. \ref{fig:experimental_known_unknown} shows that the robot safely converges to the goal from multiple initial conditions, using vision and the setup described in Section \ref{sec:experimental_setup} for familiar obstacle detection and localization, and the onboard 2D LIDAR for all the unknown obstacles. In Fig. \ref{fig:experimental_known_unknown}, we also overlay trajectories from a MATLAB simulation of a differential-drive robot with the same initial conditions and similar control gains; the simulated and physical platform follow similar trajectories in all three cases. The values of $\innerpolygontune{j_i} = \innerpolygontune{r_i}$, $\outerpolygontune{j_i} = \outerpolygontune{r_i}$ and $\innerpolygondistance{j_i} = \innerpolygondistance{r_i}$ used in these experiments are 2.0, 0.05 and 1.0.

It should be highlighted that even when the object localization process fails, collision avoidance is still guaranteed with the use of the onboard LIDAR. Nevertheless, collisions could result with obstacles that cannot be detected by the 2D horizontal LIDAR (e.g., see Fig. \ref{fig:experimental_comparison}-(a)). One could still think of extensions to the presented sensory infrastructure (e.g., the use of a 3D LIDAR) that could still guarantee safety under such circumstances.

\subsection{Experiments with Turtlebot and Onboard Perception}
\label{subsec:turtlebot_onboard}

\begin{figure}[H]
\centering
\includegraphics[width=1.0\textwidth]{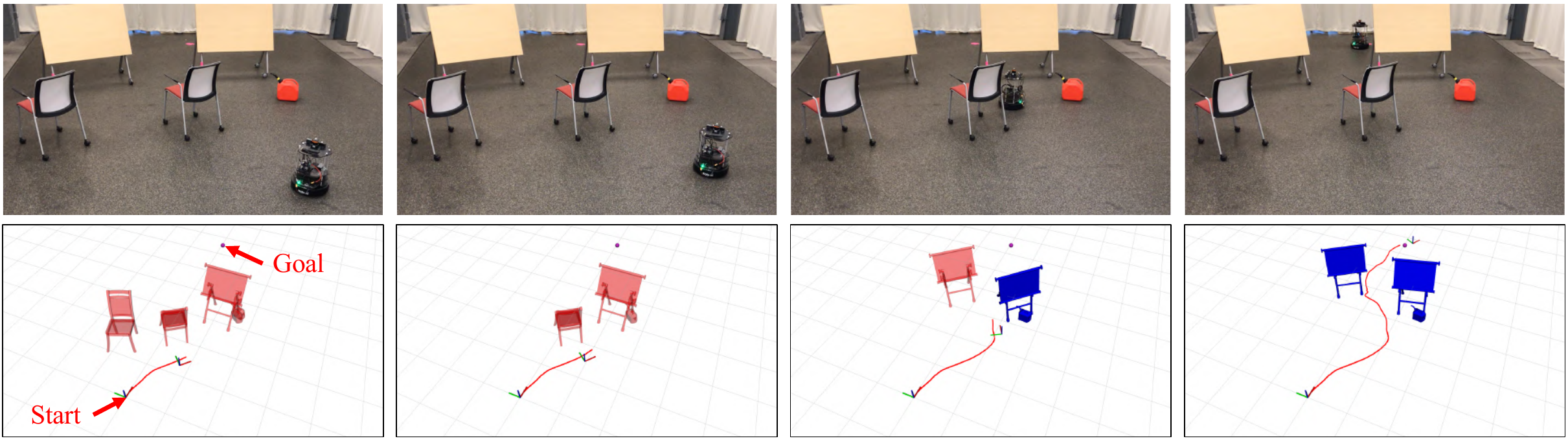}
\caption{Navigation among familiar obstacles, using Turtlebot and onboard perception. Top: snapshots of the physical workspace, Bottom: illustrations of the recorded semantic map and the robot's trajectory in RViz \cite{ros}. The robot detects and avoids the two chairs in front of it, though they are only temporarily included in the semantic map (in the absence of more frame measurements). Then it proceeds to localize and avoid the two tables and the gascan, to safely converge to the goal.} \label{fig:experimental_turtlebot_onboard}
\end{figure}

This Section briefly reports on experiments using onboard perception. As described in Section \ref{subsec:semantic_mapping}, here we use both the localized obstacles by the semantic mapping pipeline and raw, not permanently localized obstacles, resulting from a single semantic frame measurement and the optimization problem given in \eqref{eq:cost}. Fig. \ref{fig:experimental_turtlebot_onboard} illustrates an example; the robot detects and avoids the two chairs in front of it, even if they are only temporarily included in the semantic map (in the absence of more frame measurements). The robot then proceeds to localize and avoid the gascan and the two tables and safely converge to the designated goal. The values of $\innerpolygontune{j_i} = \innerpolygontune{r_i}$, $\outerpolygontune{j_i} = \outerpolygontune{r_i}$ and $\innerpolygondistance{j_i} = \innerpolygondistance{r_i}$ used in this experiment are 2.0, 0.05 and 1.0 respectively.

It should be noted that the object impermanence in the semantic map violates the formal assumptions of Theorems \ref{theorem:hybrid_fullyactuated} and \ref{theorem:hybrid_unicycle}; without permanently localizing an object, the robot could get stuck in an endless loop trying to avoid obstacles that it then ``forgets'', in unfavorable workspace configurations (e.g., like those reported in Fig. \ref{fig:merging_simple}).

\subsection{Experiments with Minitaur}
\label{subsec:minitaur}

\begin{figure}[H]
\centering
\includegraphics[width=1.0\textwidth]{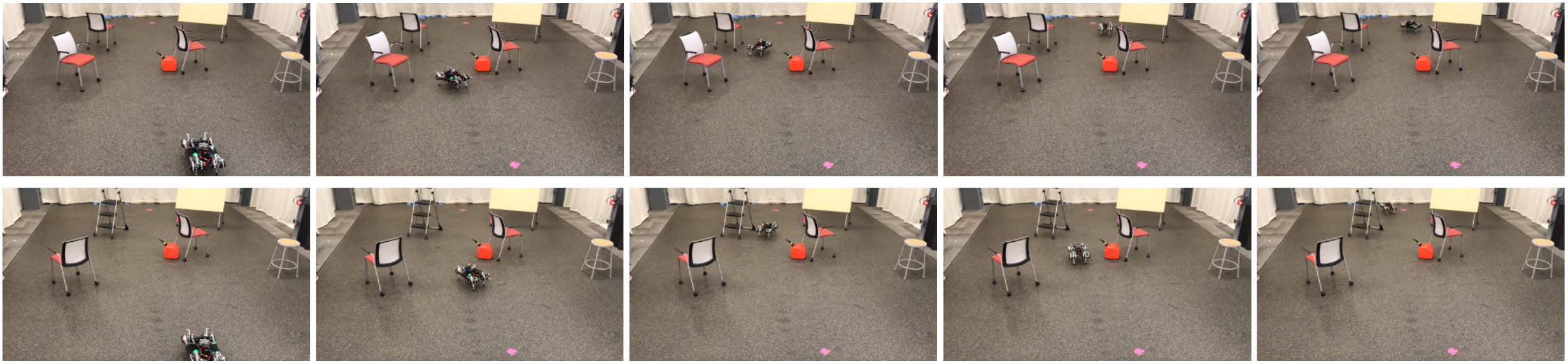}
\caption{Snapshots of Minitaur avoiding multiple familiar obstacles in two different settings, using offboard perception.} \label{fig:experimental_minitaur}
\end{figure}

Finally, Fig. \ref{fig:experimental_minitaur} presents illustrative snapshots of two navigation examples on the much more dynamic Minitaur platform. Despite the fact that Minitaur is an imperfect kinematic unicycle and the overall shakiness of the platform, the robot is capable of detecting and localizing familiar obstacles of interest and using that information to safely converge to the target. The values of $\innerpolygontune{j_i} = \innerpolygontune{r_i}$, $\outerpolygontune{j_i} = \outerpolygontune{r_i}$ and $\innerpolygondistance{j_i} = \innerpolygondistance{r_i}$ used in this experiment are 2.0, 0.05 and 1.0 respectively.

\section{Conclusion and Future Work}
\label{sec:conclusion}

\begin{figure}[H]
\centering
\includegraphics[width=1.0\textwidth]{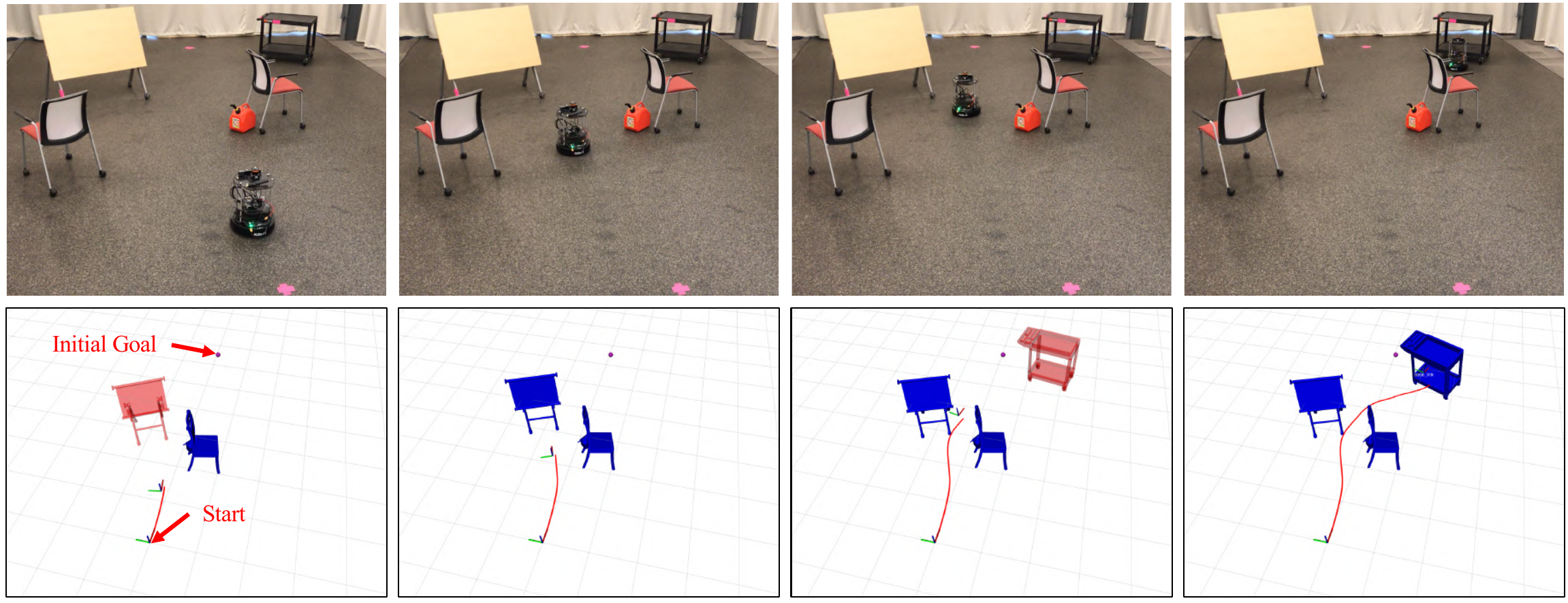}
\caption{Navigation toward a semantic target with Turtlebot. The robot is initially tasked with moving to a predefined location, unless it detects and localizes a cart; in that case it has to approach and face the cart. The last column (Top: snapshot of the physical workspace, Bottom: illustration of the recorded trajectory in RViz) shows that the robot successfully executes the task.} \label{fig:turtlebot_semantic}
\end{figure}

\begin{figure}[H]
\centering
\includegraphics[width=1.0\textwidth]{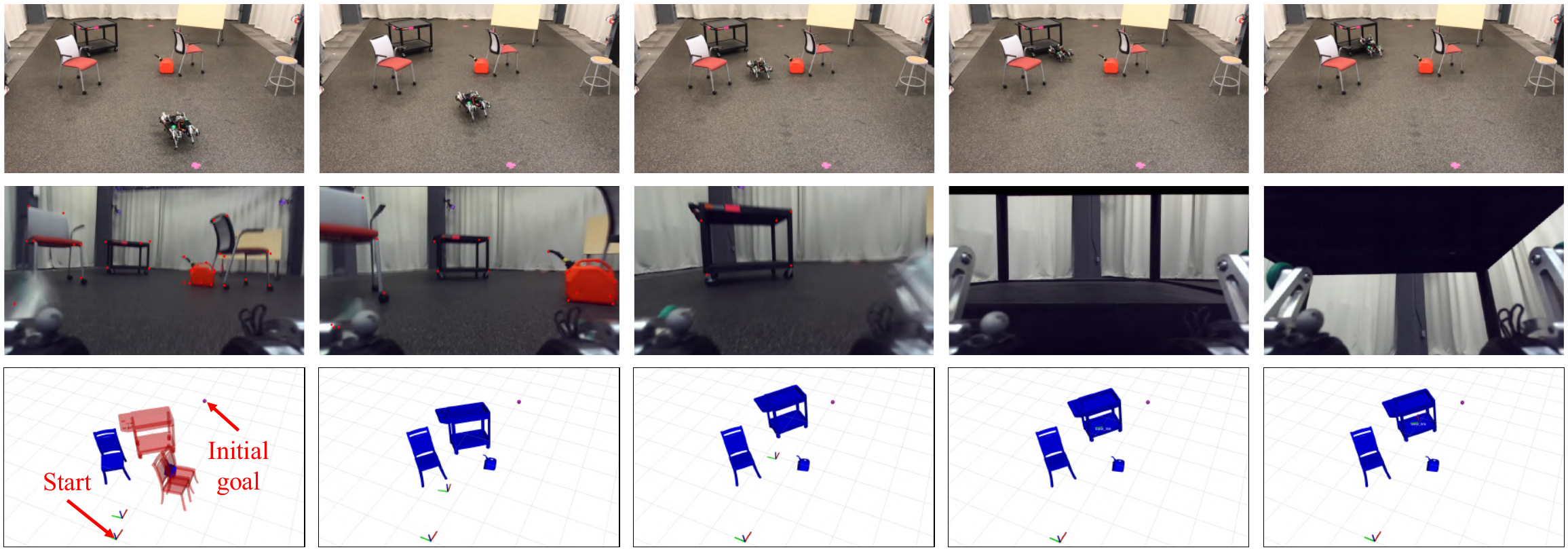}
\caption{Using reactive navigation with mobile manipulation primitives on Minitaur. Similarly to Fig. \ref{fig:turtlebot_semantic}, the robot is tasked with moving to a predefined location, unless it detects and localizes a cart; in that case it has to approach and jump to mount the cart, using a maneuver from \cite{topping_vasilopoulos_de_koditschek_2019}. Top: Recorded snapshots of the physical workspace, Middle: First-person view with semantic keypoints of familiar obstacles shown as red dots, Bottom: RViz illustration of the recorded semantic map.} \label{fig:minitaur_semantic}
\end{figure}

\subsection{Conclusion}
This paper presents a reactive navigation scheme for robots operating in planar workspaces, cluttered with obstacles of familiar geometry but \`a-priori unknown placement, and completely unknown, but strongly convex and well-separated obstacles. To the best of our knowledge, this is the first doubly reactive navigation framework (i.e., a scheme where not only the robot's trajectory but also the vector field that generates it are computed online at execution time) that can handle arbitrary polygonal shapes in real time without the need for specific separation assumptions between the familiar obstacles. The resulting algorithm combines state-of-the-art perception and object recognition techniques (based on neural network architectures) for familiar obstacles, with local range measurements (e.g., LIDAR) for the unknown obstacles, to yield provably correct navigation in geometrically complicated environments. We illustrate the practicability of this approach by reporting empirical results using modest computational hardware on a wheeled robot, and the intrinsic robustness of such reactive schemes by a second implementation on a dynamic legged platform, exhibiting imperfect fidelity to the differential drive model assumed in the formal results.

\subsection{Future Work}
Figs. \ref{fig:turtlebot_semantic} and \ref{fig:minitaur_semantic} present snapshots from experiments in settings falling outside the scope of our formal results to illustrate some of the future directions opened up by the reactive planner presented in this paper. Using the feature of semantic inference provided by the semantic mapping pipeline described in Section \ref{subsec:semantic_mapping}, the user can command the robot to target a {\it semantic} goal, instead of a merely geometric one (considered in this paper). Namely, in Fig. \ref{fig:turtlebot_semantic}, we command the Turtlebot robot to move to a geometrically predefined target, unless it sees and localizes a cart; in that case, it is tasked with approaching and facing the cart with its camera. As shown in the bottom row of Fig. \ref{fig:turtlebot_semantic}, the robot avoids familiar obstacles, localizes the cart and proceeds to properly approach it, with the right orientation. We take this approach one step further with the example shown in Fig. \ref{fig:minitaur_semantic}, using the Minitaur platform. Using the mobile manipulation primitives developed in \cite{topping_vasilopoulos_de_koditschek_2019}, we task the robot by not only localizing and approaching the cart, but also jumping to grab and mount it. This is a first step toward integrating the reactive planning architecture developed in this work in the multi-layer architecture presented in \cite{Vasilopoulos_Topping_Vega-Brown_Roy_Koditschek_2018}, for accomplishing increasingly complicated mobile manipulation tasks with underactuated legged robots in environments that are semantically partially known and geometrically unknown. Parallel work, relying on the formal guarantees presented in this paper, has already demonstrated how the same vector field planning principles and our semantic inference capabilities can be exploited in order to perform more complex missions with predefined logic that involve human following and pose tracking \cite{vasilopoulos_pavlakos_bowman_caporale_daniilidis_pappas_koditschek_2020}, tasks such as navigation among movable obstacles \cite{vasilopoulos_kantaros_pappas_koditschek_2021}, using a linear temporal logic planner \cite{kantaros_2020} and an interface layer that translates symbolic commands to point navigation tasks, or complicated mobile manipulation tasks with legged robots in unexplored 2.5D environments, using an external geometric planner to rearrange semantically tagged objects of interest \cite{vegabrown_vasilopoulos_castro_koditschek_roy_2021}.

Moreover, a remaining challenge is to generalize the present framework beyond the current restriction to 2D environments, in order to address the challenge of navigating unknown or partially known environments in higher dimension. Even though the currently presented algorithm would be restricted to shapes with genus zero (no holes), one could develop algorithms that ``patch'' the holes of shapes with non-zero genus when they are not important, affording the use of the same reactive principles for navigation. Work currently in progress investigates whether concepts from the literature on convex decomposition of polyhedra \cite{lien-amato-2007} could afford such generalization to the problem of navigating 3D workspaces with aerial drones.

Finally, we believe the methods we develop here for generating in real time simple, topologically equivalent model spaces and pulling back the model controller through the corresponding diffeomorphism can be applied to diverse, philosophically alternative approaches to our purely reactive formulation of motion planning. For example, sampling-based (probabilistically complete) offline planners have been shown to benefit from integration with even geometrically naive locally reactive methods \cite{arslan_pacelli_koditschek_2017} that can mitigate difficulties such as finding paths through narrow passages. We imagine that even greater simplification of the steering and collision-checking issues arising from sampling-based methods in partially ``familiar'' geometrically complicated environments \cite{lavalle_kuffner_2001,Bialkowski_Karaman_Otte_Frazzoli_2012} might be achieved by shifting the problem of finding a feasible path to a topologically equivalent, metrically simple abstracted model wherein planning might be significantly faster. The robot could then be tasked to follow a generated path in the abstract space (e.g., along the lines of \cite{arslan_kod_ICRA2017}) and the associated commands can be pulled back to the physical space through the diffeomorphism. Careful future inquiry will be needed to explore such deliberative-reactive hybrid uses for the online topological abstraction of familiar geometry developed here.
\section*{Acknowledgements}
This work was supported in part by AFRL grant FA865015D1845 (subcontract 669737-1), and in part by ONR grant \#N00014-16-1-2817, a Vannevar Bush Fellowship held by the last author, sponsored by the Basic Research Office of the Assistant Secretary of Defense for Research and Engineering. 

The authors thank Dr. Omur Arslan for many formative discussions and for sharing his simulation and presentation infrastructure, Prof. Elon Rimon for many interesting discussions pertaining to results from the field of computational geometry, Prof. George Pappas and Sean Bowman for sharing their semantic mapping framework, and T. Turner Topping for assistance with the Minitaur hardware experiments.

\bibliographystyle{splncs04}
\bibliography{references}

\appendix
\section*{Appendix}
\addcontentsline{toc}{section}{Appendix}
\renewcommand{\thesubsection}{\Alph{subsection}}

Appendix \ref{appendix:implicit} sketches the ideas from computational geometry \cite{shapiro2007} underlying our modular construction of implicit representations of polygonal obstacles, Appendix \ref{appendix:computational_geometry} provides some computational details associated with the diffeomorphism construction presented in Section \ref{sec:diffeomorphism}, Appendix \ref{appendix:calculation_jacobian} describes our method for the inductive computation of the diffeomorphism and its derivatives, and Appendix \ref{appendix:proofs} includes the proofs of our main results.

\subsection{Implicit Representation of Obstacles with R-functions}
\label{appendix:implicit}

\begin{figure}[H]
\centering
\includegraphics[width=0.8\textwidth]{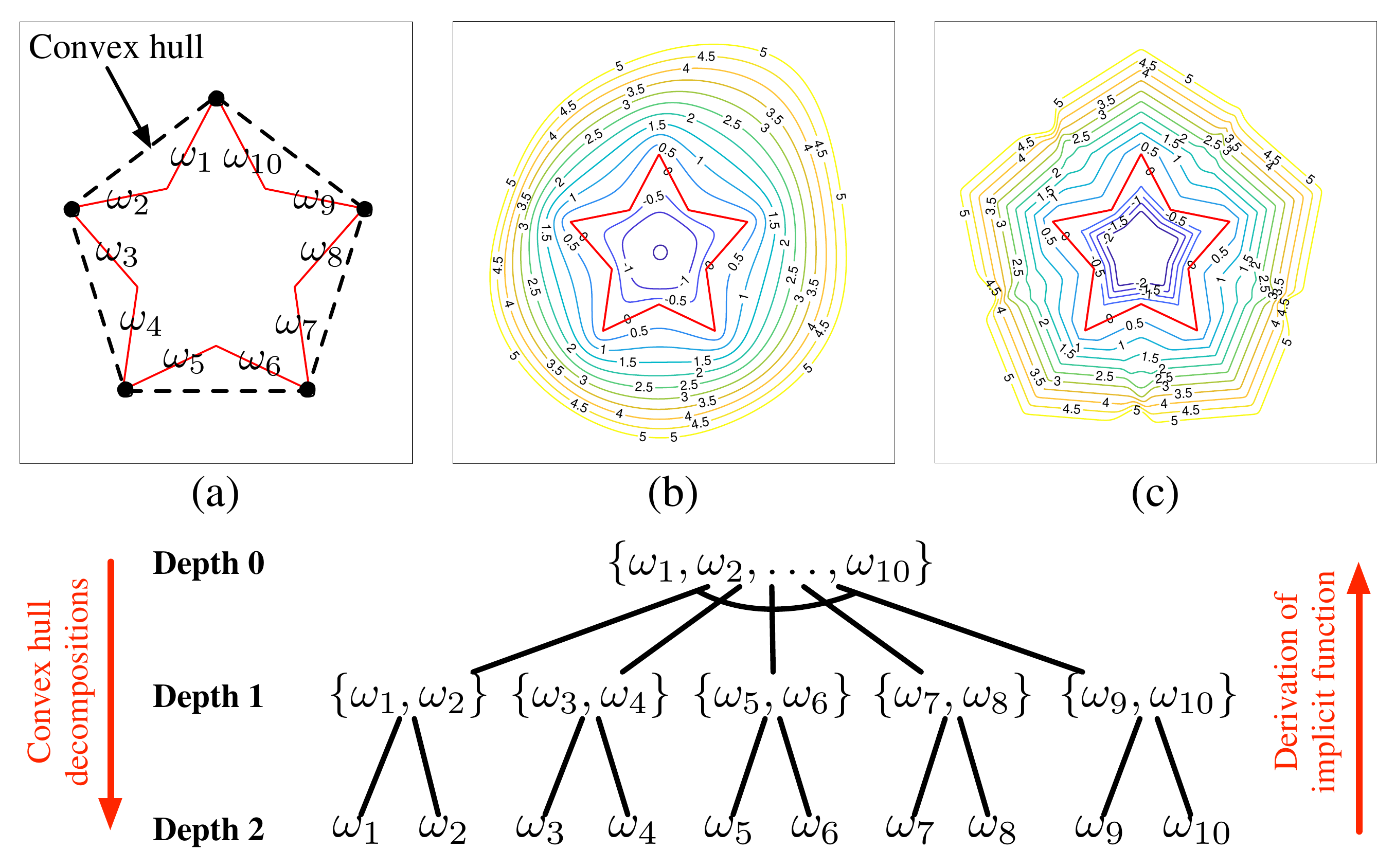}
\caption{Top: (a) An example of a star-shaped polygonal obstacle and the corresponding $\omega_j$ functions, (b) Level curves of the corresponding implicit function $\beta$ for $p=2$, (c) Level curves of the corresponding implicit function $\beta$ for $p=20$, Bottom: The AND-OR tree, constructed by the algorithm described in Appendix \ref{subsec:rfunction_algorithm} to represent this polygon. The polygon is split at the vertices of the convex hull to generate five subchains at depth 1. Each of these subchains is then split into two subchains at depth 2. The subchains at depth 2 (1) are combined via disjunction (conjunction), since they meet at non-convex (convex) vertices of the original polygon. In this way, we get our implicit function $\beta = \neg \left( (\omega_1 \vee \omega_2) \wedge (\omega_3 \vee \omega_4) \wedge (\omega_5 \vee \omega_6) \wedge (\omega_7 \vee \omega_8) \wedge (\omega_9 \vee \omega_{10}) \right)$.} \label{fig:rfunctions}
\end{figure}

In this work, looking ahead toward handling in a more modular fashion the general class of obstacle shapes encompassed by the star-tree methods from the traditional navigation function literature \cite{Rimon_Koditschek_1989,rimon1992}, we depart from individuated homogeneous implicit function representation of our memorized catalogue elements in favor of the R-function compositions \cite{Rvachev_1963}, explored by Rimon \cite{Rimon_1990} and explicated within the field of constructive solid geometry by Shapiro \cite{shapiro2007}. We believe that this modular representation of shape will be helpful in the effort now in progress to instantiate the posited mapping oracle for obstacles with known geometry, whose triangular mesh can be identified in real time using state-of-the-art techniques \cite{Kong_Lin_Lucey_2017,Kar_Tulsiani_Carreira_Malik_2015,Pavlakos2017} in order to extract implicit function representations for polygonal obstacles.

\subsubsection{Preliminary Definitions}
We begin by providing a definition of an R-function \cite{shapiro2007}.
\begin{definition}
A function $\gamma_\Phi:\mathbb{R}^n \rightarrow \mathbb{R}$ is an R-function if there exists a (binary) logic function $\Phi:\mathbb{B} \rightarrow \mathbb{B}$, called the companion function, that satisfies the relation
\begin{equation}
\Phi(S_2(w_1),\ldots,S_2(w_n)) = S_2(\gamma_\Phi(w_1,\ldots,w_n))
\end{equation}
with $(w_1,\ldots,w_n) \in \mathbb{R}^n$ and $S_2$ the Heaviside characteristic function $S_2:\mathbb{R} \rightarrow \mathbb{B}$ of the interval $[0+,\infty)$ defined as\footnote{In \cite{shapiro2007}, it is assumed that zero is always signed: either $+0$ or $-0$, which allows the authors to determine membership of zero either to the set of positive or to the set of negative numbers. This assumption is employed to resolve pathological cases, where the membership of zero causes R-function discontinuities and is not of particular importance in our setting.}
\begin{equation}
S_2(\chi) = \left\{ \begin{matrix}
0, \quad \chi \leq -0 \\ 1, \quad \chi \geq +0
\end{matrix} \right.
\end{equation}
\end{definition}
Informally, a real function $\gamma_\Phi$ is an R-function if it can change its property (sign) only when some of its arguments change the same property (sign) \cite{shapiro2007}. For example, the companion logic function for the R-function $\gamma(x,y)=x y$ is $X \Leftrightarrow Y$; we just check that $S_2(xy) = (S_2(x) \Leftrightarrow S_2(y))$.

In this work, we use the following (symbolically written) R-functions \cite{shapiro2007}
\begin{align}
\neg x & :=-x \label{eq:rfunction_negation} \\
x_1 \wedge x_2 & := x_1+x_2-\left(x_1^p+x_2^p\right)^\frac{1}{p} \label{eq:rfunction_conjunction} \\
x_1 \vee x_2 & := x_1+x_2+\left(x_1^p+x_2^p\right)^\frac{1}{p} \label{eq:rfunction_disjunction}
\end{align}
with companion logic functions the logical negation $\neg$, conjunction $\wedge$ and disjunction $\vee$ respectively and $p$ a positive integer. Intuitively, the author in \cite{shapiro2007} uses the triangle inequality with the $L_p$-norm to derive R-functions with specific properties. 

\subsubsection{Description of the Algorithm}
\label{subsec:rfunction_algorithm}
R-functions have several interesting properties but, most importantly, provide machinery to construct implicit representations for sets built from other, primitive sets. Namely, in order to obtain a real function inequality $\gamma \geq 0$ defining a set $\Omega$ constructed from primitive sets $\Omega_j$, it suffices to construct an appropriate R-function and substitute for its arguments the real functions $\omega_j$ defining the primitive sets $\Omega_j$ implicitly as $\omega_j \geq 0$ \cite[Theorem 3]{shapiro2007}. In our case, the set $\Omega$ would be a polygon $\knownobstacledilated_i$ we want to represent, the sets $\Omega_j$ would be half-spaces induced by the polygon edges, and the functions $\omega_j:\mathbb{R}^2 \rightarrow \mathbb{R}$ their corresponding hyperplane equations, given by
\begin{equation}
\omega_j(\mathbf{x}) = (\mathbf{x}-\mathbf{x}_{j})^\top \mathbf{n}_j \label{eq:hyperplane}
\end{equation}
Here $\robotposition_{j}$ is any arbitrary point on the edge hyperplane and $\mathbf{n}_j$ its normal vector, pointing towards the polygon's interior.

% \begin{algorithm}[H]
% \SetAlgoLined
% \SetKwData{stack}{stack}
% \SetKwData{nodes}{nodes}
% \SetKwData{expanded}{expanded}
% \SetKwData{subchains}{subchains}
% \SetKwData{e}{e}
% \SetKwData{temp}{temp}
% \SetKwData{temp}{temp}
% \SetKwFunction{pop}{pop}
% \SetKwFunction{split}{split}
% \SetKwFunction{append}{append}
% \SetKwInOut{Input}{Input}
% \SetKwInOut{Output}{Output}
% \Input{Array $P$ of polygon vertices, Point $\mathbf{x}$}
% \Output{Value $\beta(\mathbf{x})$}
%  \tcp{Initialize nodes of the tree: each node has a vertices, index, predecessor, depth and $\beta(\mathbf{x})$ field, in that order}
%  \nodes $\leftarrow \{P,1,0,0,\emptyset\}$\;
%  \stack $\leftarrow$ \nodes \;
%  \While{\stack $\neq \emptyset$}{
%   \expanded $\leftarrow$ \pop(\stack) \tcp*[r]{node of the tree to be expanded}
%   \subchains $\leftarrow$ \split(\expanded) \tcp*[r]{split at the convex hull}
%   \tcp{Append the subchains in the tree}
%   \ForEach{element \e of \subchains}{
%    \temp.vertices $\leftarrow$ \e.vertices\;
%    \temp.index $\leftarrow$ $\max$(\nodes.index)$+1$\;
%    \temp.predecessor $\leftarrow$ \expanded.index\;
%    \temp.depth $\leftarrow$ \expanded.depth$+1$\;
%    \temp.beta $\leftarrow \emptyset$\;
%    \append(\nodes,\temp)\;
%    \append(\stack,\temp)\;
%   }
%  }
%  \caption{Evaluating $\beta(\mathbf{x})$ at $\mathbf{x}$ from a list $P$ of polygon vertices} \label{algorithm:r_functions}
% \end{algorithm}

This result allows us to use a variant of the method presented in \cite{shapiro2007} and construct representations of polygons in the form of AND-OR trees \cite{russell_2009}, as shown in the example of Fig. \ref{fig:rfunctions}. Briefly, the interior of a polygon can be represented as the intersection of two or more {\it polygonal chains}, i.e. sequences of edges that meet at the polygon's convex hull. In the same way, each of these chains can then be split recursively into smaller subchains at the vertices of its convex hull to form a tree structure. The root node of the tree is the original polygon, with each other node corresponding to a polygonal chain; the leaves of the tree are single hyperplanes, the edges of the polygon described by functions $\omega_j$. If the split occurs at a concave vertex of the {\it original} polygon, then the subchains are combined using set union (i.e. disjunction); otherwise, they are combined using set intersection (i.e. conjunction), as shown in Fig. \ref{fig:rfunctions}. In this way, by having as input just the vertices of the polygon in counterclockwise order, we are able to construct an implicit representation for each node of the tree bottom-up, using the R-functions \eqref{eq:rfunction_conjunction} and \eqref{eq:rfunction_disjunction}, until we reach the root node of the tree. If we want $\beta_i > 0$ in the exterior of $\knownobstacledilated_i$, we can negate the result (i.e., we use the R-function \eqref{eq:rfunction_negation}) to obtain the function $\beta_i$, which is analytic everywhere except for the polygon vertices \cite{shapiro2007}. This is the reason our results in Section \ref{sec:diffeomorphism} still hold, with the map $\diffeo$ being a $C^\infty$ diffeomorphism away from the polygon vertices.

% We apply the algorithm described above to extract (offline) the AND-OR tree associated with the implicit function $\beta_{0i}$, describing the obstacle $\tilde{O}_i^*$ with its center located at the origin, as shown in Fig. \ref{fig:rfunctions}. Then, when our sensor recognizes $\tilde{O}_i^*$ online and identifies its pose as a rotation $\mathbf{R}_i$ on the plane followed by a translation $\diffeocenter{i}$ of its center, we simply find the value of $\beta_i(\mathbf{x})$ at a point $\mathbf{x}$ as $\beta_i(\mathbf{x}) = \beta_{0i}\left(\mathbf{R}_i^\top(\mathbf{x}-\diffeocenter{i})\right)$, by simply invoking the inverse homogeneous transformation that takes the center of $\tilde{O}_i^*$ back to the origin. This allows for efficient online computation of $\beta_i(\mathbf{x})$ as the robot navigates its environment.

\subsubsection{R-functions as Approximations of the Distance Function}
It is important to mention that, away from the corners and in a neighborhood of the polygon, {\it normalized} R-functions constructed using \eqref{eq:rfunction_negation}-\eqref{eq:rfunction_disjunction} behave as smooth $p$-th order approximations of the (non-differentiable) distance function to the polygon, as shown in Fig. \ref{fig:rfunctions}-(b),(c). The reader is referred to \cite{shapiro2007} for more details; in our setting, a sufficient condition for normalization is to make sure that for each $\omega_j$ given in \eqref{eq:hyperplane}, the corresponding normal vector $\mathbf{n}_j$ has unit norm \cite{shapiro2007}. This property is quite useful for our purposes, as it endows the implicit representation of our polygons with a physical meaning, compared to other representations (e.g., the homogeneous function representations in \cite{Rimon_Koditschek_1989}). Numerical experimentation showed that even $p=2$ gives sufficiently good results in our setting.

\subsection{Construction of Polygonal Collars}
\label{appendix:computational_geometry}

Definitions \ref{definition:collars}, \ref{definition:collars_root_disk} and \ref{definition:collars_root_boundary} provide the basic guidelines for constructing admissible polygonal collars that fit our formal results. However, there is not a unique way of performing this operation. Here, we describe the method employed in this paper for a single polygon $\knownobstacledilated$, contained in either $\knownobstaclesetdilatedmappeddisk$ or $\knownobstaclesetdilatedmappedintrusion$, whose triangulation tree $\triangletree{\knownobstacledilated}:=(\trianglevertices{\knownobstacledilated},\triangleedges{\knownobstacledilated})$ has already been constructed, according to Section \ref{subsec:obstacle_representation}, and the corresponding centers of transformation $\diffeocenter{j}$ have already been identified, according to Definitions \ref{definition:center}, \ref{definition:centers_root_disk} and \ref{definition:centers_root_boundary} for all triangles $j \in \trianglevertices{\knownobstacledilated}$. We assume that the value of the corresponding clearance $\betaclearance_{\knownobstacledilated}$, according to Assumption \ref{assumption:beta}, is also known.

\begin{algorithm}[H]
\begin{algorithmic}
\Function{CollarConstruction}{$\knownobstacledilated, \betaclearance_{\knownobstacledilated}$}
\State $\trianglevertices{\knownobstacledilated} \gets \texttt{sort}(\trianglevertices{\knownobstacledilated})$ \Comment{Sort in descending depth}
\Do
\State $j \gets \texttt{pop}(\trianglevertices{\knownobstacledilated})$ \Comment{Pop next triangle}
\If{$j$ is $\texttt{root}$ and $\knownobstacledilated \in \knownobstaclesetdilatedmappeddisk$}
\State $\innerpolygon{j} \gets \robotposition_{1j}\robotposition_{2j}\robotposition_{3j}\robotposition_{1j}$
\State $\mathcal{A}_j \gets \texttt{dilate}(\innerpolygon{j},\betaclearance_{\knownobstacledilated})$ \Comment{Dilate $\innerpolygon{j}$ by $\betaclearance_{\knownobstacledilated}$}
\State $\outerpolygon{j} \gets \mathcal{A}_j \cap \freespacemappedpurging{j}$
\ElsIf{$j$ is $\texttt{root}$ and $\knownobstacledilated \in \knownobstaclesetdilatedmappedintrusion$}
\State $\innerpolygon{j} \gets \diffeocenter{j}\robotposition_{2j}\robotposition_{3j}\robotposition_{1j}\diffeocenter{j}$
\State $\mathcal{A}_j \gets \texttt{dilate}(\innerpolygon{j},\betaclearance_{\knownobstacledilated})$ \Comment{Dilate $\innerpolygon{j}$ by $\betaclearance_{\knownobstacledilated}$}
\State $\mathcal{R}_j \gets (\mathcal{A}_j \cap \freespacemappedpurging{j}) \cap (H_{1j} \cap H_{2j})$
\State $\outerpolygon{j} \gets \mathcal{R}_j \cup \diffeocenter{j}\robotposition_{2j}\robotposition_{1j}\diffeocenter{j}$
\Else
\State $\innerpolygon{j} \gets \diffeocenter{j}\robotposition_{2j}\robotposition_{3j}\robotposition_{1j}\diffeocenter{j}$
\State $\mathcal{A}_j \gets \texttt{dilate}(\innerpolygon{j},\betaclearance_{\knownobstacledilated})$ \Comment{Dilate $\innerpolygon{j}$ by $\betaclearance_{\knownobstacledilated}$}
\State $\mathcal{R}_j \gets (\mathcal{A}_j \cap \freespacemappedpurging{j}) \cap (H_{1j} \cap H_{2j})$
\State $\mathcal{L}_j \gets$ List of triangles that will succeed $j$
\Do
\State $i \gets \texttt{pop}(\mathcal{L}_j)$ \Comment{Pop next triangle}
\If{$i$ is $p(j)$}
\State $\textbf{continue}$
\Else
\State $\mathcal{R}_j \gets \mathcal{R}_j - i$ \Comment{Polygon difference}
\EndIf
\doWhile{$\mathcal{L}_j \neq \varnothing$}
\State $\{\mathcal{Z}\}_k \gets \texttt{poly\_decomp}(\mathcal{R}_j)$ \Comment{\cite{keil-convex-decomposition}}
\State $\outerpolygon{j} \gets \mathcal{Z}_k$ such that $\innerpolygon{j} \subset \mathcal{Z}_k$
\EndIf
\doWhile{$\trianglevertices{\knownobstacledilated} \neq \varnothing$}
\EndFunction
\end{algorithmic}
\caption{Construction of the polygonal collars $\outerpolygon{j}$ for all triangles $j \in \trianglevertices{\knownobstacledilated}$ of a polygon $\knownobstacledilated$, whose triangulation tree $\triangletree{\knownobstacledilated}:=(\trianglevertices{\knownobstacledilated},\triangleedges{\knownobstacledilated})$ and associated clearance $\betaclearance_{\knownobstacledilated}$ are known.} \label{algorithm:collars}
\end{algorithm}

\subsection{Inductive Computation of the Diffeomorphism at Execution Time}
\label{appendix:calculation_jacobian}

From the description of the diffeomorphism $\diffeo$ in Section \ref{sec:diffeomorphism}, we see that $\diffeo$ is constructed in multiple steps by composition. Therefore, we can compute the value of $\diffeo(\robotposition)$ at $\robotposition \in \freespacemapped$ inductively, by setting $\diffeo_0(\robotposition) = \robotposition$ and computing $\diffeo_k(\robotposition) = \diffeo_{k,k-1} \circ \diffeo_{k-1}(\robotposition)$, with $k$ spanning all triangles in $\trianglevertices{\knownobstacledilated}$ for all known obstacles $\knownobstacledilated$ in both $\knownobstaclesetdilatedmappeddisk$ and $\knownobstaclesetdilatedmappedintrusion$, and $\diffeo_{k,k-1}$ given either in \eqref{eq:map_purging} or \eqref{eq:map_root}. We can then see that, due to Lemma \ref{lemma:switches_nonzero}, $\diffeo_{k,k-1}$ can be generally written in the following form
\begin{align}
    \diffeo_{k,k-1}(\robotposition) = & \switch{k,k-1}(\robotposition)\left[ \diffeocenter{k,k-1} + \deformingfactor{k,k-1}(\robotposition)(\robotposition-\diffeocenter{k,k-1})\right] \nonumber \\
    & + \left(1 - \switch{k,k-1}(\robotposition)\right)\robotposition
\end{align}
with the switch $\switch{k,k-1}$ (see \eqref{eq:sigma_ji}, \eqref{eq:sigma_ri}), deforming factor $\deformingfactor{k,k-1}$ (see \eqref{eq:deforming_factor_purging}, \eqref{eq:deforming_factor_disk}, \eqref{eq:deforming_factor_root_boundary}) and center of the transformation $\diffeocenter{k,k-1}$ (see Definitions \ref{definition:center}, \ref{definition:centers_root_disk}, \ref{definition:collars_root_boundary}) depending on the particular triangle being purged.

We can, therefore, set $D_\robotposition \diffeo_0 := \mathbf{I}$, compute
\begin{align}
    D_\robotposition \diffeo_{k,k-1} = & \left(\deformingfactor{k,k-1}(\robotposition)-1\right)(\robotposition-\diffeocenter{k,k-1}) \nabla \switch{k,k-1}(\robotposition)^\top \nonumber \\
    & + \switch{k,k-1}(\robotposition)(\robotposition-\diffeocenter{k,k-1}) \nabla \deformingfactor{k,k-1}(\robotposition)^\top \nonumber \\
    & + \left[1 + \switch{k,k-1}(\robotposition)\left(\deformingfactor{k,k-1}(\robotposition)-1\right)\right]\mathbf{I} \label{eq:inductive_jacobian}
\end{align}
and use the chain rule to write
\begin{equation}
    D_\robotposition \diffeo_k = \left(D_\robotposition \diffeo_{k,k-1} \circ \diffeo_{k-1}(\robotposition)\right) \cdot D_\robotposition \diffeo_{k-1}
\end{equation}

Finally, since \eqref{eq:dksidx} requires partial derivatives of $D_\robotposition \diffeo$, we can follow a similar procedure and the chain rule to compute the partial derivatives $\frac{\partial [D_\robotposition \diffeo_k]_{ml}}{\partial [\robotposition]_n}$, as functions of $\frac{\partial [D_\robotposition \diffeo_{k-1}]_{ml}}{\partial [\robotposition]_n}$ and $\frac{\partial [D_\robotposition \diffeo_{k,k-1}]_{ml}}{\partial [\robotposition]_n} \Big |_{\diffeo_{k-1}(\robotposition)}$, after initially setting all partial derivatives to zero: $\frac{\partial [D_\robotposition \diffeo_0]_{ml}}{\partial [\robotposition]_n} = 0$, with the indices $m,l,n \in \{1,2\}$. Namely:
\begin{align}
    \frac{\partial [D_\robotposition \diffeo_k]_{ml}}{\partial [\robotposition]_n} = & \sum_{r=1}^2 \left( \left[D_\robotposition \diffeo_{k,k-1} \circ \diffeo_{k-1}(\robotposition)\right]_{mr} \cdot \right. \nonumber \\
    & \cdot \frac{\partial [D_\robotposition \diffeo_{k-1}]_{rl}}{\partial [\robotposition]_n} \nonumber \\
    & + [D_\robotposition \diffeo_{k-1}]_{rl} \sum_{s=1}^2 [D_\robotposition \diffeo_{k-1}]_{sn} \cdot \nonumber \\
    & \left. \cdot \frac{\partial \left[D_\robotposition \diffeo_{k,k-1}\right]_{mr}}{\partial [\robotposition]_s} \Big |_{\diffeo_{k-1}(\robotposition)} \right) 
\end{align}
where, from \eqref{eq:inductive_jacobian}, we can compute
\begin{align}
    & \frac{\partial \left[D_\robotposition \diffeo_{k,k-1}\right]_{mr}}{\partial [\robotposition]_s} = (\deformingfactor{k,k-1}-1) \frac{\partial \switch{k,k-1}}{\partial [\robotposition]_r} \delta_{ms} \nonumber \\
    & + ([\robotposition]_m - [\diffeocenter{k,k-1}]_{m}) \frac{\partial \switch{k,k-1}}{\partial [\robotposition]_r} \frac{\partial \deformingfactor{k,k-1}}{\partial [\robotposition]_s} \nonumber \\
    & + (\deformingfactor{k,k-1} - 1) ([\robotposition]_m - [\diffeocenter{k,k-1}]_{m}) \frac{\partial^2 \switch{k,k-1}}{\partial [\robotposition]_r \partial [\robotposition]_s} \nonumber \\
    & + ([\robotposition]_m - [\diffeocenter{k,k-1}]_{m}) \frac{\partial \switch{k,k-1}}{\partial [\robotposition]_s} \frac{\partial \deformingfactor{k,k-1}}{\partial [\robotposition]_r} \nonumber \\
    & + \switch{k,k-1} \frac{\partial \deformingfactor{k,k-1}}{\partial [\robotposition]_r} \delta_{ms} \nonumber \\
    & + \switch{k,k-1} ([\robotposition]_m - [\diffeocenter{k,k-1}]_{m}) \frac{\partial^2 \deformingfactor{k,k-1}}{\partial [\robotposition]_r \partial [\robotposition]_s} \nonumber \\
    & + \switch{k,k-1} \frac{\partial \deformingfactor{k,k-1}}{\partial [\robotposition]_s} \delta_{mr} + (\deformingfactor{k,k-1}-1) \frac{\partial \switch{k,k-1}}{\partial [\robotposition]_s} \delta_{mr}
\end{align}
by using elements of the Hessians $\nabla^2 \switch{k,k-1}, \nabla^2 \deformingfactor{k,k-1}$.

\subsection{Proofs}
\label{appendix:proofs}

\subsubsection{Proofs of Results in Section \ref{sec:diffeomorphism}}
\label{appendix:proofs_diffeo}
\begin{proof}[Lemma \ref{lemma:singular_leaf}]
With the procedure outlined in Appendix \ref{appendix:implicit}, the only points where $\innerpolygonimplicit{j_i}$ and $\outerpolygonimplicit{j_i}$ are not smooth are vertices of $\innerpolygon{j_i}$ and $\outerpolygon{j_i}$ respectively. Therefore, with the definition of $\outerpolygonsigma{j_i}$ as in \eqref{eq:sigma_delta_ji} and the use of the smooth, non-analytic function $\zeta$ from \eqref{eq:zeta}, we see that $\outerpolygonsigma{j_i}$ is smooth everywhere, since $\diffeocenter{j_i}$ does not belong in $\freespacemappedpurging{j_i}$ and $\outerpolygonimplicit{j_i}$ is exactly 0 on the vertices of $\outerpolygon{j_i}$. Therefore, $\switch{j_i}$ can only be non-smooth on the vertices of $\innerpolygon{j_i}$ except for $\diffeocenter{j_i}$ (i.e., on the vertices of the triangle $j_i$), and on points where its denominator becomes zero. Since both $\innerpolygonsigma{j_i}$ and $\outerpolygonsigma{j_i}$ vary between 0 and 1, this can only happen when $\innerpolygonsigma{j_i} (\robotposition) = 1$ and $\outerpolygonsigma{j_i}(\robotposition) = 0$, i.e., only on $\robotposition_{1j_i}$ and $\robotposition_{2j_i}$. The fact that $\switch{j_i}$ is smooth everywhere else derives immediately from the fact that $\outerpolygonsigma{j_i}$ is a smooth function, and $\innerpolygonsigma{j_i}$ is smooth everywhere except for the triangle vertices. 
\end{proof}

\begin{proof}[Lemma \ref{lemma:purging_smooth}]
From Lemma \ref{lemma:singular_leaf}, we already know that the switch $\switch{j_i}$ is smooth away from the vertices of $j_i$. On the other hand, the singular points of the deforming factor $\deformingfactor{j_i}$ are the solutions of the equation $(\robotposition-\diffeocenter{j_i})^\top \sharednormal{j_i} = 0$ and, therefore, lie on the hyperplane passing through $\diffeocenter{j_i}$ with normal vector $\sharednormal{j_i}$ and, due to the construction of $\outerpolygon{j_i}$ as in Definition~\ref{definition:collars}, lie outside of $\outerpolygon{j_i}$ and do not affect the map $\freespacemappedpurging{j_i}$. Hence, the map $\diffeopurging{j_i}$ is smooth everywhere in $\freespacemappedpurging{j_i}$, except for the vertices of the triangle $j_i$, as a composition of smooth functions with the same properties.
\end{proof}

\begin{proof}[Proposition \ref{proposition:diffeo_purging}]
First of all, the map $\diffeopurging{j_i}$ is smooth everywhere except for the vertices of the triangle $j_i$, as shown in Lemma \ref{lemma:purging_smooth}. Therefore, in order to prove that $\diffeopurging{j_i}$ is a $C^\infty$ diffeomorphism away from the triangle vertices $\robotposition_{1j_i},\robotposition_{2j_i},\robotposition_{3j_i}$, we follow the procedure outlined in \cite{massey1992}, also followed in \cite{Rimon_Koditschek_1989}, to show that
\begin{enumerate}
    \item $\diffeopurging{j_i}$ has a non-singular differential on $\freespacemappedpurging{j_i}$ except for $\robotposition_{1j_i},\robotposition_{2j_i},\robotposition_{3j_i}$.
    \item $\diffeopurging{j_i}$ preserves boundaries, i.e., $\diffeopurging{j_i}(\partial_k\freespacemappedpurging{j_i}) \subset \partial_k\freespacemappedpurging{p(j_i)}$, with $k$ spanning both the indices of familiar obstacles $\knownobstaclesetdilatedmappeddiskindex$, $\knownobstaclesetdilatedmappedintrusionindex$ as well as the indices of unknown obstacles $\unknownobstaclesetdilatedsemanticindex$, and $\partial_k \freespace$ the $k$-th connected component of the boundary of $\freespace$ with $\partial_0\freespace$ the outer boundary of $\freespace$.
    \item the boundary components of $\freespacemappedpurging{j_i}$ and $\freespacemappedpurging{p(j_i)}$ are pairwise homeomorphic, i.e., $\partial_k\freespacemappedpurging{j_i} \cong \partial_k \freespacemappedpurging{p(j_i)}$, with $k$ spanning both the indices of familiar obstacles $\knownobstaclesetdilatedmappeddiskindex$, $\knownobstaclesetdilatedmappedintrusionindex$ as well as the indices of unknown obstacles $\unknownobstaclesetdilatedsemanticindex$.
\end{enumerate}
We begin with Property 1 and examine the space away from the triangle vertices $\robotposition_{1j_i},\robotposition_{2j_i},\robotposition_{3j_i}$. The case where $\outerpolygonsigma{j_i}$ is 0 (outside of the polygonal collar $\outerpolygon{j_i}$) is not interesting, since $\diffeopurging{j_i}$ defaults to the identity map and $D_\robotposition\diffeopurging{j_i} = \mathbf{I}$. When $\outerpolygonsigma{j_i}$ is not 0, we can compute the jacobian of the map as
\begin{align}
    D_\robotposition\diffeopurging{j_i} = & \left(\deformingfactor{j_i}(\robotposition)-1\right) (\robotposition-\diffeocenter{j_i}) \nabla \switch{j_i}(\robotposition)^\top \nonumber \\
    & + \switch{j_i}(\robotposition)(\robotposition-\diffeocenter{j_i}) \nabla \deformingfactor{j_i}(\robotposition)^\top \nonumber \\
    & + \left[ 1 + \switch{j_i}(\robotposition)\left(\deformingfactor{j_i}(\robotposition)-1\right) \right] \mathbf{I} \label{eq:jacobian_purging}
\end{align}
For the deforming factor $\deformingfactor{j_i}$ we compute from \eqref{eq:deforming_factor_purging}
\begin{equation}
    \nabla \deformingfactor{j_i}(\robotposition) = -\frac{\left(\robotposition_{1j_i} - \diffeocenter{j_i} \right) ^\top \sharednormal{j_i}}{\left[\left(\robotposition - \diffeocenter{j_i} \right) ^\top \sharednormal{j_i}\right]^2} \sharednormal{j_i}
\end{equation}
Note that we interestingly get
\begin{equation}
    \left(\robotposition-\diffeocenter{j_i} \right)^\top \nabla \deformingfactor{j_i}(\robotposition) = -\deformingfactor{j_i}(\robotposition) \label{eq:nu_inner_purging}
\end{equation}
From \eqref{eq:jacobian_purging} it can be seen that $D_\robotposition\diffeopurging{j_i} = \mathbf{A} + \mathbf{u}\mathbf{v}^\top$ with $\mathbf{A} = \left[ 1 + \switch{j_i}(\robotposition)\left(\deformingfactor{j_i}(\robotposition)-1\right) \right] \mathbf{I}$, $\mathbf{u} = \robotposition-\diffeocenter{j_i}$ and $\mathbf{v} = \left(\deformingfactor{j_i}(\robotposition)-1\right)\nabla \switch{j_i}(\robotposition) + \switch{j_i}(\robotposition) \nabla \deformingfactor{j_i}(\robotposition)$.

Due to the fact that $0 \leq \switch{j_i}(\robotposition) \leq 1$ and $0 < \deformingfactor{j_i}(\robotposition) < 1$ in the interior of an admissible polygonal collar $\outerpolygon{j_i}$ (see Definition \ref{definition:collars}), we get $1 + \switch{j_i}(\robotposition)\left(\deformingfactor{j_i}(\robotposition)-1\right) > 0$.
Hence, $\mathbf{A}$ is invertible, and by using the matrix determinant lemma and \eqref{eq:nu_inner_purging}, the determinant of $D_\robotposition\diffeopurging{j_i}$ can be computed as
\begin{align}
    & \text{det}(D_\robotposition\diffeopurging{j_i}) = \text{det}\mathbf{A} + (\text{det}\mathbf{A})\mathbf{v}^\top \mathbf{A}^{-1} \mathbf{u} \nonumber \\
    & = \left[ 1 + \switch{j_i}(\robotposition)\left(\deformingfactor{j_i}(\robotposition)-1\right) \right] \cdot \nonumber \\
    & \cdot \left[ \left(1-\switch{j_i}(\robotposition)\right) + \left(\deformingfactor{j_i}(\robotposition)-1\right)(\robotposition-\diffeocenter{j_i})^\top \nabla \switch{j_i}(\robotposition) \right]
\end{align}
Similarly the trace of $D_\robotposition\diffeopurging{j_i}$ can be computed as
\begin{align}
    & \text{tr}(D_\robotposition\diffeopurging{j_i}) = \left[ 1 + \switch{j_i}(\robotposition)\left(\deformingfactor{j_i}(\robotposition)-1\right) \right] + \left(1-\switch{j_i}(\robotposition)\right) \nonumber \\
    & + \left(\deformingfactor{j_i}(\robotposition)-1\right)(\robotposition-\diffeocenter{j_i})^\top \nabla \switch{j_i}(\robotposition)
\end{align}

Also, by construction of the switch $\switch{j_i}$, we see that $\nabla \switch{j_i}(\robotposition) = \mathbf{0}$ when $\switch{j_i}(\robotposition) = 0$. Hence, using the above expressions, we can show that $\text{det}(D_\robotposition\diffeopurging{j_i}), \text{tr}(D_\robotposition\diffeopurging{j_i}) > 0$ (and therefore establish that $D_\robotposition\diffeopurging{j_i}$ is not singular in the interior of $\outerpolygon{j_i}$, since $\freespacemappedpurging{j_i} \subseteq \mathbb{R}^2$) by showing that $(\robotposition-\diffeocenter{j_i})^\top \nabla \switch{j_i}(\robotposition) < 0$ when $\switch{j_i}(\robotposition) > 0$, where
\begin{align}
    \nabla \switch{j_i}(\robotposition) = & \frac{\outerpolygonsigma{j_i}(\robotposition) }{\left[ \innerpolygonsigma{j_i}(\robotposition) \outerpolygonsigma{j_i}(\robotposition) + \left(1- \innerpolygonsigma{j_i}(\robotposition) \right) \right]^2} \nabla \innerpolygonsigma{j_i}(\robotposition) \nonumber \\
    & + \frac{\innerpolygonsigma{j_i}(\robotposition) \left(1-\innerpolygonsigma{j_i}(\robotposition) \right)}{\left[ \innerpolygonsigma{j_i}(\robotposition) \outerpolygonsigma{j_i}(\robotposition) + \left(1- \innerpolygonsigma{j_i}(\robotposition) \right) \right]^2} \nabla \outerpolygonsigma{j_i}(\robotposition)
\end{align}
with
\begin{equation}
    \nabla \innerpolygonsigma{j_i}(\robotposition) = \left\{ \begin{matrix} -\dfrac{\innerpolygontune{j_i} \innerpolygonsigma{j_i}(\robotposition)}{\left(\innerpolygondistance{j_i} - \innerpolygonimplicit{j_i}(\robotposition) \right)^2} \nabla \innerpolygonimplicit{j_i}(\robotposition), & \innerpolygonimplicit{j_i}(\robotposition) < \innerpolygondistance{j_i} \\
    \mathbf{0}, & \innerpolygonimplicit{j_i}(\robotposition) \geq \innerpolygondistance{j_i} \end{matrix} \right.
\end{equation}
\begin{equation}
    \nabla \outerpolygonsigma{j_i}(\robotposition) = \left\{ \begin{matrix} \dfrac{\outerpolygontune{j_i} \outerpolygonsigma{j_i}(\robotposition)}{\alpha_{j_i}(\robotposition)^2} \nabla \alpha_{j_i}(\robotposition), & \outerpolygonimplicit{j_i}(\robotposition) > 0 \\
    \mathbf{0}, & \outerpolygonimplicit{j_i}(\robotposition) \leq 0 \end{matrix} \right.
\end{equation}
and $\alpha_{j_i}(\robotposition) := \outerpolygonimplicit{j_i}(\robotposition)/||\robotposition-\diffeocenter{j_i}||$. Therefore, it suffices to show that when $\switch{j_i}(\robotposition) > 0$:
\begin{align}
    (\robotposition-\diffeocenter{j_i})^\top \nabla \innerpolygonimplicit{j_i}(\robotposition) > 0 \label{eq:condition_gamma} \\
    (\robotposition-\diffeocenter{j_i})^\top \nabla \alpha_{j_i}(\robotposition) < 0 \label{eq:condition_delta}
\end{align}

Following the procedure outlined in Appendix \ref{appendix:implicit} for generating implicit functions for polygons, it can be seen that the implicit function $\innerpolygonimplicit{j_i}$, describing the exterior of the quadrilateral $\innerpolygon{j_i}$, can be written as
\begin{equation}
    \innerpolygonimplicit{j_i}(\robotposition) = \neg \left( (\innerpolygonimplicit{1j_i}(\robotposition) \wedge \innerpolygonimplicit{2j_i}(\robotposition)) \wedge (\innerpolygonimplicit{3j_i}(\robotposition) \wedge \innerpolygonimplicit{4j_i}(\robotposition)) \right)
\end{equation}
with the negation ($\neg$) and conjunction ($\wedge$) operators defined as in \eqref{eq:rfunction_negation} and \eqref{eq:rfunction_conjunction} respectively, and $\innerpolygonimplicit{1j_i}, \innerpolygonimplicit{2j_i}, \innerpolygonimplicit{3j_i}, \innerpolygonimplicit{4j_i}$ the hyperplane equations describing $\innerpolygon{j_i}$ defined as follows
\begin{align}
    \innerpolygonimplicit{1j_i}(\robotposition) := (\robotposition-\diffeocenter{j_i}) ^\top \sharednormal{1j_i}, \sharednormal{1j_i} := \mathbf{R}_{\frac{\pi}{2}} \tfrac{\diffeocenter{j_i}-\robotposition_{1j_i}}{||\diffeocenter{j_i}-\robotposition_{1j_i}||} \\
    \innerpolygonimplicit{2j_i}(\robotposition) := (\robotposition-\diffeocenter{j_i}) ^\top \sharednormal{2j_i}, \sharednormal{2j_i} := \mathbf{R}_{\frac{\pi}{2}} \tfrac{\robotposition_{2j_i}-\diffeocenter{j_i}}{||\robotposition_{2j_i}-\diffeocenter{j_i}||} \\
    \innerpolygonimplicit{3j_i}(\robotposition) := (\robotposition-\robotposition_{3j_i}) ^\top \sharednormal{3j_i}, \sharednormal{3j_i} := \mathbf{R}_{\frac{\pi}{2}} \tfrac{\robotposition_{3j_i}-\robotposition_{2j_i}}{||\robotposition_{3j_i}-\robotposition_{2j_i}||} \\
    \innerpolygonimplicit{4j_i}(\robotposition) := (\robotposition-\robotposition_{3j_i}) ^\top \sharednormal{4j_i}, \sharednormal{4j_i} := \mathbf{R}_{\frac{\pi}{2}} \tfrac{\robotposition_{1j_i}-\robotposition_{3j_i}}{||\robotposition_{1j_i}-\robotposition_{3j_i}||}
\end{align}
We therefore get
\begin{align}
    & \nabla \innerpolygonimplicit{j_i} = -\left(1 - \tfrac{\innerpolygonimplicit{1j_i} \wedge \innerpolygonimplicit{2j_i}}{\sqrt{(\innerpolygonimplicit{1j_i} \wedge \innerpolygonimplicit{2j_i})^2 + (\innerpolygonimplicit{3j_i} \wedge \innerpolygonimplicit{4j_i})^2}} \right) \nabla (\innerpolygonimplicit{1j_i} \wedge \innerpolygonimplicit{2j_i}) \nonumber \\
    & - \left(1 - \tfrac{\innerpolygonimplicit{3j_i} \wedge \innerpolygonimplicit{4j_i}}{\sqrt{(\innerpolygonimplicit{1j_i} \wedge \innerpolygonimplicit{2j_i})^2 + (\innerpolygonimplicit{3j_i} \wedge \innerpolygonimplicit{4j_i})^2}} \right) \nabla (\innerpolygonimplicit{3j_i} \wedge \innerpolygonimplicit{4j_i}) \label{eq:gamma_nabla}
\end{align}
with
\begin{align}
    \nabla (\innerpolygonimplicit{1j_i} \wedge \innerpolygonimplicit{2j_i}) = & \left(1 - \tfrac{\innerpolygonimplicit{1j_i}}{\sqrt{\innerpolygonimplicit{1j_i}^2 + \innerpolygonimplicit{2j_i}^2}} \right) \nabla \innerpolygonimplicit{1j_i} \nonumber \\
    & + \left(1 - \tfrac{\innerpolygonimplicit{2j_i}}{\sqrt{\innerpolygonimplicit{1j_i}^2 + \innerpolygonimplicit{2j_i}^2}} \right) \nabla \innerpolygonimplicit{2j_i} \nonumber \\
    = & \left(1 - \tfrac{\innerpolygonimplicit{1j_i}}{\sqrt{\innerpolygonimplicit{1j_i}^2 + \innerpolygonimplicit{2j_i}^2}} \right) \sharednormal{1j_i} \nonumber \\
    & + \left(1 - \tfrac{\innerpolygonimplicit{2j_i}}{\sqrt{\innerpolygonimplicit{1j_i}^2 + \innerpolygonimplicit{2j_i}^2}} \right) \sharednormal{2j_i} \\
    \nabla (\innerpolygonimplicit{3j_i} \wedge \innerpolygonimplicit{4j_i}) = & \left(1 - \tfrac{\innerpolygonimplicit{3j_i}}{\sqrt{\innerpolygonimplicit{3j_i}^2 + \innerpolygonimplicit{4j_i}^2}} \right) \nabla \innerpolygonimplicit{3j_i} \nonumber \\
    & + \left(1 - \tfrac{\innerpolygonimplicit{4j_i}}{\sqrt{\innerpolygonimplicit{3j_i}^2 + \innerpolygonimplicit{4j_i}^2}} \right) \nabla \innerpolygonimplicit{4j_i} \nonumber \\
    = & \left(1 - \tfrac{\innerpolygonimplicit{3j_i}}{\sqrt{\innerpolygonimplicit{3j_i}^2 + \innerpolygonimplicit{4j_i}^2}} \right) \sharednormal{3j_i} \nonumber \\
    & + \left(1 - \tfrac{\innerpolygonimplicit{4j_i}}{\sqrt{\innerpolygonimplicit{3j_i}^2 + \innerpolygonimplicit{4j_i}^2}} \right) \sharednormal{4j_i}
\end{align}
It is then not hard to show that $(\robotposition-\diffeocenter{j_i}) ^\top \nabla (\innerpolygonimplicit{1j_i} \wedge \innerpolygonimplicit{2j_i}) = \innerpolygonimplicit{1j_i} \wedge \innerpolygonimplicit{2j_i}$. The term corresponding to $(\innerpolygonimplicit{3j_i} \wedge \innerpolygonimplicit{4j_i})$ is more complicated, but we can follow a similar procedure to get
\begin{align}
    & (\robotposition-\diffeocenter{j_i}) ^\top \nabla (\innerpolygonimplicit{3j_i} \wedge \innerpolygonimplicit{4j_i}) =  \innerpolygonimplicit{3j_i} \wedge \innerpolygonimplicit{4j_i} \nonumber \\
    & - \left(1 - \tfrac{\innerpolygonimplicit{3j_i}}{\sqrt{\innerpolygonimplicit{3j_i}^2 + \innerpolygonimplicit{4j_i}^2}} \right) (\diffeocenter{j_i}-\robotposition_{3j_i})^\top \sharednormal{3j_i} \nonumber \\
    & - \left(1 - \tfrac{\innerpolygonimplicit{4j_i}}{\sqrt{\innerpolygonimplicit{3j_i}^2 + \innerpolygonimplicit{4j_i}^2}} \right) (\diffeocenter{j_i}-\robotposition_{3j_i})^\top \sharednormal{4j_i} \nonumber \\
    & < \innerpolygonimplicit{3j_i} \wedge \innerpolygonimplicit{4j_i}
\end{align}
since $(\diffeocenter{j_i}-\robotposition_{3j_i})^\top \sharednormal{3j_i} > 0$ and $(\diffeocenter{j_i}-\robotposition_{3j_i})^\top \sharednormal{4j_i} > 0$, because $\innerpolygon{j_i}$ is convex. Therefore, using the facts that $(\robotposition-\diffeocenter{j_i}) ^\top \nabla (\innerpolygonimplicit{1j_i} \wedge \innerpolygonimplicit{2j_i}) = \innerpolygonimplicit{1j_i} \wedge \innerpolygonimplicit{2j_i}$ and $(\robotposition-\diffeocenter{j_i}) ^\top \nabla (\innerpolygonimplicit{3j_i} \wedge \innerpolygonimplicit{4j_i}) < \innerpolygonimplicit{3j_i} \wedge \innerpolygonimplicit{4j_i}$, we can get the desired result using \eqref{eq:gamma_nabla} as follows
\begin{align}
    (\robotposition-\diffeocenter{j_i}) ^\top \nabla \innerpolygonimplicit{j_i}(\robotposition) > & -((\innerpolygonimplicit{1j_i} \wedge \innerpolygonimplicit{2j_i}) \wedge (\innerpolygonimplicit{3j_i} \wedge \innerpolygonimplicit{4j_i})) \nonumber \\
    = & \neg ((\innerpolygonimplicit{1j_i} \wedge \innerpolygonimplicit{2j_i}) \wedge (\innerpolygonimplicit{3j_i} \wedge \innerpolygonimplicit{4j_i})) \nonumber \\
    = & \innerpolygonimplicit{j_i}(\robotposition) > 0
\end{align}

The proof of \eqref{eq:condition_delta} follows similar patterns. Here, we focus on $\outerpolygonimplicit{j_i}$. The external polygonal collar $\outerpolygon{j_i}$ can be assumed to have $n$ sides, which means that we can write $\outerpolygonimplicit{j_i} = \left( (\outerpolygonimplicit{1j_i}\wedge \outerpolygonimplicit{2j_i}) \wedge \ldots \wedge \outerpolygonimplicit{nj_i} \right)$. Following the procedure outlined above for the proof of \eqref{eq:condition_gamma}, we can expand each term in the conjunction individually and then combine them to get
\begin{equation}
    (\robotposition-\diffeocenter{j_i}) ^\top \nabla \outerpolygonimplicit{j_i} (\robotposition) < \outerpolygonimplicit{j_i} (\robotposition) \label{eq:delta_inner}
\end{equation}
We also have
\begin{align}
    \nabla \alpha_{j_i}(\robotposition) & = \nabla \left( \frac{\outerpolygonimplicit{j_i}(\robotposition)}{||\robotposition-\diffeocenter{j_i}||} \right) \nonumber \\
    & = \frac{||\robotposition-\diffeocenter{j_i}|| \nabla \outerpolygonimplicit{j_i}(\robotposition) - \outerpolygonimplicit{j_i}(\robotposition)\tfrac{\robotposition-\diffeocenter{j_i}}{||\robotposition-\diffeocenter{j_i}||}}{||\robotposition-\diffeocenter{j_i}||^2}
\end{align}
which gives the desired result using \eqref{eq:delta_inner}
\begin{equation}
    (\robotposition-\diffeocenter{j_i}) \nabla \alpha_{j_i}(\robotposition) = \frac{(\robotposition-\diffeocenter{j_i}) \nabla \outerpolygonimplicit{j_i}(\robotposition) - \outerpolygonimplicit{j_i}(\robotposition)}{||\robotposition-\diffeocenter{j_i}||} < 0
\end{equation}
This concludes the proof that $\diffeopurging{j_i}$ satisfies Property 1.

Next, we focus on Property 2. Pick a point $\robotposition \in \partial_k \freespacemappedpurging{j_i}$. This point could lie:
\begin{enumerate}
    \item on the outer boundary of $\freespacemappedpurging{j_i}$ and away from $\knownobstacledilated_i$
    \item on the boundary of one of the $|\unknownobstaclesetdilatedsemanticindex|$ unknown but visible convex obstacles
    \item on the boundary of one of the $(|\knownobstaclesetdilatedmappeddiskindex|+|\knownobstaclesetdilatedmappedintrusionindex|-1)$ familiar obstacles that are not $\knownobstacledilated_i$
    \item on the boundary of $\knownobstacledilated_i$ but not on the boundary of the triangle $j_i$
    \item on the boundary of the triangle $j_i$
\end{enumerate}
In the first four cases, we have $\diffeopurging{j_i}(\robotposition) = \robotposition$, whereas in the last case, we have
\begin{equation}
    \diffeopurging{j_i}(\robotposition) = \diffeocenter{j_i} + \frac{\left(\robotposition_{1j_i} - \diffeocenter{j_i} \right)^\top \sharednormal{j_i}}{\left(\robotposition - \diffeocenter{j_i} \right)^\top \sharednormal{j_i}} (\robotposition-\diffeocenter{j_i})
\end{equation}
It can be verified that $\left( \diffeopurging{j_i}(\robotposition) - \robotposition_{1j_i} \right)^\top \sharednormal{j_i} = 0$, which means that $\robotposition$ is sent to the shared hyperplane between $j_i$ and $p(j_i)$ as desired. This shows that we always have $\diffeopurging{j_i}(\robotposition) \in \partial_k \freespacemappedpurging{p(j_i)}$ and the map satisfies Property 2. 

Finally, Property 3 derives from above and the fact that each boundary segment $\partial_k \freespacemappedpurging{j_i}$ is an one-dimensional manifold, the boundary of either a convex set or a polygon, both of which are homeomorphic to $S^1$ and, therefore, the corresponding boundary $\partial_k \freespacemappedpurging{p(j_i)}$.
\end{proof}

\begin{proof}[Lemma \ref{lemma:singular_root}]
The proof follows similar patterns with the proof of Lemma \ref{lemma:singular_leaf}. With the procedure outlined in Appendix \ref{appendix:implicit}, the only points where $\innerpolygonimplicit{r_i}$ and $\outerpolygonimplicit{r_i}$ are not smooth are vertices of $\innerpolygon{r_i}$ and $\outerpolygon{r_i}$ respectively. Therefore, with the definition of $\outerpolygonsigma{r_i}$ as in \eqref{eq:sigma_delta_ri} and the use of the smooth, non-analytic function $\zeta$ from \eqref{eq:zeta}, we see that $\outerpolygonsigma{r_i}$ is smooth everywhere, since $\diffeocenter{i}$ does not belong in $\freespacemappedhat$ and $\outerpolygonimplicit{r_i}$ is exactly 0 on the vertices of $\outerpolygon{r_i}$. Therefore, $\switch{r_i}$ can only be non-smooth on the vertices of $\innerpolygon{r_i}$ (i.e., on the vertices of the triangle $r_i$), and on points where its denominator becomes zero. Since both $\innerpolygonsigma{r_i}$ and $\outerpolygonsigma{r_i}$ vary between 0 and 1, this can only happen when $\innerpolygonsigma{r_i} (\robotposition) = 1$ and $\outerpolygonsigma{r_i}(\robotposition) = 0$, which is not allowed by Definition \ref{definition:collars_root_disk}, requiring $\innerpolygon{r_i} \subset \outerpolygon{r_i}$. The fact that $\switch{r_i}$ is smooth everywhere else derives immediately from the fact that $\outerpolygonsigma{r_i}$ is a smooth function, and $\innerpolygonsigma{r_i}$ is smooth everywhere except for the triangle vertices. 
\end{proof}

\begin{proof}[Proposition \ref{proposition:diffeo_root}]
The proof follows similar patterns with the proof of Proposition \ref{proposition:diffeo_purging}. As shown in Lemma \ref{lemma:root_smooth}, the map $\diffeoroot$ is smooth in $\freespacemappedhat$ away from any sharp corners. Therefore, we need to focus again on the Massey conditions \cite{massey1992} and show that
\begin{enumerate}
    \item $\diffeoroot$ has a non-singular differential on $\freespacemappedhat$ away from any sharp corners.
    \item $\diffeoroot$ preserves boundaries, i.e., $\diffeoroot(\partial_k \freespacemappedhat) \subset \partial_k \freespacemodel$, with $k$ spanning both the indices of familiar obstacles $\knownobstaclesetdilatedmappeddiskindex$, $\knownobstaclesetdilatedmappedintrusionindex$ as well as the indices of unknown obstacles $\unknownobstaclesetdilatedsemanticindex$.
    \item the boundary components of $\freespacemappedhat$ and $\freespacemodel$ are pairwise homeomorphic, i.e., $\partial_k \freespacemappedhat \cong \partial_k \freespacemodel$, with $k$ spanning both the indices of familiar obstacles $\knownobstaclesetdilatedmappeddiskindex$, $\knownobstaclesetdilatedmappedintrusionindex$ as well as the indices of unknown obstacles $\unknownobstaclesetdilatedsemanticindex$.
\end{enumerate}

We begin with Property 1 and examine the space away from any sharp corners in $\freespacemappedhat$. By construction of the polygonal collars $\outerpolygon{r_i}$ and the definition of $\diffeoroot$ in \eqref{eq:map_root}, we see that $\diffeoroot$ is either the identity map (which implies that $D_\robotposition \diffeoroot = \mathbf{I}$), or depends only on a single switch $\switch{r_k}$. In that case, we can isolate the $k$-th term of the map jacobian to write
\begin{align}
    D_\robotposition\diffeoroot = D_\robotposition\diffeoroot|_k = & \left(\deformingfactor{r_k}(\robotposition)-1\right) (\robotposition-\robotposition_k^*) \nabla \switch{r_k}(\robotposition)^\top \nonumber \\
    & + \switch{r_k}(\robotposition)(\robotposition-\diffeocenter{k}) \nabla \deformingfactor{r_k}(\robotposition)^\top \nonumber \\
    & + \left[ 1 + \switch{r_k}(\robotposition)\left(\deformingfactor{r_k}(\robotposition)-1\right) \right] \mathbf{I} \label{eq:jacobian_root}
\end{align}
It is then straightforward to follow exactly the same procedure outlined in the proof of Proposition \ref{proposition:diffeo_purging} and show that $\text{det}(D_\robotposition\diffeoroot|_k),\text{tr}(D_\robotposition\diffeoroot|_k) > 0$ for all $\robotposition \in \freespacemappedhat$ away from sharp corners. 

Next, we focus on Property 2. Pick a point $\robotposition \in \partial_k \freespacemappedhat$. This point could lie
\begin{enumerate}
    \item on the outer boundary of $\freespacemappedhat$, but not on a root triangle corresponding to an obstacle $\knownobstacledilated_i \in \knownobstaclesetdilatedmappedintrusion$
    \item on the boundary of one of the $\unknownobstaclesetdilatedsemanticindex$ unknown but visible convex obstacles
    \item on the outer boundary of $\freespacemappedhat$ and on a root triangle corresponding to an obstacle $\knownobstacledilated_i \in \knownobstaclesetdilatedmappedintrusion$, or
    \item on the boundary of one of the $|\knownobstaclesetdilatedmappeddiskindex|$ root triangles corresponding to obstacles in $\knownobstaclesetdilatedmappeddisk$
\end{enumerate}
In the first two cases, we have $\diffeoroot(\robotposition) = \robotposition$, in the third case we have $\diffeoroot(\robotposition) \in \partial_0 \freespacemodel = \partial \enclosingfreespace$ by construction of \eqref{eq:sigma_ri} and \eqref{eq:deforming_factor_root_boundary}, while in the last case
\begin{equation}
    \diffeoroot(\robotposition) = \diffeocenter{k} + \frac{\rho_k}{||\robotposition-\diffeocenter{k}||}(\robotposition-\diffeocenter{k})
\end{equation}
for some $k \in \knownobstaclesetdilatedmappeddiskindex$, sending $\robotposition$ to the boundary of the $k$-th disk in $\freespacemodel$. This shows that we always have $\diffeoroot(\robotposition) \in \partial_k \freespacemodel$ and the map satisfies Property 2.

Finally, Property 3 derives from above and the fact that each boundary segment $\partial_k \freespacemappedhat$ is an one-dimensional manifold, the boundary of either a convex set or a triangle, both of which are homeomorphic to $S^1$ and, therefore, the corresponding boundary $\partial_k \freespacemodel$.
\end{proof}

\subsubsection{Proofs of Results in Section \ref{sec:controller}}
\label{appendix:proofs_control}

\begin{proof}[Lemma \ref{lemma:stationary_points}]
The proof of this lemma derives immediately from \cite[Propositions 5,11]{arslan_kod_WAFR2016}, from which we can infer that the set of stationary points of the vector field $D_\robotposition\diffeo \cdot \controlfullyactuated^\hybridmode(\robotposition)$, defined on $\freespacemodel$, is $\{ \diffeo(\goalposition)\} \bigcup \{\mathbf{s}_i\}_{i \in \knownobstaclesetdilatedmappeddiskindex} \bigcup \{\mathcal{G}_k\}_{k \in \unknownobstaclesetdilatedsemanticindex}$, with $\diffeo(\goalposition)$ being a locally stable equilibrium of $D_\robotposition\diffeo \cdot \controlfullyactuated^\hybridmode(\robotposition)$ and each other point being a nondegenerate saddle, since \cite[Assumption 2]{arslan_kod_WAFR2016} is satisfied for the obstacles in $\freespacemodel$. To complete the proof, we just have to note that the index of an isolated zero of a vector field does not change under diffeomorphisms of the domain \cite{hirsch_1976}.
\end{proof}

\begin{proof}[Proposition \ref{proposition:stability_fullyactuated}]
Consider the smooth Lyapunov function candidate $V^{\hybridmode}(\robotposition) = ||\diffeo(\robotposition)-\diffeo(\goalposition)||^2$. Using \eqref{eq:control_fullyactuated} and writing $\robotpositionmodel = \diffeo(\robotposition)$ and $\goalpositionmodel = \diffeo(\goalposition)$, we get
\begin{align}
\frac{dV^{\hybridmode}}{dt} = & 2(\robotpositionmodel-\goalpositionmodel)^\top(\mathbf{D}_\robotposition\diffeo)\dot{\robotposition} \nonumber \\
= & -2k(\robotpositionmodel-\goalpositionmodel)^\top \left(\robotpositionmodel - \projection{\localfreespace{\robotpositionmodel}}{\goalpositionmodel} \right) \nonumber \\
= & -2k\left(\robotpositionmodel-\projection{\localfreespace{\robotpositionmodel}}{\goalpositionmodel}+\projection{\localfreespace{\robotpositionmodel}}{\goalpositionmodel}-\goalpositionmodel \right)^\top \nonumber \\ & \left(\robotpositionmodel - \projection{\localfreespace{\robotpositionmodel}}{\goalpositionmodel} \right) \nonumber \\
= & -2k || \robotpositionmodel-\projection{\localfreespace{\robotpositionmodel}}{\goalpositionmodel} ||^2 \nonumber \\
& +2k \left( \goalpositionmodel - \projection{\localfreespace{\robotpositionmodel}}{\goalpositionmodel} \right)^\top \left(\robotpositionmodel - \projection{\localfreespace{\robotpositionmodel}}{\goalpositionmodel} \right) \nonumber \\
\leq & -2k || \robotpositionmodel-\projection{\localfreespace{\robotpositionmodel}}{\goalpositionmodel} ||^2 \leq 0
\end{align}
since $\robotpositionmodel \in \localfreespace{\robotpositionmodel}$, which implies that 
\begin{equation}
\left( \goalpositionmodel - \projection{\localfreespace{\robotpositionmodel}}{\goalpositionmodel} \right)^\top\left(\robotpositionmodel - \projection{\localfreespace{\robotpositionmodel}}{\goalpositionmodel} \right) \leq 0
\end{equation}
since either $\goalpositionmodel = \projection{\localfreespace{\robotpositionmodel}}{\goalpositionmodel}$, or $\goalpositionmodel$ and $\robotpositionmodel$ are separated by a hyperplane passing through $\projection{\localfreespace{\robotpositionmodel}}{\goalpositionmodel}$. Therefore, similarly to \cite{arslan_kod_WAFR2016}, using LaSalle's invariance principle we see that every trajectory starting in $\freespacemapped$ approaches the largest invariant set in $\{ \robotposition \in \freespacemapped \, | \, \dot{V}^{\hybridmode}(\robotposition)=0 \}$, i.e. the equilibrium points of \eqref{eq:control_fullyactuated}. The desired result follows from Lemma \ref{lemma:stationary_points}, since $\goalposition$ is the only locally stable equilibrium of our control law and the rest of the stationary points are nondegenerate saddles, whose regions of attraction have empty interior in $\freespacemapped$.
\end{proof}

\begin{proof}[Proposition \ref{proposition:diffeo_unicycle}]
Note that the jacobian of $\diffeounicycle$ will be given by
\begin{equation}
D_{\robotpositionunicycle} \diffeounicycle = \left[ \begin{matrix}
D_\robotposition\diffeo & \vline & \mathbf{0}_{2 \times 1} \\ \hline D_\robotposition\angletransform & \vline & \dfrac{\partial \angletransform}{\partial \robotorientation}
\end{matrix} \right] \label{eq:differential_se2}
\end{equation}
Since we already have from Corollary \ref{corollary:diffeo_full} that $D_\robotposition\diffeo$ is non-singular, it suffices to show that $\frac{\partial \angletransform}{\partial \robotorientation} \neq 0$ for all $\robotpositionunicycle \in \freespacemapped \times S^1$. From \eqref{eq:phi} we can derive
\begin{equation}
\frac{\partial \angletransform}{\partial \robotorientation} = \frac{\text{det}(D_\robotposition\diffeo)}{||\directionvector(\robotpositionunicycle)||^2}
\end{equation}
Therefore, we immediately get that $\frac{\partial \angletransform}{\partial \robotorientation} \neq 0$ for all $\robotpositionunicycle \in \freespacemapped \times S^1$ since $\text{det}(D_\robotposition\diffeo) \neq 0$ and $||\directionvector(\robotpositionunicycle)|| \neq 0$ for all $\robotposition \in \freespacemapped$, because $D_\robotposition\diffeo$ is non-singular on $\freespacemapped$. This implies that $D_{\robotpositionunicycle}\diffeounicycle$ is non-singular on $\freespacemapped \times S^1$.

Next, we note that $\partial\left(\freespacemapped \times S^1 \right) = \partial \freespacemapped \times S^1$, since $S^1$ is a manifold without boundary. Similarly, $\partial\left(\freespacemodel \times S^1 \right) = \partial \freespacemodel \times S^1$. Hence, we can easily complete the proof following a similar procedure with the end of the proofs of Propositions \ref{proposition:diffeo_purging} and \ref{proposition:diffeo_root} to show that $\diffeounicycle$ preserves boundaries, and the boundaries of $\freespacemapped \times S^1$ and $\freespacemodel \times S^1$ are pairwise homeomorphic.
\end{proof}

\begin{proof}[Theorem \ref{theorem:control_se2}]
We have already established that $||\directionvector(\robotpositionunicycle)||$ and $\frac{\partial \angletransform}{\partial \robotorientation}$ are nonzero for all $\robotpositionunicycle \in \freespacemapped \times S^1$ in the proof of Proposition \ref{proposition:diffeo_unicycle}, which implies that $\linearinput$ and $\angularinput$ can have no singular points. Also notice that $||\directionvector(\robotpositionunicycle)||$, $\frac{\partial \angletransform}{\partial \robotorientation}$ and $D_\robotposition\angletransform\begin{bmatrix}
\cos\robotorientation & \sin\robotorientation
\end{bmatrix}^\top$ are all smooth away from corners in $\freespacemapped \times S^1$. Hence, the uniqueness and existence of the flow generated by control law \eqref{eq:control_unicycle} can be established similarly to \cite{arslan_kod_WAFR2016} through the flow properties of the controller in \cite{astolfi_1999} (that we use here in \eqref{eq:control_unicycle_reference}) and the facts that metric projections onto moving convex cells are piecewise continuously differentiable \cite{Kuntz-1994,shapiro-1988}, and the composition of piecewise continuously differentiable functions is piecewise continuously differentiable and, therefore, locally Lipschitz \cite{chaney-1990}.

Next, as shown in \eqref{eq:se2_pushforward}, the vector field $\mathbf{B}(\robotorientation)\controlunicycle^\hybridmode$ on $\freespacemapped \times S^1$ is the pullback of the complete vector field $\mathbf{B}(\robotorientationmodel)\controlunicyclemodel^\hybridmode$, guaranteed to retain $\freespacemodel \times S^1$ positively invariant under its flow as shown in \cite{arslan_kod_WAFR2016}, under the smooth change of coordinates $\diffeounicycle$ away from sharp corners in $\freespacemapped \times S^1$. This shows that the freespace $\freespacemapped \times S^1$ is positively invariant under law \eqref{eq:control_unicycle}.

Finally, consider the smooth Lyapunov function candidate $V^{\hybridmode}(\robotposition) = ||\diffeo(\robotposition)-\diffeo(\goalposition)||^2$. Then, by writing $\robotpositionmodel = \diffeo(\robotposition)$ and $\goalpositionmodel = \diffeo(\goalposition)$, we get
\begin{align}
\frac{dV^{\hybridmode}}{dt} = & 2(\robotpositionmodel-\goalpositionmodel)^\top(\mathbf{D}_\robotposition\diffeo)\dot{\robotposition} \nonumber \\
= &  2 \linearinput^\hybridmode \, (\robotpositionmodel-\goalpositionmodel)^\top(\mathbf{D}_\robotposition\diffeo) \begin{bmatrix}
\cos\robotorientation \\ \sin\robotorientation
\end{bmatrix} \nonumber \\
= & 2 \linearinputmodel^\hybridmode (\robotpositionmodel-\goalpositionmodel)^\top \begin{bmatrix} \cos\angletransform(\robotpositionunicycle) \\ \sin\angletransform(\robotpositionunicycle)
\end{bmatrix} \nonumber \\
= & -2k_v (\robotpositionmodel-\goalposition)^\top \begin{bmatrix} \cos\angletransform(\robotpositionunicycle) \\ \sin\angletransform(\robotpositionunicycle)
\end{bmatrix} \begin{bmatrix} \cos\angletransform(\robotpositionunicycle) \\ \sin\angletransform(\robotpositionunicycle)
\end{bmatrix}^\top \nonumber \\ &\left(\robotpositionmodel-\projection{\localfreespace{\robotpositionmodel} \cap H_\parallel}{\goalpositionmodel} \right) \nonumber \\
= & -2k_v (\robotpositionmodel-\goalpositionmodel)^\top \left(\robotpositionmodel-\projection{\localfreespace{\robotpositionmodel} \cap H_\parallel}{\goalpositionmodel} \right) \nonumber
\end{align}
since $\begin{bmatrix} \cos\angletransform(\robotpositionunicycle) \\ \sin\angletransform(\robotpositionunicycle)
\end{bmatrix} \begin{bmatrix} \cos\angletransform(\robotpositionunicycle) \\ \sin\angletransform(\robotpositionunicycle)
\end{bmatrix}^\top$ is just the projection operator on the line defined by the vector $\begin{bmatrix} \cos\angletransform(\robotpositionunicycle) & \sin\angletransform(\robotpositionunicycle)
\end{bmatrix}^\top$, with which $\left(\robotpositionmodel-\projection{\localfreespace{\robotpositionmodel} \cap H_\parallel}{\goalpositionmodel} \right)$ is already parallel. Following this result, we get
\begin{equation}
\frac{dV^{\hybridmode}}{dt} \leq -2k_v \Big|\Big|\diffeo(\robotpositionmodel)-\projection{\localfreespace{\robotpositionmodel} \cap H_\parallel}{\goalpositionmodel}\Big|\Big|^2 \leq 0
\end{equation}
since, similarly to the proof of Proposition \ref{proposition:stability_fullyactuated}, we have
\begin{equation}
\left(\goalpositionmodel - \projection{\localfreespace{\robotpositionmodel} \cap H_\parallel}{\goalpositionmodel} \right)^\top \left(\robotpositionmodel-\projection{\localfreespace{\robotpositionmodel} \cap H_\parallel}{\goalpositionmodel} \right) \leq 0
\end{equation}
Therefore, using LaSalle's invariance principle, we see that every trajectory starting in $\freespacemapped \times S^1$ approaches the largest invariant set in $\{ (\robotposition,\robotorientation) \in \freespacemapped \times S^1 \, | \, \dot{V}^{\hybridmode}(\robotposition)=0 \} = \{ (\robotposition,\robotorientation) \in \freespacemapped \times S^1 \, | \, \diffeo(\robotposition) = \projection{\localfreespace{\diffeo(\robotposition)} \cap H_\parallel}{\diffeo(\goalposition)} \}$. At the same time, we know from \eqref{eq:control_unicycle_reference} that $\diffeo(\robotposition) = \projection{\localfreespace{\diffeo(\robotposition)} \cap H_\parallel}{\diffeo(\goalposition)}$ implies $\linearinput^\hybridmode=0$. From \eqref{eq:control_unicycle}, for $\linearinput^\hybridmode=0$, we get that $\angularinput^\hybridmode$ will be zero at points where $\angularinputmodel^\hybridmode$ is zero, i.e. at points $(\robotposition,\robotorientation) \in \freespacemapped \times S^1$ where
\begin{equation}
\begin{bmatrix}
-\sin\angletransform(\robotpositionunicycle) \\ \cos\angletransform(\robotpositionunicycle)
\end{bmatrix}^\top \left( \diffeo(\robotposition) - \mathbf{y}_{d,G}(\diffeo(\robotposition),\angletransform(\robotorientation) \right) = 0
\end{equation}
with the angular local goal $\mathbf{y}_{d,G}$ defined as in \eqref{eq:angularlocalgoal}. Therefore the largest invariant set in $\{ (\robotposition,\robotorientation) \, | \, \diffeo(\robotposition) = \projection{\localfreespace{\diffeo(\robotposition)} \cap H_\parallel}{\diffeo(\goalposition)} \}$ is the set of points $\robotpositionunicycle = (\robotposition,\robotorientation)$ where the following two conditions are satisfied
\begin{align}
&  \diffeo(\robotposition) = \projection{\localfreespace{\diffeo(\robotposition)} \cap H_\parallel}{\diffeo(\goalposition)} \\
& \begin{bmatrix}
-\sin\angletransform(\robotpositionunicycle) \\ \cos\angletransform(\robotpositionunicycle)
\end{bmatrix}^\top \left( \diffeo(\robotposition) - \robotpositionmodel_{d,G}(\diffeo(\robotposition),\angletransform(\robotorientation)) \right) = 0
\end{align}
Using a similar argument to \cite[Proposition 12]{arslan_kod_WAFR2016}, we can, therefore, verify that the set of stationary points of law \eqref{eq:control_unicycle} is given by 
\begin{align}
& \{\goalposition\} \times (-\pi,\pi] \nonumber \\
&\bigcup \left\{(\mathbf{q},\robotorientation) \, \Big | \, \mathbf{q} \in \{(\diffeo)^{-1}(\mathbf{s}_i)\}_{i \in \knownobstaclesetdilatedmappeddiskindex} \bigcup_{k \in \unknownobstaclesetdilatedsemanticindex} \mathcal{G}_k, \right. \nonumber \\
& \left. \begin{bmatrix}
-\sin\angletransform(\mathbf{q},\robotorientation) \\ \cos\angletransform(\mathbf{q},\robotorientation)
\end{bmatrix}^\top(\mathbf{q}-\goalposition) = 0 \right \} 
\end{align}
using \eqref{eq:saddles}. We can then invoke a similar argument to Proposition \ref{proposition:stability_fullyactuated} to show that $\goalposition$ locally attracts with any orientation $\robotorientation$, while any configuration associated with any other equilibrium point is a nondegenerate saddle whose stable manifold is a set of measure zero, and the result follows.
\end{proof}

\begin{proof}[Lemma \ref{lemma:disjoint_guards}]
We can show this by contradiction. Assume that the robot is in mode $\hybridmode$ and two guards $\hybridguardrestriction^{\hybridmode,\hybridmode \cup \hybridmode_1}$ and $\hybridguardrestriction^{\hybridmode,\hybridmode \cup \hybridmode_2}$, indexed by two different subsets $\hybridmode_1 \neq \hybridmode_2$, each playing the role of $\hybridmode_u$ in \eqref{eq:restriction_guard}, nevertheless overlap, $\hybridguardrestriction^{\hybridmode,\hybridmode \cup \hybridmode_1} \cap \hybridguardrestriction^{\hybridmode,\hybridmode \cup \hybridmode_2} \neq \varnothing$. That means that there exists at least one state $\robotposition \in \hybridfreespacemode$, such that $\robotposition \in \hybridguardrestriction^{\hybridmode,\hybridmode \cup \hybridmode_1}$ and $\robotposition \in \hybridguardrestriction^{\hybridmode,\hybridmode \cup \hybridmode_2}$, for two nonempty sets $\hybridmode_1, \hybridmode_2$ with $\hybridmode \cap \hybridmode_1 = \varnothing$, $\hybridmode \cap \hybridmode_2 = \varnothing$. Since $\hybridmode_1 \neq \hybridmode_2$ by assumption, this implies that there exists at least one index $i$ that is contained in one of these index sets, but is not contained in the other. Without loss of generality, assume that $i \in \hybridmode_1$ and $i \notin \hybridmode_2$. We immediately arrive at a contradiction, since the requirement $\robotposition \in \hybridguardrestriction^{\hybridmode,\hybridmode \cup \hybridmode_2}$ requires $\ballclosure{\robotposition}{\sensorrange} \cap \knownobstacleset_{\knownobstaclesetindex \backslash (\hybridmode \cup \hybridmode_2)} = \varnothing$, but we know that the requirement $\robotposition \in \hybridguardrestriction^{\hybridmode,\hybridmode \cup \hybridmode_1}$ implies $\ballclosure{\robotposition}{\sensorrange} \cap \knownobstacle_i \neq \varnothing$ with $\knownobstacle_i \in \knownobstacleset_{\knownobstaclesetindex \backslash (\hybridmode \cup \hybridmode_2)} $.
\end{proof}

\begin{proof}[Lemma \ref{lemma:non_blocking}]
If the system is in the terminal mode $\hybridmode = \knownobstaclesetindex$, according to Definition \ref{definition:terminal_mode}, then finite time escape through the boundary of the hybrid domain is not possible, since the vector field $\controlfullyactuated^\hybridmode$ leaves its domain positively invariant under its flow, as described in Theorem \ref{theorem:control_fullyactuated}. For $\hybridmode \neq \knownobstaclesetindex$, the only way in which the flow can escape is through the boundary of an obstacle $\knownobstacle \notin \{\knownobstacle_i\}_{i \in \hybridmode}$, since $\controlfullyactuated^\hybridmode$ guarantees safety only against familiar obstacles in $\hybridmode$ and any unknown obstacles encountered along the way. We are going to show that this cannot happen by contradiction. Assume that at time $t_0$ the robot is at $\robotposition_0 \in \hybridfreespacemode$, and at time $t_1 > t_0$ it crosses the boundary of an obstacle $\knownobstacle \notin \{\knownobstacle_i\}_{i \in \hybridmode}$. This means that the robot travels distance $d>0$ between $t_0$ and $t_1$ in mode $\hybridmode$, without triggering a transition to another hybrid mode $\hybridmode'$ that includes $\knownobstacle$ (and therefore guarantees safety against it by Theorem \ref{theorem:control_fullyactuated}), which is impossible since the sensor footprint has a positive radius $\sensorrange$ and $\ballclosure{\robotposition^t}{\sensorrange}$ would have hit $\knownobstacle_i$ at some time $t < t_1$ before colliding with it.

Moreover, the restriction of the reset map in each separate mode is just the identity transform, which, by the argument made above, implies that the discrete transition itself is never blocking, assuming that the initial condition lies in the freespace $\freespace$. This is because $\freespace \subseteq \hybridfreespacemode$ for all modes $\hybridmode \in 2^{\knownobstaclesetindex}$.

Finally, hybrid ambiguity is avoided by the construction of the guard in \eqref{eq:restriction_guard}; if the robot at a position $\robotposition^{-}$ in the interior of the domain is in mode $\hybridmode$ at time $t^{-}$ before a discrete transition and in mode $\hybridmode'$ at time $t^{+}$ after the transition, we are guaranteed that the sensor footprint $\ballclosure{\robotposition^{+}}{\sensorrange}$ after the transition does not intersect any obstacle $\knownobstacle_i$ with $i \notin \hybridmode'$. This implies that $\robotposition^{+}$ lies in the interior of the domain and away from the guard, and the application of the reset map provides the unique extension to the execution.
\end{proof}

\begin{proof}[Theorem \ref{theorem:hybrid_fullyactuated}]
As stated in Section \ref{subsec:control_hybrid}, the hybrid system described in the tuple $\hybridsystem$ has $2^{|\knownobstaclesetindex|}$ modes. Unique piecewise continuous differentiability of the flow derives immediately by the unique continuous differentiability of the control law $\hybridfieldrestriction^\hybridmode$, defined in \eqref{eq:control_fullyactuated} for each separate mode $\hybridmode$, as summarized in Theorem \ref{theorem:control_fullyactuated}. Moreover, positive invariance derives from the first part of the proof of Lemma \ref{lemma:non_blocking}, which guarantees that the hybrid flow cannot escape from the hybrid domain through a point on the boundary of the domain in the complement of the guard, or the guard itself.

For stability, we note that each mode (indexed by $\hybridmode \in 2^{\knownobstaclesetindex}$) is associated with a candidate Lyapunov function $V^\hybridmode(\robotposition) = ||\diffeogeneric^\hybridmode(\robotposition)-\diffeogeneric^\hybridmode(\goalposition)||^2$, as shown in the proof of Proposition \ref{proposition:stability_fullyactuated}. Moreover, also by the results of Proposition \ref{proposition:stability_fullyactuated}, $\goalposition$ is the unique asymptotically stable equilibrium of each control vector field $\hybridfieldrestriction^{\hybridmode}$, thus, almost every execution that remains in mode $\hybridmode$ for all future time has a trajectory that asymptotically approaches the goal. Then, the key for the proof is the observation that once the robot exits a mode defined by $\hybridmode$, it can never re-enter it. This is because the robot stores information in its semantic map and this knowledge can only be incremental; in the worst case, the robot will explore all familiar obstacles in the environment, and stay in mode $\hybridmode = \knownobstaclesetindex$ for all following time.

Based on this observation, we notice that the collection of functions $\{ V^{\hybridmode} \, | \, \hybridmode \in 2^{\knownobstaclesetindex} \}$ are {\it Lyapunov-like}, in the sense of \cite[Definition 2.2]{branicky_1998}, for all time their corresponding mode is active, since they never reset. We complete the proof by invoking \cite[Theorem 2.3]{branicky_1998}, which states that if a collection of Lyapunov-like functions for a hybrid system are associated with corresponding vector fields that share the same equilibrium, then the hybrid system itself is Lyapunov stable around this equilibrium.
\end{proof}

\end{document}